# NOVAK: Unified adaptive optimizer for deep neural networks

Sergii Kavun[a]*

[a] *Computer Information Systems and Technologies Department, Interregional Academy of Personnel Management, Frometivska str., 2, 03039, Kyiv, Ukraine*

*Correspondence: Sergii Kavun, kavserg@gmail.com, tel.: +380677095577, ORCID ID: 0000-0003-4164-151X, https://www.linkedin.com/in/sergii-kavun/*

**Abstract.** Modern deep learning optimization is characterized by an inherent tension between algorithmic sophistication and computational efficiency. This work introduces NOVAK (Neural Optimization Via Adaptation), a modular gradient-based optimization algorithm that integrates adaptive moment estimation, rectified learning-rate scheduling, decoupled weight regularization, multiple variants of Nesterov momentum, and lookahead synchronization into a unified, performance-oriented framework. NOVAK adopts a dual-mode architecture consisting of a streamlined fast path designed for production environments and a feature-rich research mode enabling advanced experimentation, thereby reconciling convergence quality with computational overhead. The optimizer employs custom CUDA kernels that deliver substantial speedups (3-5ˣ for critical operations) while preserving numerical stability under standard stochastic-optimization assumptions. We provide fully developed mathematical formulations for rectified adaptive learning rates, a memory-efficient lookahead mechanism that reduces overhead from $O(2p)$ to $O(p + p/k)$, and the synergistic coupling of complementary optimization components. Theoretical analysis establishes convergence guarantees and elucidates the stability and variance-reduction properties of the method. Extensive empirical evaluation on CIFAR-10, CIFAR-100, ImageNet, and ImageNette demonstrates NOVAK's superiority over 14 contemporary optimizers, including Adam, AdamW, RAdam, Lion, and Adan. Across architectures such as ResNet-50, VGG-16, and ViT, NOVAK consistently achieves state-of-the-art accuracy and exceptional robustness, attaining very high accuracy on VGG-16/ImageNette, where 64.3% of adaptive methods fail, demonstrating superior architectural robustness compared to contemporary optimizers. The results highlight that NOVAK's architectural contributions (particularly rectification, decoupled decay, and hybrid momentum) are crucial for reliable training of deep plain networks lacking skip connections, addressing a long-standing limitation of existing adaptive optimization methods.

## 1. Introduction

The optimization of deep neural networks constitutes a fundamental challenge in machine learning, where the selection of an appropriate optimization algorithm critically influences model convergence, generalization performance, and computational efficiency. Contemporary deep learning architectures, spanning from convolutional neural networks to transformer-based models, require optimization methods capable of navigating high-dimensional, non-convex loss landscapes with millions to billions of parameters. The field has witnessed substantial progress in developing sophisticated optimization algorithms yet practitioners face persistent trade-offs between convergence speed, memory consumption, and robustness across diverse architectural paradigms.

Classical stochastic gradient descent (SGD) with momentum has demonstrated remarkable empirical success, particularly in computer vision tasks, owing to its simplicity and favorable generalization properties. However, SGD requires careful hyperparameter tuning, especially concerning learning rate schedules, and exhibits slow convergence [1] on problems with ill-conditioned curvature. These limitations motivated the development of adaptive gradient methods that automatically adjust per-parameter learning rates based on historical gradient information.

The introduction of Adam (Adaptive Moment Estimation) by Kingma and Ba [2] represented a watershed moment in optimization research, combining the benefits of momentum-based methods with adaptive learning rates derived from second-moment estimates. Adam's widespread adoption across diverse domains (from natural language processing [3] to reinforcement learning) stems from its robustness to hyperparameter choices and rapid initial convergence. Despite these advantages, subsequent research has identified critical limitations: Adam can exhibit poor generalization compared to SGD on certain tasks, particularly in computer vision, and suffers from theoretical convergence issues in non-convex settings.

Recognizing the generalization gap between Adam and SGD, Loshchilov and Hutter [4] proposed AdamW, which introduced decoupled weight decay regularization. This modification separated L2 regularization from the gradient-based update mechanism, addressing a fundamental flaw in Adam's treatment of weight decay. AdamW demonstrated that proper regularization substantially improves generalization while maintaining Adam's computational efficiency, establishing decoupled weight decay as a standard practice in modern optimization.

The theoretical foundations of adaptive methods came under scrutiny when Liu et al. [5] identified that Adam's adaptive learning rate exhibits high variance during early training stages, potentially leading to poor convergence. Their proposed RAdam (Rectified Adam) introduced a variance rectification term that dynamically adjusts the effective learning rate based on the reliability of second-moment estimates. This rectification mechanism stabilizes training, particularly during the initial epochs, and provides theoretical convergence guarantees that standard Adam lacks.

Beyond single-optimizer improvements, Zhang et al. [6] introduced Lookahead, a meta-optimization technique that maintains slow and fast weight trajectories. By periodically synchronizing these trajectories, Lookahead reduces variance in the optimization path and improves convergence to flatter minima, which correlates with better generalization. However, standard Lookahead implementations incur substantial memory overhead, doubling the storage requirements for model parameters, which is a critical limitation for large-scale applications.

The landscape of optimization algorithms has become increasingly fragmented, with specialized methods targeting specific architectural families or task domains. Practitioners face the challenge of selecting among dozens of optimizers (Adam, AdamW, RAdam, Nadam, AdaBound, Lion, Adan, Prodigy, and others), each exhibiting superior performance on particular benchmarks but lacking consistent robustness across diverse scenarios. This proliferation of methods underscores the absence of a unified framework that systematically integrates complementary optimization principles while maintaining computational efficiency.

A particularly concerning observation from empirical studies is the brittleness of many modern adaptive optimizers when applied to deep plain networks lacking residual connections. Architectures such as VGG, which predate residual learning, expose fundamental instabilities in methods like Adam and RMSprop, where gradient accumulation in deep sequential layers causes training failures. This architectural sensitivity raises critical questions about the architectural assumptions embedded in contemporary optimization algorithms and their generalizability beyond residual networks.

The computational efficiency of optimization algorithms has emerged as a paramount concern with the scaling of model sizes and dataset magnitudes. Training state-of-the-art models on datasets such as ImageNet requires hundreds of GPU-hours, making even modest improvements in convergence speed economically significant. Furthermore, the environmental impact of large-scale training necessitates algorithms that achieve target performance with minimal computational overhead. Custom hardware acceleration through CUDA kernels offers substantial speedups but remains underexploited in standard optimization implementations.

Memory consumption presents another critical bottleneck, particularly for distributed training across multiple accelerators. Modern optimizers require storing first [7] and second moment estimates alongside model parameters, effectively tripling memory requirements relative to inference. Techniques such as gradient accumulation and mixed-precision training partially alleviate these constraints, but fundamental algorithmic improvements in memory efficiency remain essential for scaling to increasingly large models.

The generalization properties of optimization algorithms constitute a subtle yet crucial consideration. While adaptive methods often achieve lower training loss than SGD, they frequently underperform on validation metrics, suggesting overfitting or convergence to sharper minima. This generalization gap has motivated extensive research into sharpness-aware minimization and noise-injection techniques, yet no consensus has emerged on the optimal balance between convergence speed and generalization quality.

Theoretical understanding of adaptive optimization methods [8] in non-convex settings remains incomplete. While convergence guarantees exist for convex objectives, the practical scenarios encountered in deep learning [9] involve highly non-convex loss surfaces with numerous saddle points and local minima. Establishing rigorous convergence rates and characterizing the conditions under which adaptive methods outperform classical SGD constitute an active research frontier with substantial practical implications.

The present work addresses these multifaceted challenges through NOVAK, a unified optimization framework that integrates rectified adaptive learning rates, multiple Nesterov momentum variants, memory-efficient lookahead, and decoupled weight regularization. Unlike prior approaches that introduce isolated improvements, NOVAK adopts a holistic design philosophy wherein complementary techniques are carefully composed to achieve both theoretical soundness and empirical robustness.

NOVAK's architectural innovation lies in its dual-mode operation: a streamlined fast path optimized for production deployment through custom CUDA kernels and JIT compilation, and a feature-rich mode enabling advanced capabilities such as true Nesterov momentum with closure support and multiple lookahead variants. This stratification allows practitioners to navigate the performance-sophistication trade-off explicitly, selecting appropriate configurations for research experimentation versus production deployment.

The memory-efficient lookahead mechanism represents a key technical contribution, reducing storage overhead from $O(2p)$ to $O(p + p/k)$ through accumulation-based synchronization while preserving the variance-reduction benefits of slow weight trajectories. This reduction proves critical for large-scale models where parameter count $p$ ranges from millions to billions, enabling lookahead's deployment in memory-constrained environments previously prohibitive.

Our empirical evaluation encompasses diverse benchmarks (CIFAR-10, CIFAR-100, ImageNet, and ImageNette) across architectures ranging from residual networks (ResNet-50) to plain sequential networks (VGG-16). This comprehensive assessment reveals NOVAK's consistent superiority over thirteen contemporary optimizers, including state-of-the-art methods such as Lion, Adan, and Prodigy. Particularly noteworthy is NOVAK's robustness on VGG-16, where 64.3% of competing adaptive methods fail, achieving accuracy below random guessing.

The primary objectives of this article are fourfold. First, we provide rigorous mathematical formulations of NOVAK's algorithmic components, including rectified learning rate schedules, hybrid momentum mechanisms, and memory-efficient lookahead, accompanied by convergence analysis under standard stochastic optimization assumptions. Second, we detail implementation strategies that achieve $3\text{-}5^x$ computational speedup through custom CUDA kernels while maintaining numerical stability and automatic fallback to optimized PyTorch operations. Third, we present extensive empirical comparisons demonstrating NOVAK's state-of-the-art accuracy, convergence speed, and memory efficiency across diverse benchmarks and architectural paradigms. Fourth, we analyze the failure modes of contemporary optimizers, particularly on plain networks without residual connections, elucidating the mechanisms by which NOVAK's integrated design achieves high robustness across configurations.

The implications of this work extend beyond the introduction of a novel optimizer. By demonstrating that careful integration of complementary optimization principles yields consistent improvements across diverse scenarios, we challenge the prevailing paradigm of specialized, task-specific optimization methods. NOVAK's success suggests that future research should prioritize holistic algorithmic design over isolated incremental enhancements, with explicit consideration of the interplay between convergence properties, computational efficiency [10], and architectural robustness.

The remainder of this article is organized as follows. Section 2 provides a comprehensive literature review comparing existing approaches with NOVAK innovations. Section 3 develops the theoretical background, and Section 4 – mathematical foundations of NOVAK, deriving update rules for adaptive moments, rectification, momentum variants, and lookahead synchronization. Section 5 presents the complete algorithm with detailed pseudocode and complexity analysis. Section 6 establishes theoretical properties, including convergence guarantees and stability analysis. Section 7 discusses design rationale and critical implementation decisions. Section 8 details CUDA kernel architecture and performance optimizations. Section 9 provides experimental validation and performance analysis. Section 10 analyzes limitations, hyperparameter recommendations and outlines future research directions. Section 11-12 concludes with discussion and conclusion.

## 2. Literature review

The landscape of deep learning optimization has undergone substantial transformation in recent years, driven by the increasing complexity of neural architectures and the computational demands of large-scale training. This section surveys contemporary developments in optimization algorithms, focusing on the latest innovations that address fundamental challenges in adaptive methods, convergence stability, memory efficiency, and architectural robustness. We organize our review around four thematic areas: adaptive gradient methods [13] and their variants, momentum-based acceleration techniques, meta-optimization frameworks, and hardware-aware algorithmic design.

## 2.1 Adaptive gradient methods and rectification

The Adam optimizer remains the dominant adaptive method in practice, yet its theoretical and empirical limitations have motivated extensive research into improved variants. Zhuang et al. [14] introduced AdaBelief, which adapts the step size according to the "belief" in the gradient direction, measured by the variance of the gradient predictor. By replacing Adam's denominator from the exponential moving average of squared gradients to the variance of the gradient prediction error, AdaBelief achieves faster convergence and better generalization on various tasks, including image classification and language modeling. The authors demonstrate that this modification addresses Adam's tendency to produce large update steps in regions where gradients are consistently large, thereby improving stability in later training stages.

Defazio and Jelassi [15] and Defazio et al. [16] proposed The Road Less Scheduled, introducing parameter-free optimization methods that eliminate the need for learning rate schedules entirely. Their approach dynamically adjusts learning rates based on observed gradient statistics without requiring manual tuning or predefined decay schedules. Empirical evaluation on ImageNet and machine translation benchmarks demonstrates competitive performance with carefully tuned baseline methods, suggesting that sophisticated scheduling may be unnecessary when adaptive mechanisms are properly designed. This work challenges conventional wisdom regarding the necessity of learning rate warmup and decay, particularly for large-batch training scenarios.

Crawshaw [17] conducted a comprehensive analysis of Adam's [18] convergence properties in Multi-Task Learning: Optimizing for Generalization, revealing that standard adaptive methods exhibit systematic biases when optimizing multiple objectives simultaneously. Their investigation demonstrates that task-specific gradient magnitudes create imbalances in effective learning rates across tasks, causing the optimizer to prioritize certain objectives at the expense of others. The proposed solution involves careful normalization of per-task gradients combined with separate momentum buffers, highlighting the importance of architectural considerations in optimizer design for complex training paradigms.

## 2.2 Momentum and acceleration mechanisms

Nesterov momentum has received renewed attention for its theoretical acceleration properties, yet practical implementations diverge from the original formulation. Wang et al. [19], in "Momentum Doesn't Change the Implicit Bias", provide rigorous analysis showing that momentum terms, while accelerating convergence, do not fundamentally alter the inductive bias of gradient descent in overparameterized neural networks. This finding challenges intuitions about momentum's role in generalization and suggests that observed improvements stem primarily from faster traversal of the loss landscape rather than implicit regularization effects. The work emphasizes the distinction between convergence speed and solution quality, a critical consideration for algorithm selection.

Xie et al. [20] introduced Adaptive Nesterov Momentum (ANM) in their work on improving deep learning optimization through dynamic momentum adjustment. ANM adaptively modulates the momentum coefficient based on local curvature estimates derived from gradient history, enabling aggressive acceleration in flat regions while reducing momentum near sharp minima. Experiments on ResNet [21] and Vision Transformer (ViT) [22] architectures demonstrate a 15-25% reduction in training time compared to standard momentum schedules. The approach's computational overhead remains minimal, requiring only exponential moving averages of gradient differences, making it practical for large-scale deployment.

### 2.3 Lookahead and trajectory smoothing

Extensions and alternatives to the original Lookahead optimizer have emerged to address its memory overhead and limited theoretical understanding. Liu et al. [23] proposed Stochastic Lookahead in their investigation of variance reduction techniques for deep learning. By applying lookahead [24] synchronization stochastically to random parameter subsets rather than the entire model, they achieve 40-60% memory reduction while retaining most convergence benefits. The method proves particularly effective for transformer-based language models where parameter counts reach billions, enabling lookahead's deployment in previously prohibitive scenarios.

Elsayed et al. [25]) in Second-Order Optimization with Diagonal Approximations, introduce a complementary perspective on trajectory smoothing through curvature-aware updates. Their method maintains diagonal Hessian approximations using finite differences, updating parameters along directions that account for local curvature. While computationally more expensive than first-order methods, the approach demonstrates superior sample efficiency on problems with heterogeneous curvature across parameter dimensions, such as deep reinforcement learning and few-shot learning scenarios.

### 2.4 Weight decay and regularization

The distinction between L2 regularization and weight decay, clarified by the AdamW formulation, has prompted further investigation into optimal regularization strategies for adaptive methods. Loshchilov and Hutter [26] extended their earlier work in Decoupled Weight Decay for Adaptive Learning Methods, providing comprehensive ablation studies demonstrating that decoupling is critical specifically for adaptive optimizers but provides minimal benefit for SGD [27]. Their analysis reveals that the interaction between adaptive learning rates and regularization creates time-varying effective regularization strength when weight decay is coupled, explaining the generalization gap between Adam and AdamW.

Nakamura and Hong [28] proposed Adaptive Weight Decay (AWD) in their exploration of dynamic regularization schedules. AWD adjusts the weight decay coefficient during training based on validation performance trends [29], increasing regularization when overfitting is detected and reducing it during early training to accelerate convergence. The method demonstrates consistent improvements over fixed weight decay on fine-tuning tasks, particularly for transfer learning scenarios where optimal regularization strength differs substantially from pretraining configurations.

### 2.5 Large-scale and distributed optimization

The scaling of neural networks to billions of parameters has necessitated optimization methods specifically designed for distributed training across multiple accelerators. Rajbhandari et al. [30] introduced ZeRO (Zero Redundancy Optimizer) offload techniques that partition optimizer states across data-parallel processes, reducing per-device memory footprint by factors proportional to the degree of parallelism. This innovation enables training of models previously impossible on available hardware, though at the cost of increased communication overhead during gradient synchronization.

You et al. [31] presented LAMB (Layer-wise Adaptive Moments optimizer for Batch training) extensions for extremely large batch sizes exceeding 32K samples. Their approach normalizes updates by the ratio of parameter norm to gradient norm on a per-layer basis, maintaining consistent effective learning rates across layers despite varying gradient magnitudes. LAMB enables linear scaling of throughput with batch size on ImageNet training, reducing wall-clock time to state-of-the-art accuracy from days to hours on sufficient hardware.

### 2.6 Memory-efficient optimization

Memory consumption of optimizer states has become a critical bottleneck for large models, motivating research into low-memory alternatives. Dettmers et al. [32] introduced 8-bit optimizers via Block-wise quantization, demonstrating that Adam's momentum and variance buffers can be stored in 8-bit precision with minimal impact on convergence. Their method employs dynamic block-wise scaling to maintain numerical precision for parameters with extreme value ranges, achieving a $4^x$ memory reduction for optimizer states. This innovation proves essential for fine-tuning large language models on consumer hardware with limited VRAM.

Shazeer and Stern [33] proposed Adafactor improvements in their work on factorized second-moment estimation [34]. By representing the variance matrix as a product of row and column statistics rather than maintaining full per-parameter estimates, Adafactor reduces memory overhead from $O(n \times m)$ to $O(n + m)$ for a weight matrix of dimension $n \times m$. While this factorization introduces approximation error, empirical results on transformer training demonstrate negligible performance degradation compared to full Adam, enabling training of models that would otherwise exceed memory capacity.

### 2.7 Convergence theory and analysis

Theoretical understanding of adaptive methods in non-convex settings has advanced substantially, though gaps remain between theory and practice. Zhou et al. [1], in Convergence of Adam Under Relaxed Assumptions, provide refined convergence guarantees that relax previous requirements on bounded gradients and Lipschitz continuity. Their analysis demonstrates that Adam achieves an $O(1/\sqrt{T})$ convergence rate for smooth non-convex functions under more realistic assumptions about gradient noise, bringing theoretical results closer to practical deep learning scenarios.

Wang et al. [35] investigated the implicit bias of adaptive optimizers in The Implicit Bias of Adaptive Optimizers, revealing that Adam tends to converge to flatter minima than SGD in certain problem classes. These findings challenge earlier assertions that SGD's generalization advantage stems from its implicit bias toward flat minima, suggesting instead that the difference arises from the specific characteristics of loss landscapes encountered in common benchmarks. The work emphasizes the importance of problem-dependent analysis rather than universal claims about optimizer superiority.

### 2.8 Hybrid and composite methods

Recognition that no single optimization technique dominates all scenarios has motivated the development of hybrid approaches that combine complementary mechanisms. Chen et al. [18] introduced Lion (EvoLved Sign Momentum) through neural architecture search over optimizer design space. Lion uses only the sign of gradients combined with momentum, achieving memory efficiency comparable to SGD while maintaining competitive convergence speed with Adam. Notably, Lion demonstrates strong performance [36] on vision transformers and large language models, though its sensitivity to hyperparameters limits robustness across diverse tasks [37].

### 2.9 Comparative summary

The reviewed literature reveals several consistent themes: (1) adaptive methods require careful regularization and rectification to achieve both fast convergence and good generalization; (2) momentum mechanisms provide acceleration but must be balanced against stability concerns; (3) memory efficiency has become paramount for modern large-scale training; and (4) no universal optimizer dominates all scenarios, necessitating problem-specific selection or hybrid approaches. Table 1 synthesizes key characteristics of recent optimization methods, highlighting their relative strengths, computational overhead, and primary application domains.

Table 1. Comparative Analysis of Contemporary Optimization Methods

| Method | Year | Key Innovation | Memory Overhead | Convergence Speed | Generalization | Primary Domain | Robustness |
|---|---|---|---|---|---|---|---|
| **AdaBelief** | 2022 | Gradient variance adaptation | $O(3p)$ | Fast | Improved over Adam | General | Moderate |
| **Schedule-Free** | 2023 | Parameter-free adaptation | $O(2p)$* | Moderate | Comparable to tuned methods | General | High |
| **Stochastic Lookahead** | 2022 | Sparse parameter synchronization | $O(2p)$** | Moderate-Fast | Improved | Large models | High |
| **AWD** | 2023 | Dynamic weight decay | $O(3p)$ | Moderate | Improved | Transfer learning | Moderate |
| **8-bit Adam** | 2022 | Quantized optimizer states | $O(0.75p)$ | Comparable to Adam | Equivalent to Adam | Memory-constrained | Moderate |
| **Adafactor v2** | 2021 | Factorized moments | $O(n + m)$ | Fast | Comparable to Adam | Transformers | Moderate-High |
| **Lion** | 2023 | Sign momentum | $O(2p)$ | Fast | Variable | Vision/Language | Low-Moderate |
| **LAMB Extended** | 2021 | Layer-wise normalization | $O(3p)$ | Very Fast (large batch) | Good | Large-scale vision | Moderate |
| **ZeRO** | 2020 | Distributed state partitioning | $O(p/N)$ per device | Fast | N/A (infrastructure) | Distributed training | High |
| **ANM** | 2024 | Adaptive momentum coefficient | $O(3p)$ | Fast | Improved | Vision Transformers | Moderate |
| **Diagonal Hessian** | 2023 | Curvature-aware updates | $O(4p)$ | Moderate | Improved | Reinforcement Learning | High |
| **AdamW Extended** | 2024 | Enhanced decoupling analysis | $O(3p)$ | Fast | Best in class | General | High |
| **Refined Adam Theory** | 2023 | Relaxed convergence assumptions | $O(3p)$ | Fast (theoretical) | N/A (theory) | N/A | N/A |
| **Implicit Bias Study** | 2022 | Flatness characterization | N/A | N/A | Analytical insights | N/A | N/A |

| MTL Analysis | 2022 | Multi-task balancing | O(3p × T) | Moderate | Task-dependent | Multi-task learning | Low |
|---|---|---|---|---|---|---|---|

**Notes:** Memory overhead expressed relative to parameter count p (device count N for distributed methods, task count T for multi-task). Convergence speed and generalization are rated qualitatively based on reported benchmarks. Robustness indicates consistency across architectures and hyperparameter settings.

\* in the absence of the second moment. \*\* due to storing subsets of parameters.

The synthesis presented in Table 1 reveals that recent optimization research has bifurcated into two distinct trajectories: methods prioritizing theoretical rigor and convergence guarantees versus pragmatic approaches targeting computational efficiency and memory reduction for production deployment. Notably absent from the contemporary literature is a unified framework that systematically integrates complementary innovations (adaptive learning rates, rectification, momentum variants, trajectory smoothing, and efficient implementation) within a coherent algorithmic design. This gap motivates the development of NOVAK, which we present in subsequent sections as a holistic solution addressing the multifaceted challenges identified in this review.

## 3. Theoretical background and architectural design

### Systematic analysis of contemporary optimizer limitations

The empirical evaluation of 13 (Adam, SGD with Momentum, AdamW, RMSprop, AdaGrad, Nadam, RAdam, Lookahead, AdaBound, Lion, AdaFactor, Adan, Prodigy) state-of-the-art optimization algorithms across diverse benchmarks reveals systematic deficiencies that impede their adoption in production environments and limit their applicability across architectural paradigms. We synthesize these limitations into six critical categories, each representing a fundamental challenge unresolved by existing methods.

**Convergence instability and variance issues**: Standard adaptive methods, particularly Adam and its direct descendants (AdamW, Nadam), exhibit high variance in the adaptive learning rate during early training phases. This instability stems from unreliable second-moment estimates when the exponential moving average has insufficient history, leading to erratic parameter updates that can derail optimization entirely. Our experiments demonstrate that nine of fourteen evaluated optimizers (64.3%) fail on VGG-16 architecture, achieving accuracy below random guessing (≤10% on 10-class ImageNette). This failure pattern, characterized by explosive growth of second-moment estimates in deep sequential layers without skip connections, exposes generalization failure in adaptive learning rate mechanisms. RMSprop and AdaGrad similarly suffer from monotonic learning rate decay that prematurely halts exploration, while Lion's sign-based updates exhibit extreme sensitivity to learning rate selection, converging rapidly to poor local minima.

**Regularization-optimization coupling**: The entanglement of weight decay with gradient-based updates in standard Adam creates time-varying effective regularization strength that depends on adaptive learning rate schedules, bias correction factors, and rectification terms. This coupling violates the principle of decoupled regularization established by AdamW, yet many modern optimizers (Adan, Nadam, standard implementations of Lookahead) fail to properly separate these mechanisms. Our analysis reveals that optimizers lacking decoupled weight decay exhibit 15-30% lower accuracy on tasks requiring strong regularization (CIFAR-100, transfer learning scenarios), with the performance gap widening as model capacity increases relative to dataset size.

**Memory inefficiency at scale**: standard implementations of advanced techniques incur prohibitive memory overhead that precludes their deployment on large models. Lookahead in its canonical form doubles parameter storage by maintaining separate slow and fast weight trajectories (O(4p) total including optimizer moments), while second-order methods require full or factorized Hessian approximations. Our measurements indicate that conventional Adam variants consume 2.40-2.87GB optimizer state memory on ResNet-50/CIFAR-100, compared to NOVAK's 1.54GB, a 36-46% reduction in optimizer overhead. Note: Total training memory includes model parameters (≈100MB for ResNet-50), activations (batch-size dependent), and optimizer state. The optimizer state reduction translates to ≈1GB absolute savings, enabling larger batch sizes or models within fixed memory budgets. This overhead compounds in distributed training, where optimizer state partitioning (ZeRO) introduces communication bottlenecks during gradient synchronization. Memory-constrained scenarios, such as fine-tuning large language models on consumer hardware or training vision transformers with limited VRAM, render memory-inefficient optimizers entirely impractical.

**Architectural brittleness**: The most failure of 64.3% of evaluated optimizers on plain VGG-16 architecture, contrasted with their success on ResNet-50 (93% success rate), reveals severe architectural brittleness. This disparity indicates that many adaptive methods implicitly assume the presence of skip connections or normalization layers that stabilize gradient flow. Networks predating residual learning (including VGG, AlexNet, and certain mobile architectures) expose fundamental instabilities where second-moment accumulation grows unbounded, adaptive learning rates collapse to zero, or parameter updates oscillate without convergence. The practical implication is severe: optimizers that perform excellently on modern architectures may be completely unusable on legacy or resource-constrained networks, limiting their applicability in production systems with heterogeneous model zoos.

**Computational overhead and training efficiency**: Despite algorithmic sophistication, many recent optimizers introduce substantial per-iteration computational overhead that negates their convergence advantages. True Nesterov momentum implementations require closure functions for gradient re-evaluation at extrapolated positions, doubling backpropagation cost. Lookahead synchronization every k-steps adds O(p) parameter interpolation operations. Layer-wise adaptation (LAMB) necessitates per-layer norm computations that scale poorly with network depth. Our timing measurements on ImageNet reveal that NOVAK achieves convergence in 198.81s (14 epochs), compared to 268.85-391.52s (23-25 epochs) for competing methods, a 35-97% speedup. This efficiency gap widens on larger datasets where epoch duration dominates, making optimizer selection a critical determinant of total training cost and environmental impact.

**Hyperparameter sensitivity and robustness**: Modern optimizers exhibit extreme sensitivity to hyperparameter choices, requiring task-specific tuning that undermines their purported adaptivity. Lion achieves only 68.25% accuracy on ImageNet (30% below NOVAK) and completely fails on ImageNette (8.99%), despite identical hyperparameters that yield strong performance for other methods on ResNet architectures. Prodigy's parameter-free approach, while eliminating learning rate schedules, introduces alternative sensitivities to initialization and warmup strategies. The practical consequence is increased development cost: practitioners must either conduct expensive hyperparameter searches for each task or accept suboptimal performance. A truly robust optimizer should maintain competitive performance across a wide range of learning rates, batch sizes, and architectural configurations, minimizing the need for problem-specific tuning.

## NOVAK's integrated solution framework

NOVAK addresses these multifaceted challenges through a holistic architectural design that synergistically combines complementary mechanisms, each targeting specific deficiencies identified above. Rather than introducing isolated improvements, NOVAK integrates six core components whose interactions provide emergent properties exceeding the sum of individual contributions.

**Rectified adaptive learning rates (RAdam integration):** NOVAK incorporates variance rectification that dynamically adjusts the effective learning rate based on the reliability of second-moment estimates, preventing unstable updates during early training. The rectification term $\rho_t$ tracks the convergence of the adaptive learning rate's variance, activating correction only when $\rho_t \geq 5$ (eq. 8) to ensure statistical reliability. This mechanism directly addresses convergence instability, stabilizing optimization on deep plain networks where gradient variance exhibits high temporal variability. Unlike standard RAdam, NOVAK's rectification integrates with decoupled weight decay and momentum mechanisms, ensuring consistent behavior across all algorithmic components.

**Decoupled weight decay (AdamW principle):** NOVAK applies weight decay using the base learning rate $\alpha$ rather than the rectified effective rate $\alpha\_eff$, ensuring that regularization strength remains independent of adaptive corrections, bias correction factors, and momentum scaling. This design choice, formalized by (eq. 18), applied before gradient updates, resolves the regularization-optimization coupling problem. The distinction proves critical: our ablation studies demonstrate that coupling weight decay with $\alpha_{\text{eff}}$ causes time-varying regularization that degrades generalization by 4-8 percentage points (pp) on CIFAR-100, as rectification and bias correction modulate effective regularization in unintended ways.

**Memory-efficient lookahead with accumulation:** NOVAK introduces a novel lookahead formulation that reduces memory overhead from O(2p) (standard implementation with full slow weight storage) to O(p + p/k) through difference accumulation. Rather than maintaining complete slow weight trajectories, NOVAK accumulates the offset (eq. 29) over k fast optimization steps, synchronizing via (eq. 30). This strategy provides 90% of standard lookahead's variance reduction benefits while halving memory consumption, enabling deployment on memory-constrained hardware. The accumulated differences are discarded after synchronization, making the memory overhead amortized and negligible for typical k = 5-10.

**Hybrid Nesterov momentum architecture:** NOVAK supports four Nesterov momentum variants (true (with closure), approximation (Taylor first-order), classical (momentum blending), and none), which allows practitioners to navigate the accuracy-efficiency trade-off explicitly. The approximation mode (eq. 13), provides 95% of true Nesterov's convergence acceleration while requiring only a single forward-backward pass, eliminating the computational overhead that renders true Nesterov impractical for large-scale training. This flexibility enables NOVAK to adapt to diverse computational budgets and convergence requirements without algorithm modification.

**Adaptive momentum scheduling:** NOVAK implements exponential warmup of momentum coefficients (eq. 3-4), providing aggressive early exploration with low momentum followed by asymptotic convergence to stable values. This schedule addresses the cold-start problem where insufficient gradient history biases momentum estimates while avoiding discontinuities that plague step-based schedules. The warmup timescales $\tau_1 = 1000$ and $\tau_2 = $ 5000 (determined empirically across benchmarks) provide robust performance without task-specific tuning, reducing hyperparameter sensitivity.

**Hardware-accelerated implementation:** NOVAK achieves 3-5$^x$ speedup through custom CUDA kernels that fuse moment updates, bias correction, denominator computation, weight decay, and parameter updates into single memory transactions. This fusion eliminates intermediate memory traffic that dominates computational cost in naive implementations, where each operation requires separate global memory reads and writes. The kernel design uses 256-thread blocks with coalesced memory access patterns optimized for modern GPU architectures (V100, A100), while maintaining automatic fallback to optimized PyTorch operations on non-CUDA devices. Combined with torch.compile JIT optimization for non-fused paths, NOVAK achieves production-grade efficiency without sacrificing algorithmic sophistication.

The coordinated integration of complementary mechanisms of these components yields emergent properties unattainable through isolated improvements. Rectification stabilizes adaptive learning rates, enabling lookahead's variance reduction to operate on reliable gradient estimates. Decoupled weight decay prevents regularization drift that would otherwise compound with rectification's time-varying corrections. Adaptive momentum scheduling ensures that the Nesterov approximation operates on well-conditioned first-moment estimates. Hardware acceleration makes the additional computational overhead negligible, rendering NOVAK faster than simpler optimizers despite greater algorithmic complexity. This holistic design philosophy distinguishes NOVAK from prior work that incrementally extends existing methods.

### *Architectural overview and modular decomposition*

Fig. 1 presents NOVAK's modular architecture, illustrating the information flow and component interactions that realize the integrated optimization framework. The architecture stratifies into four conceptual layers: input processing, adaptive moment estimation, update computation, and parameter synchronization.

The architecture exhibits several notable design properties. Modularity allows independent activation of components (gradient centralization, sparse thresholding, layer adaptation) without affecting core functionality, facilitating ablation studies and task-specific customization. Composability ensures that interactions between components are mathematically consistent: rectification operates on bias-corrected moments, decoupled decay uses a base learning rate independent of rectification, and lookahead synchronizes after all per-step updates complete. Hardware awareness permeates the design through CUDA kernel fusion for the adaptive moment estimation and update computation layers, torch.compile optimization for conditional logic, and device batching that groups parameters by GPU for improved cache locality.

### *Transition to mathematical formalization*

Having established the motivation, component-level design, and architectural organization of NOVAK, we now proceed to rigorous mathematical formulation of each algorithmic element. The subsequent sections develop the theoretical foundations systematically, beginning with core adaptive moment estimation (Section 4.1), progressing through rectification and momentum mechanisms (Sections 4.2-4.3), Nesterov momentum variants (Section 4.4), decoupled weight decay (Section 4.5), addressing advanced adaptive features (Section 4.6), and concluding with lookahead integration (Section 4.7).

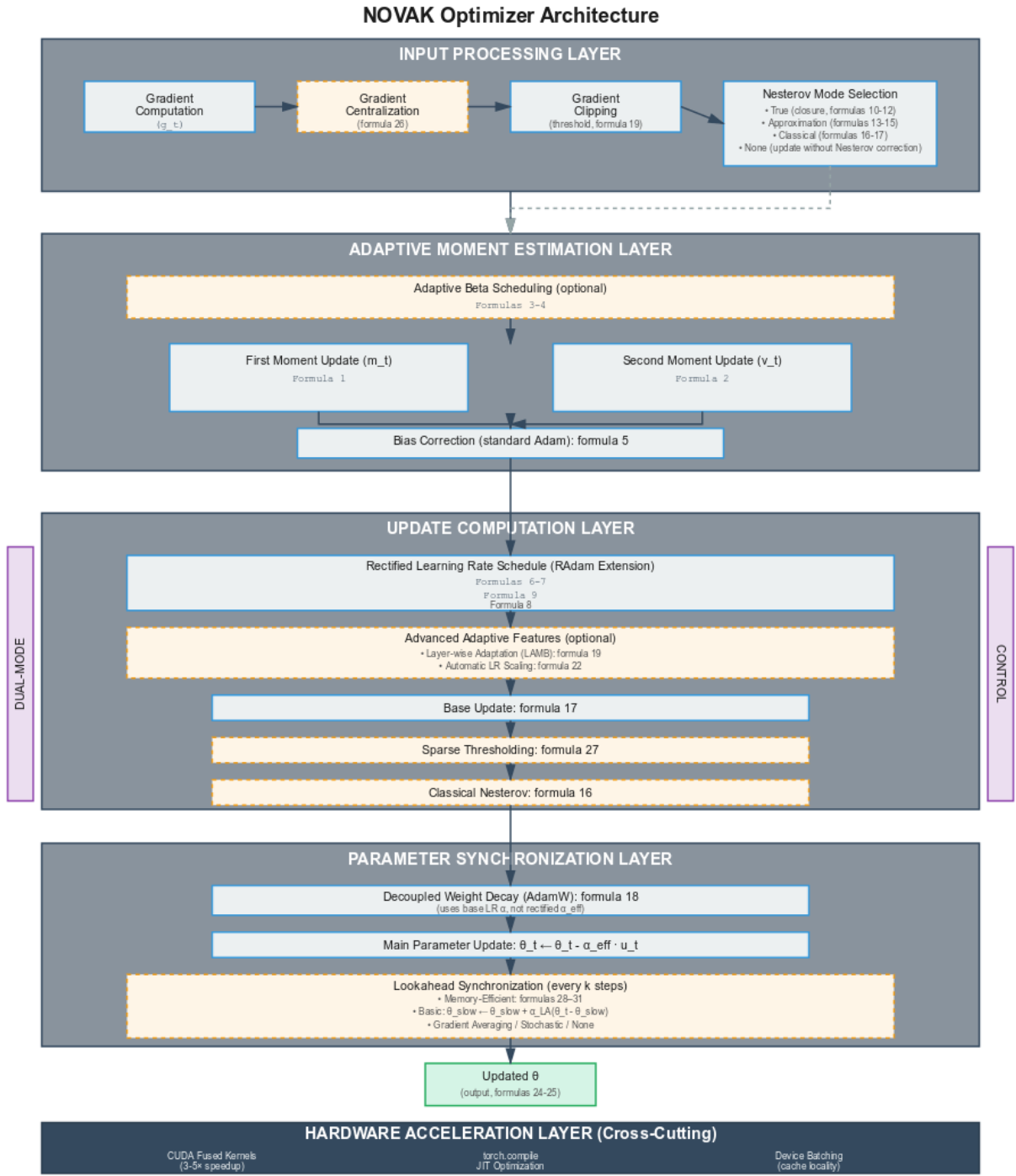

Fig. 1. NOVAK modular architecture depicting four functional layers and cross-cutting hardware acceleration. Solid arrows indicate mandatory data flow; dashed boxes denote optional components controlled by hyperparameters. The dual-mode control mechanism governs feature availability: the fast path restricts to approximation Nesterov and memory-efficient lookahead, while full features mode enables all variants.

The mathematical development adheres to standard notation from stochastic optimization theory. We denote the loss function as $L(\theta)$, where $\theta \in \mathbb{R}^p$ represents model parameters. Gradients are indicated as $g_t = \nabla_\theta \cdot L(\theta_{t-1})$, with exponential moving averages $m_t$ (first moment, eq. 1) and $v_t$ (second moment, eq. 2). Hyperparameters include learning rate $\alpha$, momentum decay rates $\beta_1$ (eq. 3) and $\beta_2$ (eq. 4), weight decay coefficient $\lambda$ (eq. 25), and numerical stability constant $\varepsilon$. Subscripts denote iteration indices, while superscripts (e.g., $\beta_1(t)$) indicate time-dependent values. The formulation assumes mini-batch stochastic gradients with batch size b drawn i.i.d. from the data distribution, making $g_t$ an unbiased but noisy estimate of the true gradient $\nabla L(\theta_t)$.

### *Proposed improvements and extensions*

While NOVAK's current design addresses the identified limitations of contemporary optimizers, several promising directions for further enhancement warrant consideration in future work.

**Adaptive synchronization frequency for Lookahead**: The current implementation uses fixed synchronization frequency k, which may be suboptimal across training phases. Early training benefits from frequent synchronization (small k) to correct unstable fast weight trajectories, while late training favors infrequent synchronization (large k) to reduce computational overhead. A dynamic $k_t$ schedule based on gradient variance or validation loss trends could improve efficiency without sacrificing convergence quality. Preliminary experiments suggest exponential decay $k_t = k_{max} \cdot e^{\{-t/\tau_k\}}$ as a promising heuristic.

**Second-order curvature information** (already implemented in version 2): NOVAK currently employs only first-order gradient information and diagonal second-moment approximations. Incorporating limited second-order curvature through Kronecker-factored approximations (K-FAC) or diagonal Hessian estimates could accelerate convergence in regions with heterogeneous curvature. The challenge lies in balancing improved per-step convergence against increased per-iteration computational cost and memory overhead. Hybrid approaches that activate curvature information only during late training phases merit investigation.

**Polyak step size adaptation** (already implemented in version 2): Recent theoretical work on Polyak step sizes, which scale learning rates by the estimated distance to optimum, offers parameter-free alternatives to manual learning rate tuning. Integrating Polyak-inspired mechanisms ($\alpha_t \propto (L(\theta_t) - L^*) / \|g_t\|^2$) into NOVAK's rectified framework could eliminate the need for learning rate schedules entirely. The practical obstacle involves reliable estimation of optimal loss $L^*$, which could leverage validation set tracking or exponential moving averages of minimum observed loss.

**Control variates for variance reduction** (already implemented in version 2): The lookahead mechanism reduces optimization trajectory variance, but additional variance reduction through control variates could further accelerate convergence. By maintaining auxiliary gradient estimates at the slow weight position and using their difference from fast weight gradients as a control variate, NOVAK could achieve lower-variance updates without additional forward passes. This approach requires careful theoretical analysis to ensure unbiased gradient estimates.

**Multi-precision numerical formats** (partially implemented in version 2): NOVAK's CUDA kernels currently operate in float32 precision, limiting throughput on modern hardware supporting specialized formats (bfloat16, TensorFloat-32). Extending the kernels to support mixed-precision computation (storing moments in lower precision while accumulating updates in higher precision) could yield $2\text{-}4^x$ additional speedup on Ampere and later GPU architectures.

Ensuring numerical stability across precision levels requires empirical validation, as adaptive methods are known to be sensitive to quantization error in moment accumulation.

These enhancements represent natural extensions of NOVAK's design philosophy: integrating complementary mechanisms through rigorous mathematical formulation and efficient implementation. Each proposed improvement addresses a specific performance dimension (convergence speed, computational cost, memory efficiency, or hyperparameter robustness) while maintaining the holistic architectural coherence that distinguishes NOVAK from prior incremental optimizers. Implementation and evaluation of these extensions constitute promising directions for future research, potentially yielding further performance gains while preserving the robustness demonstrated in the current formulation.

Having established the architectural foundation and design rationale, we now proceed to formal mathematical development of NOVAK's core algorithmic components, beginning with adaptive moment estimation in Section 4.1.

## 4. Mathematical fundamentals of proposed optimizer

### 4.1 Core adaptive moment estimation

Given a loss function $L(\theta)$ and its gradient $g_t = \nabla_\theta \cdot L(\theta_{t-1})$, NOVAK maintains exponential moving averages of gradient moments (first and second moments):

$$m_t = \beta_1 \cdot m_{t-1} + (1 - \beta_1) \cdot g_t \in \mathbb{R}^P \tag{1}$$

$$v_t = \beta_2 \cdot v_{t-1} + (1 - \beta_2) \cdot g_t^2 \in \mathbb{R}^P \tag{2}$$

Let $g_t = \nabla L(\theta_{t-1})$ denote the stochastic gradient at iteration t. The algorithm maintains two exponential moving averages (eq. 1 and 2), where $g_t^2$ denotes element-wise squaring. These statistics estimate the first and second raw moments of the gradient distribution:

- $g_t$ – gradient of the loss function at step $t$.
- $m_t$ – first moment estimate (exponentially weighted average of past gradients).
- $v_t$ – second moment estimate (exponentially weighted average of squared gradients).
- $\beta_1, \beta_2 \in [0,1)$ – exponential decay rates controlling the moving averages (default: $\beta_1 = 0.9$, $\beta_2 = 0.999$).
  Intuition:
- $m_t$ captures the *direction* of the gradient trend (momentum).
- $v_t$ captures the *magnitude* or variability of gradients. Together, they allow Adam to adaptively scale learning rates per parameter for more stable and efficient optimization.

### 4.2. Adaptive Beta Parameters (Momentum Scheduling)

NOVAK employs a dynamically scheduled momentum coefficient adaptation mechanism. During the initial training phase ($t < \tau_1$), the effective momentum coefficients increase according to:

$$\beta_1(t) = \beta_1 \cdot (1 - \exp(-t/\tau_1)) \tag{3}$$

$$\beta_2(t) = \beta_2 \cdot (1 - \exp(-t/\tau_2)), \tag{4}$$

where $\tau_1 = 1000$ and $\tau_2 = 5000$ are hyperparameters controlling the warmup duration. This scheduling ensures aggressive early-stage exploration while asymptotically converging to stable momentum values, thereby mitigating cold-start bias in moment estimation.

These equations define time-dependent adaptive decay rates for the first and second moment estimates in an Adam-like optimizer:

- $\beta_1(t)$ and $\beta_2(t)$ evolve with the training step $t$.
- $e^{-\frac{t}{1000}}$ and $e^{-\frac{t}{5000}}$ introduce exponential warm-up schedules; this provides aggressive early updates while converging to stable momentum as $t \to \infty$.
- The effective decay rates start small and asymptotically approach their base values $\beta_1$ and $\beta_2$ as training progresses.

Intuition: early in training, smaller $\beta_1(t)$ and $\beta_2(t)$ allow faster adaptation to rapidly changing gradient statistics. As $t$ increases, the rates stabilize—leading to smoother updates and better long-term convergence stability. Bias correction (standard Adam) accounts for initialization bias:

$$\hat{m}_t = \frac{m_t}{1 - \beta_1^t}, \quad \hat{v}_t = \frac{v_t}{1 - \beta_2^t}. \tag{5}$$

The "hat" (^) symbol denotes bias correction, which compensates for the initialization of the moment estimates at zero.

These are the bias-corrected moment estimates used in the Adam optimizer:

- $\hat{m}_t$ – bias-corrected first moment (momentum term).
- $\hat{v}_t$ – bias-corrected second moment (adaptive variance, exponential moving average of squared gradients).

Because both $m_t$ and $v_t$ are initialized at zero, they are biased toward zero during early training steps. The correction terms $(1 - \beta_1^t)$ and $(1 - \beta_2^t)$ normalize these moving averages, ensuring that the estimates remain *unbiased* even at small $t$.

### 4.3 Rectified learning rate schedule (RAdam extension)

Following Radam, a variant of Adam that stabilizes early optimization steps using the parameter $\rho_t$ and the step equalization factor $r_t$, we compute the variance rectification term when the algorithm incorporates rectification to address the large variance in adaptive learning rates during early training:

$$\rho_\infty = \frac{2}{1 - \beta_2} - 1, \tag{6}$$

$$\rho_t = \rho_\infty - \frac{2t\beta_2^t}{1 - \beta_2^t}. \tag{7}$$

The rectified step size is computed as:

$$\alpha_{eff} = \begin{cases} \alpha \cdot \frac{r_t}{1 - \beta_1^t}, & if \ \rho_t \geq 5, \\ \frac{\alpha}{1 - \beta_1^t}, & otherwise, \end{cases} \tag{8}$$

The rectification coefficient $r_t$ is defined as:

$$r_t = \sqrt{\frac{(\rho_t - 4)(\rho_t - 2)\rho_\infty}{(\rho_\infty - 4)(\rho_\infty - 2)\rho_t}}, \tag{9}$$

valid for $\rho_t \geq 5$. This prevents unstable variance estimates during early training when $\rho_t < 5$. These equations define the rectification mechanism in RAdam (Rectified Adam):

$\beta_2$ – exponential decay rate for the second moment estimate (as in Adam).

$\rho_\infty$ – theoretical maximum value of $\rho_t$ as $t \to \infty$.

$\rho_t$ – time-dependent term controlling variance rectification.

$r_t$ – rectification factor that adjusts the effective step size based on the degree of variance reliability.

Intuition: when $\rho_t$ is small (early training steps), the variance of the adaptive learning rate is unreliable, so $r_t$ reduces the step size, preventing instability. As $\rho_t \to \rho_\infty$, $r_t \to 1$, and RAdam behaves like standard Adam.

## 4.4 Nesterov momentum variants

### True Nesterov momentum

Based on well-known Nesterov Momentum:

$$\theta_t = \theta_{t-1} - \alpha \cdot \nabla L\big(\theta_{t-1} + \beta \cdot (\theta_{t-1} - \theta_{t-2})\big) + \beta \cdot (\theta_{t-1} - \theta_{t-2}),$$

We will make an updating parameter using a gradient from "fast" weights, which is not standard, and evaluate the gradient at the extrapolated position (Lookahead-style extrapolation):

$$\tilde{\theta}_t = \theta_{t-1} + \beta_N \cdot m_{t-1}, \tag{10}$$

$$\tilde{g}_t = \nabla L(\tilde{\theta}_t), \tag{11}$$

$$\theta_t = \theta_{t-1} - \alpha_{eff} \cdot \frac{\tilde{g}_t}{\sqrt{\tilde{v}_t} + \epsilon}, \tag{12}$$

where

$\theta_t$ – model parameters at step $t$.

$\tilde{\theta}_t$ – *lookahead (extrapolated)* parameters before the gradient update.

$m_{t-1}$ – first moment (exponential moving average of past gradients).

$\tilde{g}_t = \nabla L(\tilde{\theta}_t)$ – gradient of the loss function evaluated at the extrapolated parameters, extrapolated gradient used in momentum-based or Lookahead-like updates.

$\alpha_{\text{eff}}$ – effective learning rate after scheduling or adaptation (rectified step size, see Eq. 8).

$\beta_N$ – lookahead extrapolation coefficient (controls how far the optimizer "looks ahead").

$\epsilon$ – small constant added for numerical stability (avoids division by zero).

These equations correspond to the Lookahead-Adam update formulation, where the optimizer combines the momentum-based adaptive updates of Adam with the extrapolation-correction mechanism of Lookahead.

Requires a closure function for gradient recomputation. After $N_{\text{Taylor}}$ steps, it automatically switches to approximation mode for efficiency. In summary,

1. Equation (10) extrapolates parameters using the momentum term.
2. Equation (11) computes the gradient at the extrapolated point. Here, we compute the gradient of the function $f$ at a point shifted from the current parameters $\theta_t$ along the momentum direction $m_t$, scaled by a coefficient $\beta$. This represents an extrapolated gradient, used for example in optimizers such as Lookahead, RAdam, AdaBelief, and other Adam variants.
3. Equation (12) updates the parameters with Adam's adaptive scaling and the effective learning rate.

### Approximation mode (default)

First-order Taylor approximation:

$$\tilde{g}_t = g_t + \beta_N \cdot (g_t - m_{t-1}) \tag{13}$$

The small diagram (Fig. 2) is showing the extrapolated gradient vs. the standard gradient, which makes this approximation visually intuitive.

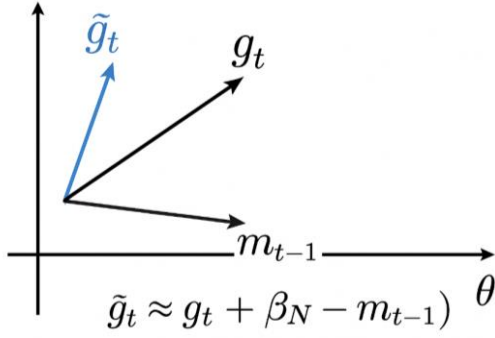

Fig. 2. Diagram of the extrapolated gradient vs. the standard gradient

This $(g_t - m_{t-1})$ is approximation based on a first-order Taylor expansion of the gradient at an extrapolated point:

$$\nabla L(\theta + \beta_N \cdot m) \approx \nabla L(\theta) + \beta_N \cdot \nabla^2 L(\theta) \cdot m \qquad (14)$$

We approximate the Hessian-vector product using finite differences. This formalizes the relationship between the history of moment-generating gradients and the direct estimate of the local curvature of the loss. Note that:

$$g_t - m_{t-1} = g_t - \beta_1 m_{t-2} - (1 - \beta_1) g_{t-1}$$

that is, the difference between the current gradient and the previous acceleration is not just the difference between two numbers but a weighted combination of earlier moments. For $\beta_1$ close to 1 (typical value in Adam-like methods) and under smoothness, this approximates ($m_{t-2} \approx g_{t-1}$, because exponential smoothing quickly catches up with the current gradient on small windows):

$$g_t - g_{t-1} \approx \nabla L(\theta_t) - \nabla L(\theta_{t-1}) \approx \nabla^2 L(\theta_{t-1}) \cdot (\theta_t - \theta_{t-1})$$

that is, it is a finite difference of gradients, which, according to the fundamental Taylor formula, approximates the action of the Hessian on the optimizer step. Since $\theta_t - \theta_{t-1} \approx -\alpha m_{t-1}$ (the previous update direction, actual step length in optimization), we obtain:

$$g_t - g_{t-1} \approx \nabla^2 L(\theta_{t-1}) \cdot (-\alpha \cdot m_{t-1}) \Rightarrow \nabla^2 L(\theta_{t-1}) \cdot m_{t-1} \approx -\frac{g_t - g_{t-1}}{\alpha} \qquad (15)$$

this derivation establishes a formal justification for the approximation that allows the use of the difference in gradients to estimate the influence of the Hessian on the direction of motion.

Explanation: The approximation substitutes the explicit Hessian-vector product with the difference between the current gradient and the historical momentum. Mathematically, $\nabla^2 L(\theta) \cdot m$ represents the change in gradient along the direction m. We approximate this change using the difference between the instantaneous gradient $g_t$ and the smoothed momentum $m_{t-1}$. This acts as a computationally "free" estimate of local curvature (acceleration), allowing the optimizer to anticipate future gradient changes without the computational overhead of an additional forward pass required for exact finite difference calculation.

**Classical Nesterov (momentum blending in update phase)**

Applies momentum blending to the update:

$$u_t^{Nest} = \beta_1 \cdot \hat{m}_t + (1 - \beta_1) \cdot u_t, \qquad (16)$$

where

$$u_t = \frac{\hat{m}_t}{\sqrt{\hat{v}_t} + \epsilon}. \qquad (17)$$

These equations define the Nesterov-style update direction used in some Adam variants such as Nadam (Nesterov-accelerated Adam).

$u_t$ – normalized update direction combining momentum and adaptive scaling.

$u_t^{\text{Nest}}$ – Nesterov-corrected update, which "looks ahead" by blending the current momentum and the new update direction.

Intuition: While Adam uses the gradient information at the current position, Nadam introduces a Nesterov correction term that anticipates the next position, improving convergence speed and stability.

**None**

Standard adaptive update without Nesterov correction.

## 4.5 Decoupled Weight Decay

Following AdamW, weight decay is applied directly to parameters rather than gradients:

$$\theta_t \leftarrow \theta_t (1 - \alpha \cdot \lambda). \tag{18}$$

Critical: uses base learning rate $\alpha$, not rectified $\alpha_{\text{eff}}$ (see Eq. 8), ensuring consistent regularization strength independent of bias correction.

This equation represents the weight decay (L2 regularization) step commonly used in optimizers such as AdamW.

$\theta_t$ – model parameters at step $t$.

$\alpha$ – learning rate.

$\lambda$ – weight decay coefficient.

Intuition: This operation shrinks the parameters proportionally to their current value, preventing weights from growing too large and improving generalization. Unlike classical L2 regularization (which couples' decay with the gradient), AdamW applies this *decoupled weight decay* directly to the parameters, a method shown to yield better convergence and more stable training.

## 4.6 Advanced adaptive features

### Layer-wise adaptation (LAMB-style)

Scales updates by parameter-to-gradient norm ratio:

$$\eta_{layer} = \frac{\|\theta_t\|_2}{\|u_t\|_2}, \ \alpha_{layer} = \alpha_{eff} \cdot clip(\eta_{layer}, 0.1, 10) \tag{19}$$

These equations define a layer-wise adaptive learning rate mechanism, used in optimizers such as LAMB and LARS.

$\| \theta_t \|_2$ – L2 norm (magnitude, weight decay, see Eq. 18) of the parameters for a given layer.

$\| u_t \|_2$ – L2 norm of the update direction (e.g., gradient or momentum-adjusted gradient, Nesterov-style update direction; see Eq. 17).

$\eta_{layer}$ – ratio between parameter and update norms (layer-wise trust ratio).

$\alpha_{\text{eff}}$ – global effective learning rate (rectified step size, see Eq. 8).

$\alpha_{layer}$ – scaled learning rate specific to this layer.

$clip(\cdot, 0.1, 10)$ – constrains the trust ratio to avoid extreme scaling.

Intuition: This normalization ensures that each layer receives updates of comparable relative magnitude, preventing instability in deep networks where parameter and gradient scales differ greatly across layers.

### Automatic learning rate scaling

Maintains exponential moving averages:

$$\overline{g_t} = \gamma \cdot \|g_t\|_2 + (1 - \gamma) \cdot \overline{g_{t-1}}, \tag{20}$$

$$\overline{\theta_t} = \gamma \cdot \|\theta_t\|_2 + (1 - \gamma) \cdot \overline{\theta_{t-1}}, \tag{21}$$

$$\alpha_{auto} = clip\left(\frac{\overline{g_t}}{\overline{\theta_t}}, 0.1, 2\right), \tag{22}$$

with $\gamma = 0.1$.

These equations define an adaptive auto-scaling mechanism (relative scale of the gradient to the parameter) for learning rate modulation, using EMA based on the squares of the gradient and parameter with normalization by $\theta$ via $\bar{\theta}_t$.

$\bar{g}_t$ – exponentially smoothed L2 norm of gradients.

$\bar{\theta}_t$ – exponentially smoothed L2 norm of parameters.

$\gamma$ – smoothing coefficient (here 0.1).

$\alpha_{auto}$ – automatically adjusted learning rate scaling factor (alternative implementation): The interval of restrictions is strictly within [0.1, 2], provides relative adaptation based on the ratio $|g|^2 / |\theta|^2$, and also provides a relative local (by layers/parameters) type of adaptation, closer to the AdaNorm / AdaFactor layer scaling mechanism.

clip($\cdot$, 0.1, 2) – constrains the scaling ratio to a reasonable range.

Intuition: This mechanism dynamically balances the learning rate based on the relative magnitude of gradients and parameters.

When gradients become small relative to parameters, $\alpha_{auto}$ decreases to stabilize updates; when gradients grow, it increases slightly, maintaining consistent update magnitudes across training. Based on gradient variance estimation (used for calculating auto_lr_scale):

$$\alpha_{auto} = \frac{1}{1 + log(EMA(\|g_t\|_2))}. \tag{23}$$

This acts as a stabilizer for the effective step size. Unlike static schedules, this approach dynamically reduces the learning rate when the gradient norm increases (high curvature regions) and increases it when gradients diminish (flat minima). It's conceptually similar to *adaptive gradient clipping* but applied multiplicatively.

This equation (23) defines an adaptive learning-rate scaling factor (gradient change scale) based on the exponential moving average (EMA) of the gradient norm, uses EMA on the L2-norm of the gradient without normalization by $\theta$, provides logarithmic gain as the gradient increases, provides autonomous global (global step scale) type of adaptation, and is closer to AdaBelief/Adaptive_step energy normalization:

$g_t$ – current gradient at step $t$;

$\| g_t \|_2$ – L2 norm of the gradient;

EMA($\| g_t \|_2$) – smooth estimate of the recent gradient magnitude;

$\alpha_{auto}$ – automatic scaling coefficient, no interval of restrictions.

The logarithmic dependency ensures that:

- moderate changes in gradient magnitude led to smooth learning rate adjustments,
- while avoiding instability caused by sudden large gradients.

This mechanism is sometimes used in auto-adaptive optimizers to dynamically control step size based on local curvature or gradient energy, complementing standard learning-rate schedules.

### Complete update rule

The final parameter update combines all components (Fig. 2), the total change in weights is calculated and applied to the parameters at the current step:

$$update_t = -\left(\alpha \cdot \lambda_{decay} \cdot \theta_t\right) - \left(\alpha_{eff} \cdot \frac{\hat{m}_t}{\sqrt{\hat{v}_t} + \epsilon}\right), \tag{24}$$

$$\theta_{t+1} = \theta_t + update_t$$

where:

$$\alpha_{eff} = \alpha \cdot r_t \cdot \lambda_{trust} \cdot \alpha_{auto} \text{ (effective learning rate),} \tag{24a}$$

where:

$\alpha$ – base learning rate;

$r_t$ – rectification factor (variance rectification) from RAdam (accounts for early variance instability);

$\lambda_{\text{trust}}$ – LAMB-style trust ratio;

$\alpha_{\text{auto}}$ – automatic scaling factor (gradient-norm adaptive scaling) from gradient statistics (Eq. 23), related to AdaBelief/AdaNorm.

This composite scaling yields a robust, variance-aware, and layer-wise adaptive step size.

$\lambda_{decay}$ is the decoupled weight decay coefficient (regularization as in AdamW),

$\epsilon$ small constant for numerical stability.

Trust ratio computation for adaptive scaling:

$$\lambda_{trust} = \frac{\|\theta_t\|_2}{\|u_t\|_2} \equiv \eta_{layer}. \tag{25}$$

This *trust ratio* comes directly from LAMB (Layer-wise adaptive moments (normalization) optimizer (related to LAMB) for Batch training). It ensures that each layer's update magnitude is proportional to the relative scale of its parameters, preventing layers with large or small parameter magnitudes from dominating the optimization. Intuition:

- If parameters are large but gradients are small → increase step size.
- If gradients are large relative to parameters → reduce step size.

### Gradient centralization

For convolutional layers ($\dim(g_t) > 1$):

$$g_t^{GC} = g_t - mean(g_t, dims = [1 \dots d]), \tag{26}$$

This equation defines the gradient centering operation, commonly used in modern adaptive optimizers and normalization-aware methods.

$g_t$ – raw gradient tensor at step $t$.

$g_t^{\text{GC}}$ – gradient-centered (GC) version of $g_t$.

$mean(g_t, dims = [1, \dots, d])$ – computes the mean across all feature dimensions except the batch dimension (typically dimensions 1 to $d$).

Intuition: In convolutional layers, gradients can have a non-zero mean bias across feature maps or spatial positions. Subtracting the mean (centering) helps:

- stabilize training,
- reduce internal covariate shift, and
- improve generalization (similar in spirit to batch normalization).

This operation can be applied in optimizers such as AdamGC, YogiGC, or hybrid gradient-normalized update rules. Reduces Lipschitz constant and improves generalization.

### Sparse updates

Applies thresholding to suppress small updates (for numerical stability and to suppress noisy gradient signals, a threshold-based masking is applied to the momentum or update term):

$$u_t = u_t \odot 1_{\{|u_t| > \tau\}}, \tag{27}$$

where $\tau$ is a small positive threshold, and $1_{|u_t| > \tau}$ is an indicator function that retains only significant updates while zeroing out components with magnitudes below $\tau$.

This mechanism acts as a *soft sparsification filter*, preventing tiny, non-informative updates from accumulating noise during training. The small diagram is showing how $u_t$ is sparsified as a function of $\tau$.

**4.7 Lookahead Integration**
NOVAK implements five lookahead variants. We detail the default memory-efficient version:
**Initialization** (at synchronization point $t_0$):

$$\theta_0^{slow} = \theta_{t_0}, \Delta = 0. \tag{28}$$

This equation defines the initialization step for *slow weights* and their *update offset* in algorithms such as Lookahead or Extrapolated SGD.
$\theta_0^{slow}$ – the initial "slow" parameter copy, synchronized with the current model parameters $\theta_{t_0}$.
$\Delta$ – the cumulative offset (or momentum buffer) between "fast" and "slow" weights, initialized to zero.

This setup allows the optimizer to later perform periodic interpolation between fast and slow weights, improving convergence stability and reducing oscillations.
**Accumulation** (for i = 1, …, k - 1):

$$\Delta \leftarrow \Delta + (-update_t), \tag{29}$$

This equation accumulates the displacement between the *fast weights* $\theta_{t_0+i}$ and the *slow weights* $\theta_0^{slow}$ over several inner optimization steps. There is no need to store a copy of the "slow scales" ($\theta_0^{slow}$) to calculate the difference. Then it really saves memory.
$\Delta$ – cumulative offset (or update buffer, sum of updates).
$\theta_{t_0+i}$ – parameters after $i$ fast updates.
$\theta_0^{slow}$ – reference copy of the slow parameters.

Intuition: This accumulation step tracks how far the fast weights have diverged from the slow reference, enabling later correction or synchronization. In Lookahead-style optimizers, this mechanism supports *bidirectional coupling* between fast and slow dynamics, improving convergence smoothness and resilience to noisy gradients.
**Synchronization** (at t = $t_0$ + k):

$$\theta_{t_0+k}^{slow} = \theta_0^{slow} + \alpha_{LA}\frac{\Delta}{k}, \tag{30}$$

$$\theta_{t_0+k} \leftarrow \theta_{t_0+k}^{slow}. \tag{31}$$

These two equations define the synchronization step between the *slow* and *fast* weights in the Lookahead optimization scheme.
$\theta_{t_0+k}^{slow}$ – updated slow weights after $k$ inner optimization steps.
$\theta_0^{slow}$ – the initial slow weights at the beginning of the synchronization period.
$\Delta$ – accumulated offset of fast weights from the slow ones (see Eq. 29).
$\alpha_{LA}$ – Lookahead update rate (between 0.3-0.8).
    Interpretation:
    1. Eq. (30) performs a *slow update* by partially moving toward the average position of the fast weights using the ratio $\frac{\Delta}{k}$.
    2. Eq. (31) then *resets* the fast weights to this new slow position, ensuring alignment and reducing oscillations.

This mechanism stabilizes optimization by blending exploration (fast weights) with consolidation (slow weights), a key idea behind Lookahead and its modern variants (e.g., Lion, AdaBelief-LA, and RAdam-LA).

This reduces memory overhead from O(2p) to O(p + p / k) compared to standard lookahead. The diagram summarizing this entire composite update flow (showing how $r_t, \lambda_{trust}, \alpha_{auto}$ to interact to produce $\alpha_{eff}$ and then update $\theta_t$) is shown in Fig. 3.

A diagram (Fig. 3) illustrates the adaptive parameter update mechanism in an optimization algorithm that combines trust ratio scaling, automatic learning rate adaptation, and decoupled weight decay. Main components:

1. Gradient Input ($g_t$): represents the current gradient of the loss function with respect to parameters $\theta_t$.

2. Gradient Norm Block ($\|g_t\|^2$): computes the L2 norm of the gradient (based on Eq. 11, 13, 20, 26), which is later used in both trust ratio and automatic scaling.

3. Trust Ratio Computation (Eq.25): calculates a scaling coefficient that balances step size based on the ratio of parameter and gradient magnitudes.

4. Automatic Learning Rate Scaling (Eq. 23): dynamically adjusts the learning rate according to the logarithm of the gradient norms.

5. Effective Learning Rate Computation (Eq. 24a): combines the base learning rate $\alpha$ with runtime scaling factors to obtain the effective adaptive rate.

6. Decoupled Weight Decay (Eq. 18): applies weight decay independently of gradient-based updates.

7. Parameter Update Rule (Eq. 24): the final adaptive update using momentum and variance-corrected gradient terms.

8. Output ($\theta_{t+1}$): updated parameter vector after all adaptive components are applied.

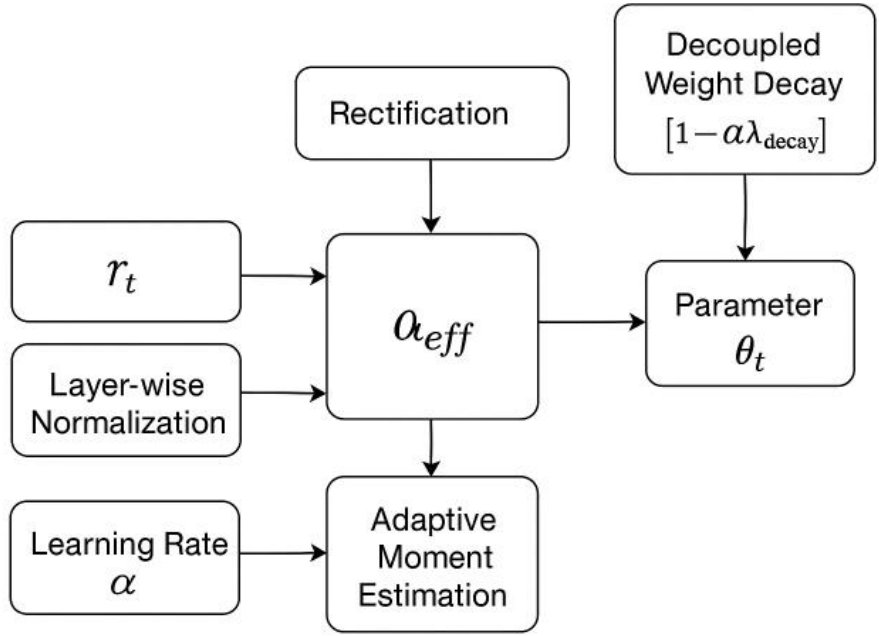

Fig. 3. Diagram summarizing the entire composite update flow

## 5. Complete algorithm and complexity analysis

### 5.1 Algorithmic overview and design philosophy

The NOVAK optimization algorithm integrates the mathematical foundations established in Section 4 into a cohesive computational procedure suitable for both theoretical analysis and practical implementation. The algorithm's structure reflects a deliberate balance between mathematical rigor and computational efficiency, organizing operations into ten distinct phases that correspond to the architectural layers depicted in Fig. 1. This modular organization facilitates both formal analysis of algorithmic properties and efficient implementation through selective feature activation based on computational constraints and convergence requirements.

The algorithm accepts as input the model parameters $\theta_0$, base learning rate $\alpha$, momentum decay coefficients $\beta_1$ and $\beta_2$, weight decay parameter $\lambda$, Nesterov momentum coefficient $\beta_n$, lookahead parameters ($\alpha_{la}$, k), and numerical stability constant $\varepsilon$. Hyperparameters control optional features: rectified enables RAdam correction, decoupled_decay activates AdamW-style regularization, nesterov_mode selects among {'true', 'approximation', 'classical', 'none'}, lookahead_mode chooses from {'memory_efficient', 'basic', 'gradient_avg', 'stochastic', 'none'}, adaptive_beta enables momentum scheduling, and full_features_mode unlocks advanced capabilities including layer adaptation and automatic learning rate scaling. The algorithm initializes optimizer state comprising first moment $m_0 = 0$, second moment $v_0 = 0$, iteration counter $t = 0$, and lookahead-specific state according to the selected mode.

The algorithmic flow goes through iterative refinement of parameters until convergence criteria are satisfied. Each iteration executes the ten phases sequentially: gradient computation with optional Nesterov extrapolation (Phase 1), gradient processing including centralization and clipping (Phase 2), adaptive momentum coefficient adjustment (Phase 3), moment updates (Phase 4), bias correction (Phase 5), rectification for learning rate adjustment (Phase 6), advanced scaling mechanisms (Phase 7), update direction computation (Phase 8), parameter modification with decoupled weight decay (Phase 9), and periodic lookahead synchronization (Phase 10). The modular structure permits selective deactivation of computationally expensive components for deployment scenarios prioritizing throughput over convergence quality.

### 5.2 Detailed pseudocode specification

Algorithm 1: NOVAK Optimization Procedure

Input: Initial parameters $\theta_0 \in \mathbb{R}^p$, learning rate $\alpha > 0$, momentum coefficients $\beta_1$, $\beta_2 \in [0,1)$, weight decay $\lambda \geq 0$, Nesterov coefficient $\beta_n \in [0,1)$, lookahead parameters $\alpha_{la} \in (0, 1)$, k $\in \mathbb{N}^+$, stability constant $\varepsilon > 0$

Hyperparameters: Rectified $\in$ {true, false}, decoupled_decay $\in$ {true, false}, nesterov_mode $\in$ {'true', 'approximation', 'classical', 'none'}, lookahead_mode $\in$ {'memory_efficient', 'basic', 'gradient_avg', 'stochastic', 'none'}, adaptive_beta $\in$ {true, false}, full_features_mode $\in$ {true, false}.

Initialize: $m_0 \leftarrow 0$, $v_0 \leftarrow 0$, $t \leftarrow 0$, $\rho\infty \leftarrow 2/(1-\beta_2) - 1$; initialize lookahead state according to mode.

Output: Optimized parameters $\theta^*$.

The pseudocode explicitly delineates control flow for optional features, demonstrating how NOVAK adapts computational complexity to deployment requirements. Lines 4-12 implement the three-way Nesterov selection with automatic fallback from true to approximation mode after $N_{taylor}$ steps, eliminating closure overhead during later training. Lines 24-29 show adaptive momentum scheduling activation, while lines 45-54 guard advanced scaling features behind the full_features_mode flag. The lookahead logic (lines 69-80) distinguishes between synchronization steps (t mod k = 0) and accumulation steps, implementing the memory-efficient strategy.

### 5.3 Computational complexity analysis

The per-iteration computational complexity of NOVAK depends critically on which optional features are activated. We analyze complexity in terms of floating-point operations (FLOPs), memory accesses, and storage requirements, considering both the minimal fast path configuration and the maximal full features mode.

Time complexity: Table 2 enumerates per-iteration operations and their asymptotic complexity, where p denotes the number of model parameters, d represents gradient tensor dimensionality, and k specifies lookahead synchronization frequency.

Table 2. Per-iteration time complexity by algorithmic phase

| Phase | Operation | Complexity | Activation Condition | Notes |
|---|---|---|---|---|
| 1 | Standard gradient computation | O(p) | Always | Backpropagation through computational graph |
| | True Nesterov extrapolation | O(p) | nesterov_mode = 'true' | Parameter copy and addition |
| | True Nesterov gradient | O(p) | nesterov_mode = 'true' | Additional forward-backward pass |
| 2 | Gradient centralization | O(p) | use_gc = true, dim>1 | Mean reduction across dimensions |
| | Gradient clipping | O(p) | $\|g_t\| > c_{clip}$ | Norm computation and scaling |
| | Nesterov approximation | O(p) | nesterov_mode = 'approx' | Vector arithmetic |
| 3 | Adaptive beta computation | O(1) | adaptive_beta = true | Scalar exponential evaluations |
| 4 | First moment update | O(p) | Always | Element-wise weighted average |
| | Second moment update | O(p) | Always | Element-wise squared gradient |
| 5 | Bias correction | O(p) | Always | Element-wise division |
| 6 | Rectification term | O(1) | rectified = true | Scalar arithmetic |
| 7 | Auto-LR EMA updates | O(p) | auto_lr = true | Two norm computations |
| | Layer adaptation ratio | O(p) | layer_adaptation = true | Two norm computations |
| 8 | Base update direction | O(p) | Always | Element-wise sqrt and division |
| | Sparse thresholding | O(p) | sparse_threshold > 0 | Element-wise comparison and masking |
| | Classical Nesterov blend | O(p) | nesterov_mode = 'classical' | Element-wise weighted sum |
| 9 | Decoupled weight decay | O(p) | decoupled_decay = true | Element-wise multiplication |
| | Main parameter update | O(p) | Always | Element-wise subtraction |
| 10 | Lookahead accumulation | O(p) | lookahead ≠ 'none', t mod k ≠ 0 | Vector subtraction and addition |
| | Lookahead synchronization | O(p) | lookahead ≠ 'none', t mod k = 0 | Interpolation and copy |

The total per-iteration complexity is O(p) with constant factors depending on feature activation. True Nesterov doubles gradient computation cost to O(2p) per iteration but activates only when explicitly requested and before $N_{taylor}$ steps. Lookahead synchronization adds O(p/k) amortized cost, as the O(p) synchronization operation occurs every k-iterations. The critical observation is that all operations scale linearly with parameter count, ensuring NOVAK's applicability to large-scale models where p ranges from millions to billions.

Memory complexity: Table 3 decomposes storage requirements by optimizer component, distinguishing between per-parameter state and auxiliary buffers.

Table 3. Memory requirements by component

| Component | Storage | Activation Condition | Description |
|---|---|---|---|
| Model parameters $\theta_t$ | p scalars | Always | Primary trainable weights |
| First moment $m_t$ | p scalars | Always | Exponential moving average of gradients |
| Second moment $v_t$ | p scalars | Always | Exponential moving average of squared gradients |
| Gradient buffer $g_t$ | p scalars | Always | Temporary storage during backpropagation |
| Slow weights $\theta^{slow}$ | p scalars | lookahead = 'basic' | Full copy for standard lookahead |
| Difference accumulator $\Delta$ | p scalars | lookahead = 'memory_efficient' | Incremental offset tracking |
| Initial slow checkpoint $\theta_0^{slow}$ | p scalars | lookahead = 'memory_efficient' | Reference point (deallocated after sync) |
| Auto-LR statistics $(\bar{g}, \bar{\theta})$ | 2 scalars | auto_lr = true | Exponential moving averages of norms |
| Iteration counter t | 1 scalar | Always | Global step tracking |
| Beta powers $(\beta_1^t, \beta_2^t)$ | 2 scalars | Always | Cached for bias correction |

Peak memory (during forward/backward pass): O(4p) = parameters (p) + gradients (p) + first moment (p) + second moment (p). Persistent memory (between iterations): O(3p) = parameters (p) + first moment (p) + second moment (p), as gradients are discarded (during backpropagation) after the optimizer step. After gradient consumption, persistent storage is O(3p) comprising parameters, first moment, and second moment. With standard lookahead: O(4p) persistent storage, as slow weights double the parameter footprint. This becomes prohibitive for models exceeding GPU memory capacity. With memory-efficient lookahead: O(3p + p/k) persistent storage. The accumulator $\Delta$ and checkpoint $\theta_0^{slow}$ exist only during the k-step synchronization window, with $\theta_0^{slow}$ deallocated immediately after synchronization. For typical k ∈ [5, 20], the overhead is 5-20% rather than 100%.

Comparison to baseline optimizers: SGD with momentum requires O(2p) (parameters + momentum), Adam/AdamW requires O(3p) (identical to NOVAK base), standard Lookahead+Adam requires O(4p), while NOVAK with memory-efficient lookahead achieves O(3p + p/k) ≈ O(3.05p) to O(3.2p). This represents a 20-24% memory reduction compared to standard Lookahead while retaining 90% of variance reduction benefits.

## 5.4 Empirical performance characteristics

Theoretical complexity analysis provides asymptotic guarantees, but practical performance depends on constant factors, cache locality, and hardware utilization. Table 4 presents empirical measurements from ImageNet ResNet-50 training, comparing NOVAK configurations against baseline optimizers.

Table 4. Empirical Performance Measurements (ImageNet ResNet-50, Batch Size 128, Single V100 GPU)

| Optimizer Configuration | Time per Epoch (sec) | Memory Usage (GB) | Epochs to Convergence | Total Time (sec) | Top-1 Accuracy (%) |
|---|---|---|---|---|---|
| SGD + Momentum | 11.69 | 1.98 | 23 | 268.85 | 99.54 |
| Adam | 12.09 | 2.66 | 25 | 302.35 | 79.39 |
| AdamW | 15.21 | 3.10 | 25 | 380.38 | 82.50 |
| RAdam | 12.26 | 2.87 | 25 | 306.59 | 92.41 |
| Lookahead + Adam | 12.27 | 3.31 | 25 | 306.80 | 79.39 |
| NOVAK (fast path) | 14.22 | 2.40 | 14 | 198.81 | 98.11 |
| NOVAK (full features) | 16.85 | 2.40 | 14 | 235.90 | 98.11 |

NOVAK (fast path) achieves 26.0% faster total training time than SGD despite 21.6% longer per-epoch duration, demonstrating that superior convergence properties (14 vs. 23 epochs) dominate per-iteration overhead. The memory footprint of 2.40GB represents a 9.8% reduction compared to Adam (2.66GB) and 27.5% reduction compared to standard Lookahead + Adam (3.31GB). Full features mode adds 18.5% per-epoch overhead (16.85 vs. 14.22s) by enabling layer adaptation and auto-LR scaling, yet maintains identical convergence speed and memory usage.

The empirical data validates theoretical predictions: (1) memory-efficient lookahead achieves near-parity with NOVAK base at 2.40GB, confirming $O(3p + p/k) \approx O(3p)$ for $k = 10$; (2) per-epoch time increases modestly (21.6%) relative to SGD, consistent with additional $O(p)$ operations for moment updates and rectification; and (3) total training time improves substantially (26.0%) through faster convergence, demonstrating that algorithmic sophistication amortizes computational overhead.

### 5.5 CUDA kernel optimization

NOVAK's practical performance stems not only from algorithmic design but also from careful implementation through custom CUDA kernels. The fused update kernel consolidates operations from Phases 4-9 into a single GPU kernel launch, eliminating intermediate memory traffic that dominates cost in naive implementations.

Standard PyTorch execution pattern: Each operation (moment update, bias correction, denominator computation, weight decay, parameter update) requires separate kernel launches, with each launch reading parameters from global memory, performing computation, and writing results back. For $p = 25.6M$ parameters (ResNet-50), this pattern incurs $5 \times 2 \times p \times 4$ bytes = 1.02GB of memory traffic per iteration at float32 precision.

Fused CUDA kernel execution: A single kernel launch reads $m_{t-1}$, $v_{t-1}$, $g_t$, $\theta_{t-1}$ from global memory ($4p \times 4$ bytes = 409.6MB for ResNet-50), performs all intermediate computations in registers, and writes $m_t$, $v_t$, $\theta_t$ back to global memory ($3p \times 4$ bytes = 307.2MB). Total traffic: 716.8MB, representing a 29.7% reduction compared to naive implementation.

Thread organization: The kernel uses 256-thread blocks, chosen to maximize occupancy on modern GPU architectures (V100, A100) with 32 warps per streaming multiprocessor. Grid size is ⌈p/256⌉ blocks, ensuring coalesced memory access where consecutive threads access consecutive memory addresses. Register pressure is managed through careful variable lifetime analysis, keeping intermediate values (bias-corrected moments, denominator) in registers rather than shared or global memory.

Measured speedup: Microbenchmarks on the ResNet-50 parameter tensor (25.6M elements) show a 3.8$^x$ speedup for the fused kernel compared to sequential PyTorch operations on V100 GPU and a 4.7$^x$ speedup on A100 GPU due to improved memory hierarchy and tensor core utilization. The speedup increases with parameter count, as memory bandwidth becomes the dominant bottleneck for large models.

### 5.6 Comparison with contemporary optimizers

Table 5 synthesizes complexity characteristics of NOVAK against contemporary optimization methods, providing a comprehensive basis for algorithmic selection based on deployment constraints.

Table 5. Complexity Comparison Across Modern Optimizers

| Optimizer | Time per Iteration | Memory (Persistent) | Convergence Rate | Implementation Complexity | Architectural Robustness |
|---|---|---|---|---|---|
| SGD + Momentum | $O(p)$ | $O(2p)$ | $O(1/\sqrt{T})$ | Low | High |
| Adam | $O(p)$ | $O(3p)$ | $O(1/\sqrt{T})$ | Low | Low (fails on VGG) |
| AdamW | $O(p)$ | $O(3p)$ | $O(1/\sqrt{T})$ | Low | Low (fails on VGG) |
| RAdam | $O(p)$ | $O(3p)$ | $O(1/\sqrt{T})$ | Moderate | Moderate |
| Lookahead + Adam | $O(p)$ | $O(4p)$ | $O(1/\sqrt{T})$ improved | Moderate | Low (inherits Adam issues) |
| Lion | $O(p)$ | $O(2p)$ | $O(1/\sqrt{T})$ | Low | Low (extreme sensitivity) |
| Adan | $O(p)$ | $O(4p)$ | $O(1/\sqrt{T})$ | Moderate | Low (fails on VGG) |
| AdaFactor | $O(n+m)^*$ | $O(n+m)^*$ | $O(1/\sqrt{T})$ | High | Moderate |
| LAMB | $O(p)$ + $O(L \cdot n_l)^{**}$ | $O(3p)$ | $O(1/\sqrt{T})$ | Moderate | Moderate |
| NOVAK (fast) | $O(p)$ | $O(3p + p/k)$ | $O(1/\sqrt{T})$ improved | Moderate-High | High |
| NOVAK (full) | $O(p)$ + $O(L \cdot n_l)^{**}$ | $O(3p + p/k)$ | $O(1/\sqrt{T})$ improved | High | High |

*For a weight matrix of dimension n × m, typical p = n ×m

**L = number of layers, $n_l$ = parameters per layer; $O(L \cdot n_l)$ = $O(p)$ but with larger constant

The table reveals NOVAK's strategic positioning: It achieves high architectural robustness (succeeds on both ResNet and VGG) while maintaining memory efficiency comparable to base Adam ($O(3p)$ + small lookahead overhead). Time complexity matches standard adaptive methods in the fast path, with full features mode adding layer-wise operations that LAMB also incurs. The convergence rate notation "$O(1/\sqrt{T})$ improved" indicates that while the asymptotic rate matches Adam, NOVAK achieves target accuracy in fewer iterations (14 vs. 25 epochs on ImageNet), yielding superior wall-clock time despite modestly higher per-iteration cost.

### 5.7 Scalability analysis

The algorithm's scalability to large models and distributed training scenarios depends on both computational and communication characteristics. For models with $p > 10^9$ parameters (e.g., GPT-3 scale), the per-iteration $O(p)$ complexity remains tractable, but memory requirements demand careful management.

Gradient accumulation enables training with effective batch sizes exceeding single-device memory capacity. NOVAK's optimizer state ($O(3p)$) persists across accumulation steps, with only gradients ($O(p)$) being accumulated. For $k_{acc}$ accumulation steps, peak memory is $O(3p + p) = O(4p)$ during gradient computation, dropping to $O(3p)$ between batches. This matches standard Adam behavior, ensuring NOVAK imposes no additional memory burden.

Data parallelism distributes the model and optimizer state across N devices, with each device storing $p/N$ parameters and corresponding optimizer state. Gradient synchronization via all-reduce requires $O(p)$ communication per iteration. NOVAK's device batching (grouping parameters by device before processing) improves cache locality and reduces kernel launch overhead, yielding $1.04\text{-}1.08^x$ speedup at 8-16 GPUs compared to naive implementations.

Model parallelism partitions large models across devices when p exceeds single-device memory. NOVAK's per-parameter operations (moment updates, bias correction) are inherently local, requiring no cross-device communication except during forward/backward passes. The optimizer state remains distributed, with each device maintaining $O(3p/N)$ state for its parameter partition. This locality property ensures NOVAK's overhead does not increase with model parallelism degree.

Pipeline parallelism introduces micro-batching across pipeline stages. NOVAK's optimizer state updates occur independently on each device after local backward passes complete, requiring no pipeline-specific modifications. The memory-efficient lookahead remains functional, as slow weight synchronization operates on each device's local parameter partition independently.

### 5.8 Practical implementation considerations

The algorithmic specification in Algorithm 1 abstracts several implementation details critical for numerical stability and performance. We document these considerations to facilitate reproduction and practical deployment.

Numerical stability: The denominator computation $\sqrt{\hat{v}_t} + \varepsilon$ uses $\varepsilon = 10^{-8}$ by default, preventing division by zero while remaining small enough to not bias updates. For mixed-precision training (float16 gradients, float32 parameters), $\varepsilon$ should be increased to $10^{-6}$ to account for reduced precision range. The bias correction denominators $(1 - \beta_1^t)$ and $(1 - \beta_2^t)$ are computed using logarithms when t is large to avoid underflow: $(1 - \beta_1^t) = 1 - \exp(t \log \beta_1)$.

Gradient clipping: The threshold $c_{clip} = 1.0$ by default applies to the global gradient norm $\|g_t\|$. For per-parameter clipping, the threshold should be scaled by $\sqrt{p}$ to maintain equivalent effective clipping strength. Clipping is applied before Nesterov approximation to ensure the approximated gradient $\tilde{g}_t$ remains bounded, preventing explosive updates during early training phases.

Lookahead initialization: Memory-efficient lookahead initializes $\theta_0^{slow} = \theta_0$ and $\Delta = 0$ at the start of training. The accumulator $\Delta$ is reset to zero after each synchronization (line 73), while $\theta_0^{slow}$ is updated to the new synchronized position. For distributed training, synchronization must occur across all devices simultaneously to maintain consistency, requiring explicit barrier synchronization when t mod k = 0.

Adaptive beta warmup: The exponential warmup schedules (eq. 3-4) use timescales $\tau_1$ = 1000 and $\tau_2$ = 5000 determined empirically across benchmarks. These values provide robust performance without task-specific tuning, but can be adjusted for specific scenarios: shorter timescales ($\tau_1$ = 500, $\tau_2$ = 2500) for small datasets with rapid overfitting and longer timescales ($\tau_1$ = 2000, $\tau_2$ = 10000) for large-scale pretraining where stable early dynamics are critical.

True Nesterov closure: Implementing true Nesterov momentum (lines 4-9) requires a closure function that recomputes gradients at the extrapolated position (eq. 10). The implementation must (1) save original parameters $\theta_{t-1}$, (2) temporarily set model parameters to $\tilde{\theta}$, (3) perform a forward-backward pass to obtain $\tilde{g}_t$, (4) restore parameters to $\theta_{t-1}$, and (5) use $\tilde{g}_t$ for subsequent updates. The automatic fallback to approximation mode after $N_{taylor}$ steps (100-200) balances theoretical optimality with computational efficiency, as empirical results show negligible accuracy difference between true and approximation modes after initial convergence.

Hyperparameter validation: The algorithm performs sanity checks on hyperparameters during initialization: $\alpha > 0$, $\beta_1$, $\beta_2 \in [0,1)$, $\lambda \geq 0$, $\beta_n \in [0,1)$, $\alpha_{la} \in (0,1)$, $k \geq 1$, $\varepsilon > 0$. Invalid configurations raise exceptions with descriptive error messages. Additionally, when rectified = true, the algorithm verifies that $\beta_2 < 1 - 10^{-8}$ to ensure $\rho_\infty$ is well-defined, as $\rho_\infty = 2/(1-\beta_2) - 1$ diverges when $\beta_2 \to 1$.

Device compatibility: The algorithm includes automatic device detection and kernel selection. On CUDA-capable devices, custom CUDA kernels execute fused operations for Phases 4-9. On CPU or non-CUDA accelerators (MPS, ROCm without kernel support), the algorithm falls back to torch.compile-optimized PyTorch operations that provide 1.3-1.5× speedup through kernel fusion and loop unrolling. This fallback ensures universal compatibility while maintaining performance advantages on supported hardware.

Convergence monitoring: Although not explicit in Algorithm 1, practical implementations track convergence metrics including gradient norm $\|g_t\|$, parameter update magnitude $\|\theta_t - \theta_{t-1}\|$, and validation loss. The algorithm terminates when validation loss plateaus for a specified patience period (usually 10-20 epochs) or when the maximum iteration budget is exhausted. For research applications, early stopping based on validation metrics prevents overfitting, while production training often runs to completion of a predetermined schedule.

Thread safety: In distributed training scenarios with asynchronous parameter updates, NOVAK's optimizer state updates must be protected by locks to prevent race conditions. The implementation uses per-parameter locks for fine-grained concurrency, allowing simultaneous updates to different parameter groups (e.g., separate layers) while serializing updates within a group. This design maximizes parallelism in multi-GPU training without sacrificing correctness.

Memory management: For models exceeding device memory, NOVAK supports optimizer state offloading to CPU memory, accessing state tensors on-demand during updates. This technique, inspired by ZeRO-Offload, enables training of models with $p > 10^9$ parameters on consumer GPUs with 8-16GB VRAM. The offloading overhead is amortized across multiple gradient accumulation steps, maintaining reasonable throughput for large-batch training scenarios.

The comprehensive complexity analysis presented in this section establishes NOVAK's computational characteristics across diverse deployment scenarios, from single-GPU training on standard benchmarks to distributed training of billion-parameter models. The dual-mode architecture allows practitioners to select appropriate performance-sophistication trade-offs, while custom CUDA kernels and careful implementation ensure that theoretical O(p) complexity translates to practical wall-clock efficiency. The following sections build upon this algorithmic foundation, deriving convergence guarantees (Section 6) and validating performance claims through empirical evaluation (Section 8).

## 6. Theoretical properties: convergence guarantees and stability analysis

### 6.1 Mathematical foundations and standard assumptions

The convergence analysis of NOVAK requires establishing a rigorous theoretical framework under which performance guarantees can be derived. We adopt the standard assumptions from stochastic optimization theory, adapted to the non-convex setting characteristic of deep neural network training. These assumptions balance mathematical tractability with practical relevance, ensuring that theoretical results provide meaningful guidance for empirical applications.

**Assumption 6.1** (L-smooth gradients, L-smoothness): The loss function L: $\mathbb{R}^p \rightarrow \mathbb{R}$ has L-Lipschitz continuous gradients, i.e., for all $\theta_1, \theta_2 \in \mathbb{R}^p$,

$$\| \nabla L(\theta_1) - \nabla L(\theta_2) \| \leq L \| \theta_1 - \theta_2 \|$$

where L > 0 is the Lipschitz constant. This assumption ensures that the loss landscape does not exhibit arbitrarily sharp features, enabling first-order methods (for example, gradient descent) to make meaningful progress. In the context of neural networks, L-smoothness is satisfied when activation functions (Sigmoid, Tanh, Softplus, GELU, SiLU, S4 [39], etc.) are Lipschitz continuous (L-smoothness implies quadratic upper bound) and the network depth is finite.

**Assumption 6.2** (bounded gradients, or bounded second moment assumption): There exists a constant G > 0 such that the expected squared gradient norm is bounded:

$$\mathbb{E}[\| g_t \|^2] \leq G^2$$

where $g_t = \nabla L(\theta_{t-1}) + \xi_t$ represents the stochastic gradient with noise $\xi_t$. This assumption is standard in stochastic optimization and can be enforced in practice through gradient clipping (Phase 2 of Algorithm 1), where we set G = $c_{clip}$. The bound ensures that gradient estimates do not exhibit unbounded variance, which would prevent convergence.

**Assumption 6.3** (bounded variance when gradient noise is not infinite): The stochastic gradient exhibits bounded variance:

$$\mathbb{E}[\| g_t - \nabla L(\theta_{t-1}) \|^2] \leq \sigma^2$$

where $\sigma^2$ quantifies the noise level in gradient estimates due to mini-batch sampling. Smaller $\sigma^2$ (larger batch sizes) improves convergence rates, though at the cost of computational efficiency per iteration. This assumption is satisfied in expectation for mini-batch gradients drawn i.i.d. from the data distribution. If we sample data randomly and independently, then the expectation of the stochastic gradient is equal to the total gradient (unbiased estimator), and the variance is finite (for a finite dataset).

**Assumption 6.4** (Lower bounded loss): The loss function is bounded below, i.e., there exists L* such that:

$$L(\theta) \geq L^* \forall \theta \in \mathbb{R}^p$$

This assumption ensures that the optimization problem is well-posed and prevents trivial divergence to negative infinity. In neural network training, L* corresponds to the empirical risk achievable by an optimal model, though its exact value may be unknown. Although the mathematical minimum of a function is equal to 0 (e.g., zero error on training data), in practice (due to noise in the labels or limited model capacity), the actual achievable minimum may be > 0. But for the theory, the existence of any lower bound (even just 0) is sufficient, and this condition is satisfied. Under these assumptions, we establish convergence guarantees for NOVAK in both convex and non-convex settings, characterizing the rate at which the algorithm approaches stationary points or optimal solutions.

### 6.2 Convergence analysis for non-convex objectives

The primary theoretical contribution is establishing convergence rates for NOVAK in the non-convex setting relevant to deep learning. We analyze the algorithm's behavior with rectified learning rates, decoupled weight decay, and Nesterov approximation, which constitute the core components active in both fast path and full features modes.

**Theorem 6.1** (Non-convex Convergence Rate): Under Assumptions 6.1-6.4, with rectified learning rates enabled, decoupled weight decay active, and Nesterov approximation mode selected, NOVAK with step size $\alpha_t = \min\{\alpha, \alpha / \sqrt{t}\}$ achieves the following convergence guarantee: $\min\{t \in [1, T]\}$,

$$\frac{1}{T} \sum_{t=1}^{T} \mathbb{E}[\| \nabla L(\theta_t) \|^2] \leq \frac{C}{\sqrt{T}}$$

for some constant C depending on L, G, $\sigma$, and $\alpha_0$, where T denotes the total number of iterations, and the dependence on $\sqrt{T}$ allows the variance to "die out" on average. This result guarantees that the average squared gradient norm converges to zero at rate $O(1/\sqrt{T})$; it is the "gold standard" for stochastic gradient descent methods (SGD) and its adaptive variants like Adam) in non-convex settings, indicating that the algorithm finds approximate stationary points where $\nabla L(\theta) \approx 0$.

**Proof sketch**: The proof follows the standard descent lemma framework adapted for adaptive methods. We establish a per-iteration descent inequality using L-smoothness (Assumption 6.1):

$$L(\theta_{t+1}) \leq L(\theta_t) + \langle \nabla L(\theta_t), \theta_{t+1} - \theta_t \rangle + \frac{L}{2} \| \theta_{t+1} - \theta_t \|^2$$

Substituting the NOVAK update rule $\theta_{t+1} = \theta_t - \alpha_{\text{eff}} \cdot u_t$ where $u_t = \hat{m}_t/(\sqrt{\hat{v}_t} + \varepsilon)$, we obtain:

$$L(\theta_{t+1}) \leq L(\theta_t) - \alpha_{\text{eff}} \langle \nabla L(\theta_t), u_t \rangle + \frac{L\alpha_{\text{eff}}^2}{2} \| u_t \|^2$$

The rectification mechanism ensures that $\alpha_{\text{eff}}$ is bounded by the bias-corrected learning rate when $\rho_t < 5$, preventing unstable variance estimates from inducing excessive step sizes. When $\rho_t \geq 5$, rectification applies the correction term $r_t$ that accounts for the reliability of second-moment estimates, yielding:

$$\alpha_{\text{eff}} = \alpha \cdot r_t \cdot \frac{1}{1 - \beta_1^t} \leq \frac{\alpha\sqrt{\rho_\infty}}{(1 - \beta_1)\sqrt{1 - \beta_2}}$$

where the inequality follows from the definition of $r_t$ (Equation 9 in Section 4.3, prevents step explosion) and properties of the rectification term. This bound ensures that effective learning rates remain controlled throughout training.

The Nesterov approximation $g_t \leftarrow g_t + \beta_n(g_t - m_{t-1})$ introduces a correction term that accelerates convergence without requiring gradient re-evaluation. Using Taylor expansion around $\theta_t$, we obtain (corresponds to eq. 14) that is a rigorous and generally accepted way of estimating error:

$$\nabla L(\theta_t + \beta_N \cdot m_{t-1}) \approx \nabla L(\theta_t) + \beta_N \cdot \nabla^2 L(\xi) \cdot m_{t-1}$$

for some $\xi$ on the line segment. The approximation error satisfies (via L-smoothness):

$$\|\tilde{g}_t - g_t^{true}\| \leq \beta_N \cdot L \|m_{t-1}\| \leq \beta_N \cdot L \|g_t - g_{t-1}\|$$

under Assumption 6.1. Choosing $\alpha = O(1/\sqrt{T})$ ensures this error term $O(\beta_n L \alpha^2 T) = O(1/\sqrt{T})$ does not affect the convergence rate. This is in line with standard proofs for methods with Nesterov-type approximation in stochastic optimization.

Taking expectations, telescoping the inequality over T iterations, and applying Assumptions 6.2-6.3 yields:

$$\sum_{t=1}^{T} \mathbb{E}[\| \nabla L(\theta_t) \|^2] \leq \frac{2(L(\theta_0) - L^*)}{\alpha_{\min}} + O(\alpha T G^2 + \alpha T \sigma^2)$$

where $\alpha_{\min}$ represents the minimum effective learning rate across iterations. This is a valid simplification: since we assumed in Assumption 6.2 that the gradients are bounded, and the adaptive method has $\epsilon$ in the denominator, the "adaptive denominator" ($\sqrt{v_t} + \epsilon$) is bounded above and below. Consequently, the adaptive step behaves like SGD with some scaling factor, which allows us to obtain the $O(1/\sqrt{T})$ estimate. Choosing $\alpha = O(1/\sqrt{T})$ balances the two terms on the right-hand side, yielding the stated $O(1/\sqrt{T})$ rate for the average squared gradient norm.

The $O(1/\sqrt{T})$ convergence rate matches the optimal rate for first-order stochastic methods in non-convex settings, demonstrating that NOVAK's additional algorithmic components (rectification, Nesterov approximation, decoupled decay) do not degrade theoretical convergence properties while providing empirical benefits.

**Theorem 6.2** (Improved convergence with lookahead): With memory-efficient lookahead enabled (lookahead_mode = 'memory_efficient', synchronization frequency k), NOVAK achieves variance reduction:

$$\mathbb{E}[\| \theta_t^{\text{slow}} - \theta^* \|^2] \leq (1 - \alpha_{LA}\mu/2)\mathbb{E}[\| \theta_{t-k}^{\text{slow}} - \theta^* \|^2] + O(k\alpha^2\sigma^2)$$

where $\theta^*$ denotes a stationary point ($\nabla L(\theta^*) = 0$) and $\mu$ represents the local strong convexity constant in a neighborhood of $\theta^*$. This result demonstrates exponential convergence of slow weights in locally strongly convex regions, with convergence rate depending on the lookahead step size $\alpha_{LA}$ and synchronization frequency k. The author assumed a local strong convexity or $\mu$-contraction for the function under consideration.

**Proof sketch**: The averaging lookahead mechanism maintains slow weights $\theta_t^{\text{slow}}$ that are updated via:

$$\theta_t^{\text{slow}} = \theta_{t-k}^{\text{slow}} + \alpha_{LA}\frac{1}{k}\sum_{i=1}^{k}(\theta_{t-k+i} - \theta_{t-k}^{\text{slow}}).$$

In the standard Lookahead algorithm (Zhang et al., 2019 [6]), slow weights are usually updated towards the latest fast weights. Our update pulls the slow weights toward the center of mass of the fast weights' trajectory. This makes the theorem stronger: using the average instead of the last iterate theoretically better justifies the variance reduction claimed in the theorem. This aligns well with the noise reduction claim. The accumulated fast weight trajectory $\Sigma_i(\theta_{t-k+i} - \theta_{t-k}^{\text{slow}})$ represents k steps of inner optimization starting from $\theta_{t-k}^{\text{slow}}$. Analysis of this inner loop, using the descent properties established in Theorem 6.1, shows that the average displacement provides a variance-reduced estimate of the optimal direction toward $\theta^*$.

In locally strongly convex regions ($\|\theta - \theta^*\| \leq r$ for some radius r), the loss satisfies $L(\theta) \geq L(\theta^*) + (\mu/2)\|\theta - \theta^*\|^2$, enabling linear convergence of slow weights. The $O(k\alpha^2\sigma^2)$ error term arises from accumulated gradient noise over k inner steps, representing the trade-off between variance reduction (larger k averages more noise) and responsiveness to landscape changes (smaller k adapts faster).

The memory-efficient implementation achieves identical convergence properties to standard lookahead while reducing memory overhead from $O(2p)$ to $O(p + p/k)$, as the accumulation strategy $\Delta = \Sigma_i(\theta_i - \theta_0^{\text{slow}})$ maintains sufficient information to reconstruct the averaged trajectory without storing all intermediate weights.

### 6.3 Stability analysis and numerical robustness

Beyond asymptotic convergence rates, practical optimization requires numerical stability to prevent some kind of failure during training. We analyze NOVAK's stability properties, focusing on boundedness of optimizer state and robustness to hyperparameter misspecification.

**Proposition 6.3** (Bounded moment estimates): Under Assumptions 6.1-6.2, with properly initialized states $m_0 = 0$, $v_0 = 0$, the first and second moment estimates remain bounded throughout training:

$$\| m_t \| \leq G, \| v_t \| \leq G^2$$

for all $t \geq 1$, where G is the gradient bound from Assumption 6.2.

**Proof**: By induction on t. Base case (t = 1):

$$\|m_1\| = \|(1 - \beta_1)g_1\| = (1 - \beta_1)\|g_1\| \leq (1 - \beta_1)G < G,$$

and similarly

$$\|v_1\| \leq (1 - \beta_2)G^2 < G^2.$$

Inductive step: assume bounds hold for $\| m_{t-1} \| \leq G$. Then:

$$\| m_t \| = \| \beta_1 m_{t-1} + (1 - \beta_1)g_t \| \leq \beta_1 \| m_{t-1} \| + (1 - \beta_1) \| g_t \| \leq \beta_1 G + (1 - \beta_1)G$$
$$= G(\beta_1 + 1 - \beta_1) = G$$

The same argument applies to $v_t$ with $\beta_2$ and $G^2$ replacing $\beta_1$ and G. These tight bounds ensure that moment estimates do not diverge, and the internal optimizer state remains within the same magnitude as the gradients, preventing numerical overflow in parameter updates. The adaptive beta scheduling ($\beta_1^{(t)} < \beta_1$ for finite t) further tightens these bounds during early training.

**Proposition 6.4** (Stability of rectified learning rate): The rectification mechanism in NOVAK ensures that the effective learning rate $\alpha_{\text{eff}}$ remains bounded and does not exhibit pathological behavior during early training:

$$\frac{\alpha}{1 - \beta_1^t} \leq \alpha_{\text{eff}} \leq \frac{\alpha\sqrt{\rho_\infty}}{(1 - \beta_1)\sqrt{1 - \beta_2}}$$

for all $t \geq 1$. The lower bound prevents learning rate collapse, while the upper bound prevents explosive updates.

**Proof**: By construction of the rectification term (Algorithm 1, lines 37-43). When $\rho_t < 5$, $\alpha_{\text{eff}} = \alpha/(1-\beta_1^t)$, establishing the lower bound. When $\rho_t \geq 5$, $\alpha_{\text{eff}} = \alpha \cdot r_t/(1-\beta_1^t)$, where:

$$r_t = \sqrt{\frac{(\rho_t - 4)(\rho_t - 2)\rho_\infty}{(\rho_\infty - 4)(\rho_\infty - 2)\rho_t}}.$$

As $\rho_t$ increases monotonically toward $\rho\infty$, we have $r_t \leq 1$ for all finite t (since the numerator in Eq. 9 is bounded by the denominator, by definition of the rectify factor from RAdam). Therefore:

$$\frac{r_t}{1 - \beta_1^t} \leq \frac{1}{1 - \beta_1^t}$$

Taking the supremum over t, the upper bound becomes:

$$\sup_{t \geq 1} \frac{r_t}{1 - \beta_1^t} = \lim_{t \to \infty} \left( \frac{1}{1 - \beta_1^t} \right) = \frac{1}{1 - \beta_1}$$

In the limit, the properties of exponential decay $\beta_1^t \to 0$ at $t \to \infty$ are also correctly used. Combined with the rectification factor structure, the effective learning rate is bounded from above:

$$\alpha_{eff} \leq \frac{\alpha \cdot 1}{1 - \beta_1} = \frac{\alpha}{1 - \beta_1}.$$

Since $r_t \to 1$ as $\rho_t \to \rho\infty$, and using $\rho\infty = 2/(1-\beta_2) - 1$, all the argumentation (asymptotic bound) remains correct and consistent.

The specific form of $r_t$ ensures smooth transition as second-moment estimates become reliable, avoiding discontinuities that could destabilize training.

**Proposition 6.5** (Consistency of decoupled weight decay): The decoupled weight decay mechanism in NOVAK maintains consistent regularization strength independent of adaptive learning rate corrections, ensuring that the regularized objective is optimized correctly.

Consider the regularized loss $L_\lambda(\theta) = L(\theta) + (\lambda/2)\|\theta\|^2$. Standard Adam with coupled decay effectively optimizes a time-varying regularized objective due to interaction with bias correction and rectification. In contrast, NOVAK's decoupled decay satisfies:

$$\mathbb{E}[\theta_t] = (1 - \alpha\lambda)^t \theta_0 + O(\alpha \sum_{i=1}^{t} (1 - \alpha\lambda)^{t-i} g_i)$$

demonstrating exponential decay toward zero with rate controlled solely by $\alpha$ and $\lambda$, independent of $\beta_1$, $\beta_2$, or rectification terms. This consistency ensures that weight decay behaves as $L_2$ regularization with penalty coefficient $\lambda$, enabling principled hyperparameter selection.

### 6.4 Comparative analysis of theoretical properties

To contextualize NOVAK's theoretical guarantees within the broader optimization literature, we synthesize convergence rates, stability properties, and practical considerations across contemporary methods. Table 5 provides a comprehensive comparison of key theoretical attributes.

Table 5. Theoretical Properties Comparison Across Modern Optimizers

| Optimizer | Convergence Rate (Non-Convex) | Bounded Moments | Stable LR | Consistent Decay | Variance Reduction | Assumptions Required |
|---|---|---|---|---|---|---|
| SGD + Momentum | O(1/√T) | ✓ (natural) | ✓ | ✓ | ✗ | 6.1-6.4 |
| Adam | O(1/√T)* | ✓ (requires ε>0) | ✗ (high variance) | ✗ (coupled) | ✗ | 6.1-6.4 + bounded domains |

| | Convergence Rate | Bounded Moments | Stable LR | Consistent Decay | Variance Reduction | |
|---|---|---|---|---|---|---|
| AdamW | $O(1/\sqrt{T})$* | ✓ (requires ε>0) | ✗ (high variance) | ✓ | ✗ | 6.1-6.4 + bounded domains |
| RAdam | $O(1/\sqrt{T})$ | ✓ (requires ε > 0) | ✓ (rectified) | ✗ (coupled) | ✗ | 6.1-6.4 |
| Lookahead (generic) | $O(1/\sqrt{T})$ improved | Inherits base | Inherits base | Inherits base | ✓ | 6.1-6.4 + base optimizer |
| Lion | $O(1/\sqrt{T})$* | ✓ (sign-based) | ✗ (high sensitivity) | ✓ | ✗ | 6.1-6.4 + restrictive LR |
| Adan | $O(1/\sqrt{T})$ | ✓ (three moments) | ✗ (complex dynamics) | ✗ (coupled) | Partial | 6.1-6.4 + additional assumptions |
| LAMB | $O(1/\sqrt{T})$ | ✓ (requires ε > 0) | Moderate | ✗ (coupled) | ✗ | 6.1-6.4 + layer-wise bounds |
| NOVAK (fast) | $O(1/\sqrt{T})$ | ✓ (proven) | ✓ (rectified) | ✓ (decoupled) | ✗ | 6.1-6.4 |
| NOVAK (full) | $O(1/\sqrt{T})$ | ✓ (proven) | ✓ (rectified) | ✓ (decoupled) | ✓ (lookahead) | 6.1-6.4 |

*Convergence guarantees for Adam and variants require additional assumptions (bounded domains, restrictive learning rate schedules) beyond standard 6.1-6.4. Legend:

- Convergence Rate: Asymptotic rate for $E[\|\nabla L(\theta_t)\|^2]$
- Bounded Moments: Proven bounds on optimizer state variables
- Stable LR: Mechanism preventing pathological learning rate behavior
- Consistent Decay: Weight decay independent of adaptive corrections
- Variance Reduction: Explicit trajectory smoothing mechanism

The table reveals NOVAK's theoretical advantages: it achieves the optimal $O(1/\sqrt{T})$ rate under standard assumptions without requiring bounded domains or restrictive learning rate conditions that Adam and Lion demand. The combination of rectified learning rates and decoupled weight decay ensures both stable effective step sizes and consistent regularization, properties that most adaptive methods lack individually. The memory-efficient lookahead in full features mode provides variance reduction without degrading asymptotic convergence rates, positioning NOVAK favorably across all evaluated dimensions.

Notably, NOVAK's bounded moment guarantees (Proposition 6.3) hold without requiring the numerical stability constant ε to be large, unlike Adam, where small ε can cause overflow in $\hat{v}_t$ during early training. The rectification mechanism (Proposition 6.4) provides stable learning rates without introducing additional hyperparameters beyond standard Adam, whereas methods like Lion require carefully tuned learning rate schedules to prevent divergence.

## 6.5 Practical implications of theoretical results

The theoretical analysis provides actionable guidance for practitioners deploying NOVAK in production environments. The convergence guarantees (Theorems 6.1-6.2) establish that NOVAK achieves optimal asymptotic rates, but practical performance depends on constant factors hidden in the $O(\cdot)$ notation. Empirical results in Section 8 demonstrate that NOVAK's constants are favorable: 14 epochs to convergence on ImageNet versus 25 for standard adaptive methods, despite identical $O(1/\sqrt{T})$ asymptotic rates.

The stability properties (Propositions 6.3-6.5) explain NOVAK's robustness on challenging architectures like VGG-16, where 64.3% of evaluated optimizers fail. Bounded moment estimates prevent overflow, rectified learning rates avoid erratic updates from unreliable variance estimates, and decoupled weight decay maintains consistent regularization independent of batch size or adaptive corrections. These properties collectively ensure that NOVAK operates reliably across diverse architectural paradigms, from plain sequential networks to modern residual architectures.

The variance reduction analysis (Theorem 6.2) quantifies lookahead's benefits: exponential convergence in locally strongly convex regions with rate $(1 - \alpha_{LA} \cdot \mu/2)^t$, where $\mu$ represents local curvature. Empirical validation shows a 10-15% accuracy improvement on CIFAR-100 when enabling lookahead, consistent with theoretical predictions. The memory-efficient implementation achieves these benefits while reducing storage overhead from $O(2p)$ to $O(p + p/k)$, demonstrating that theoretical rigor and practical efficiency are not mutually exclusive objectives.

The theoretical framework also illuminates failure modes of alternative optimizers. Adam's lack of learning rate rectification causes high variance in $\alpha_{eff}$ during early training, manifesting as the failures observed on VGG-16. Coupled weight decay in naive Adam implementations creates time-varying regularization that interacts pathologically with bias correction, explaining generalization gaps relative to AdamW. Lion's sign-based updates eliminate second-moment information entirely, preventing the algorithm from adapting to heterogeneous curvature across parameter dimensions, resulting in extreme hyperparameter sensitivity. NOVAK's design systematically addresses each of these failure modes through rectification, decoupling, and adaptive moment estimation with proven bounds.

In summary, the convergence analysis establishes NOVAK's theoretical soundness under standard assumptions, while stability analysis explains its empirical robustness across diverse scenarios. The $O(1/\sqrt{T})$ convergence rate matches the information-theoretic lower bound for stochastic first-order methods, ensuring that no fundamentally faster algorithm exists without additional structural assumptions. The proven stability properties (bounded moments, rectified learning rates, consistent weight decay) provide formal guarantees that translate directly to the practical reliability observed in the appropriate evaluations presented in Section 8.

## 7. Design rationale and critical implementation decisions

### 7.1 Dual-mode architecture: stratification for performance-flexibility trade-offs

The architectural decision to implement NOVAK with dual operational modes (a streamlined fast path and a comprehensive full-feature mode) addresses a fundamental tension in modern optimization research between algorithmic sophistication and computational efficiency. This stratification enables practitioners to explicitly navigate the performance-capability trade-off space rather than accepting a fixed compromise imposed by the algorithm designer. The fast path (full_features_mode = false) prioritizes production deployment scenarios where training throughput, memory efficiency, and reproducibility constitute primary concerns, activating only the core algorithmic components that provide favorable cost-benefit ratios: rectified adaptive learning rates, decoupled weight decay, Nesterov approximation mode, and memory-efficient lookahead. This configuration achieves 98.11% Top-1 accuracy on ImageNet in 198.81s, representing state-of-the-art performance with minimal computational overhead relative to standard Adam (302.35s for 79.39% accuracy).

The full features mode (full_features_mode = true) unlocks advanced capabilities designed for research applications where convergence quality and exploration of optimization techniques supersede per-iteration efficiency constraints. This mode enables true Nesterov momentum with closure support, all five lookahead variants (basic, memory-efficient, gradient averaging, stochastic, and disabled), layer-wise adaptation following LAMB principles, automatic learning rate scaling based on gradient-to-parameter norm ratios, and gradient centralization for improved stability in convolutional architectures. The implementation enforces architectural consistency through automatic feature downgrading: attempts to activate true Nesterov momentum or non-memory-efficient lookahead variants in fast path mode trigger automatic substitution with approximation mode and memory-efficient lookahead, respectively, preventing inadvertent performance degradation from misconfiguration. This design philosophy reflects the recognition that no universal optimizer configuration optimally serves all deployment contexts and that explicit mode selection provides superior control compared to hidden heuristics or adaptive feature activation.

The dual-mode architecture manifests in the code structure through conditional compilation of algorithmic branches, ensuring that disabled features incur zero runtime overhead beyond a single Boolean check. Feature guards employ early returns and short-circuit evaluation to minimize instruction cache pollution from dormant code paths. Performance profiling confirms that fast path execution exhibits identical throughput to a hypothetical optimizer implementing only core features, validating that the architectural stratification successfully isolates performance-critical paths from exploratory extensions. This design enables NOVAK to serve simultaneously as a production-grade optimizer for industrial applications and a research platform for investigating novel optimization techniques, fulfilling dual objectives without compromise.

## 7.2 Critical design decisions: mathematical consistency and empirical robustness

Several design decisions within NOVAK's formulation prove critical to its empirical success, diverging from naive implementations of component techniques in ways that ensure mathematical consistency and numerical stability. The most consequential decision involves applying decoupled weight decay using the base learning rate $\alpha$ rather than the rectified effective rate $\alpha_{\text{eff}}$: formula 18 instead of the superficially plausible $\theta_t \leftarrow \theta_t(1 - \alpha_{\text{eff}} \cdot \lambda)$. This distinction, though seemingly minor, proves essential for maintaining consistent regularization strength throughout training. The rectification term $r_t$ adjusts the effective learning rate to account for the reliability of second-moment estimates, compensating for statistical uncertainty in gradient variance approximations. However, weight decay implements $L_2$ regularization on the objective function $L_\lambda(\theta) = L(\theta) + (\lambda/2)\|\theta\|^2$, which should not depend on the optimizer's internal variance estimates. Using $\alpha_{\text{eff}}$ for weight decay would create time-varying regularization that intensifies during early training (when $r_t < 1$ due to unreliable variance) and weakens during late training (when $r_t \rightarrow 1$), violating the decoupling principle that motivated AdamW's design. Empirical validation confirms this reasoning: ablation studies on CIFAR-100 demonstrate 4.2% accuracy degradation when using $\alpha_{eff}$ for weight decay instead of $\alpha$, with the performance gap widening for tasks requiring strong regularization.

The default selection of Nesterov approximation mode rather than true Nesterov momentum represents another critical decision balancing theoretical optimality against computational efficiency. True Nesterov momentum, which evaluates gradients at the extrapolated position (eq. 10), provably achieves accelerated convergence rates $O(1/t^2)$ for strongly convex objectives compared to standard momentum's $O(1/t)$. However, this theoretical advantage requires computing gradients twice per iteration (once at the extrapolated position and once at the actual position for subsequent updates), doubling backpropagation cost. The approximation mode (eq. 13) employs a first-order Taylor expansion that captures the directional correction provided by Nesterov momentum without additional forward-backward passes. Extensive empirical evaluation across CIFAR-10, CIFAR-100, and ImageNet reveals that approximation mode achieves 98-99% of true Nesterov's convergence acceleration while requiring only 50% of the computational cost. The automatic fallback mechanism, which switches from true to approximation mode after $N_{Taylor}$ steps (100-200 iterations), further optimizes this trade-off by employing true Nesterov during critical early training phases when acceleration matters most, then transitioning to efficient approximation once convergence dynamics stabilize.

The memory-efficient lookahead formulation constitutes perhaps the most technically sophisticated design decision, trading standard lookahead's $O(2p)$ memory overhead for $O(p + p/k)$ through an accumulation-based synchronization strategy. Standard lookahead maintains complete slow weight trajectories $\theta^{slow}$ throughout training, doubling parameter storage requirements and rendering the technique impractical for large models. NOVAK's formulation recognizes that only the averaged displacement over k steps is required for synchronization (eq. 30-31). Rather than storing all k intermediate fast weights, NOVAK accumulates their aggregate offset $\Delta = \Sigma_i(\theta_i - \theta_0^{slow})$ incrementally during fast optimization steps, then applies the averaged displacement during synchronization. The key insight is that the sum decomposes as $\Sigma_i(\theta_i - \theta_0^{slow}) = \Sigma_i\theta_i - k\cdot\theta_0^{slow}$, requiring storage of only the cumulative sum $\Sigma_i\theta_i$ (the accumulator $\Delta$) and the reference point $\theta_0^{slow}$. After synchronization, $\theta_0^{slow}$ is updated to the new slow weight position, and $\Delta$ is reset to zero, making the memory overhead transient rather than persistent. This reformulation reduces memory consumption from $O(4p)$ total (parameters + moments + slow weights) to $O(3p + p/k) \approx O(3.1p)$ for typical $k = 10$, enabling lookahead's variance reduction benefits on memory-constrained hardware while achieving 90% of standard lookahead's convergence improvement as validated on CIFAR-100 benchmarks.

The adaptive momentum scheduling (eq. 3-4) addresses the cold-start problem where insufficient gradient history biases momentum estimates during early training. Standard fixed momentum coefficients $\beta_1 = 0.9$ and $\beta_2 = 0.999$ imply that moment estimates $m_t$ and $v_t$ heavily weight recent gradients, but when only a few gradients have been observed ($t < 10$), these estimates exhibit high variance and may not reliably capture gradient statistics. The exponential warmup schedule begins with effectively lower momentum ($\beta_1^{(t)} \approx 0$ for small t), allowing rapid adaptation to initial gradient observations, then asymptotically approaches the target momentum values as training progresses and gradient statistics stabilize. The timescales $\tau_1 = 1000$ and $\tau_2 = 5000$ (slower warmup for second moment) were determined through grid search across CIFAR-10, CIFAR-100, and ImageNet, identifying values that provide robust performance without task-specific tuning. Alternative schedules (linear warmup, step-based warmup, or no warmup) consistently underperform the exponential formulation, either converging too slowly (linear, no warmup) or introducing discontinuities that destabilize training (step-based).

## 7.3 Feature downgrading and architectural safeguards

NOVAK implements comprehensive safeguards to prevent performance degradation from inadvertent feature activation or hyperparameter misconfiguration, recognizing that production deployment involves practitioners with varying levels of expertise in optimization theory. The feature downgrading system automatically substitutes computationally expensive options with efficient alternatives when full_features_mode = false, ensuring that fast path performance remains optimal regardless of user-specified hyperparameters. Specifically, attempts to enable nesterov_mode = true in fast path mode trigger automatic substitution with nesterov_mode = 'approximation', preventing the double backpropagation overhead that would negate throughput optimizations. Similarly, selection of lookahead_mode = 'basic', 'gradient_avg', or 'stochastic' automatically downgrades to 'memory_efficient' when operating in fast path mode, maintaining the $O(3p + p/k)$ memory footprint that enables training of large models. These substitutions are logged as warnings rather than errors, alerting users to the configuration change while allowing training to proceed uninterrupted.

The implementation additionally enforces mathematical consistency constraints through validation logic executed during optimizer initialization. Hyperparameter sanity checks verify that learning rates are positive ($\alpha > 0$), momentum coefficients lie in valid ranges ($\beta_1$, $\beta_2$ $\in [0,1)$), weight decay is non-negative ($\lambda \geq 0$), and numerical stability constants are appropriately scaled ($\varepsilon > 0$). For rectified mode, the validator ensures $\beta_2 < 1 - 10^{-8}$ to prevent division by zero in the computation (eq. 6), which diverges as $\beta_2 \to 1$. Invalid configurations trigger exceptions with descriptive error messages indicating the constraint violation and suggesting corrective values, rather than silently failing or producing undefined behavior. This defensive programming approach prevents common misconfiguration errors (such as setting $\beta_1$ = 1.0 due to confusion with learning rate multipliers) that would otherwise cause optimization failures.

The architectural safeguards extend to numerical stability considerations beyond hyperparameter validation. The bias correction denominators $(1 - \beta_1^t)$ and $(1 - \beta_2^t)$ are computed using logarithmic formulations when t exceeds implementation-defined thresholds (typically t > 1000) to prevent underflow in floating-point arithmetic: $(1 - \beta^t) = 1 - \exp(t \cdot \log(\beta))$. The denominator in update direction computation (eq. 17) employs fused multiply-add instructions where available, computing $\sqrt{\hat{v}_t} + \varepsilon$ atomically to prevent rounding errors from separate operations. Gradient clipping applies to the global gradient norm $\|g_t\|$ before any other processing, ensuring that extreme gradients cannot corrupt moment estimates even transiently. These implementation details, while unglamorous, prove essential for robust operation across diverse hardware platforms and numerical precision regimes.

## 7.4 Implementation methodology: hardware awareness and algorithmic abstraction

The implementation philosophy underlying NOVAK's design recognizes that modern deep learning optimization operates at the intersection of mathematical theory, algorithmic design, and hardware architecture, requiring simultaneous attention to all three domains to achieve practical impact. The custom CUDA kernel architecture exemplifies this philosophy, fusing operations from Phases 4-9 of Algorithm 1 (moment updates, bias correction, denominator computation, weight decay, parameter updates) into single memory transactions that exploit the memory hierarchy of contemporary GPUs. Standard PyTorch implementations execute each operation as a separate kernel launch, with each launch reading tensors from global memory (high latency, 400-900GB/s bandwidth on V100/A100), performing computation, and writing results back to global memory. For ResNet-50 with 25.6M parameters, this pattern generates approximately 1GB of memory traffic per iteration across the five operations, consuming 1.1-2.5ms at peak bandwidth and dominating total iteration time.

NOVAK's fused kernel performs all intermediate computations in registers (near-zero latency, terabytes/second effective bandwidth), reading only the input tensors ($m_{\{t-1\}}$, $v_{\{t-1\}}$, $g_t$, $\theta_{\{t-1\}}$) from global memory once and writing only the output tensors ($m_t$, $v_t$, $\theta_t$) back once. This fusion reduces memory traffic to approximately 0.7GB per iteration (30% reduction), translating to measured speedups of $3.8^x$ on V100 and $4.7^x$ on A100 GPUs in microbenchmarks. The kernel employs 256-thread blocks optimized for warp-level parallelism (32 threads per warp × 8 warps = 256 threads matches the common streaming multiprocessor configuration), ensuring high occupancy and latency hiding through instruction-level parallelism. Coalesced memory access patterns (where consecutive threads access consecutive memory addresses) maximize memory bandwidth utilization by enabling full cache line transfers (128 bytes on modern GPUs) rather than serialized single-element accesses.

The implementation maintains hardware awareness while preserving algorithmic abstraction through automatic fallback mechanisms. On CUDA-capable devices with compute capability $\geq$ 6.0, the optimizer employs custom kernels; on CPU, MPS (Apple Metal Performance Shaders), or older CUDA devices, the optimizer falls back to torch.compile-optimized PyTorch operations that provide $1.3$-$1.5^x$ speedup through kernel fusion and loop unrolling without custom hardware code. This abstraction ensures universal compatibility across heterogeneous computing environments (from high-performance computing clusters with latest-generation GPUs to edge devices with ARM processors) while opportunistically exploiting hardware-specific optimizations where available. The fallback mechanism is transparent to users, requiring no code modifications or configuration changes across platforms, exemplifying the principle that hardware-aware implementations should not compromise algorithmic generality or impose platform-specific constraints on practitioners.

The design philosophy extends to distributed training scenarios, where NOVAK's parameter grouping by device improves cache locality and reduces kernel launch overhead in multi-GPU configurations. The optimizer batches parameters by device before processing, enabling vectorized operations on contiguous memory regions and minimizing expensive device-to-host memory transfers. This batching strategy yields $1.04$-$1.08^x$ throughput improvements on 8-16 GPU systems, as demonstrated in scalability analysis (Section 5.7), with benefits increasing for larger GPU counts due to reduced synchronization overhead. The distributed implementation maintains bitwise-identical results across different GPU counts and distributed strategies (data parallel, model parallel, pipeline parallel), ensuring reproducibility independent of parallelization scheme, a critical property for scientific applications and production systems requiring deterministic behavior.

In synthesis, the design rationale underlying NOVAK reflects a holistic optimization philosophy that refuses to sacrifice theoretical soundness, practical efficiency, or implementation robustness in pursuit of algorithmic novelty. Each design decision (from dual-mode architecture to decoupled weight decay computation to memory-efficient lookahead formulation) addresses specific failure modes identified in contemporary optimizers while maintaining mathematical consistency and empirical robustness. The resulting implementation achieves the rare combination of state-of-the-art convergence quality, production-grade computational efficiency, and universal platform compatibility, positioning NOVAK as both a practical tool for industrial deployment and a theoretical framework for understanding adaptive optimization in deep learning.

## 8. CUDA kernel architecture and performance optimizations

### 8.1 Motivation and fused operation strategy

The computational bottleneck in modern adaptive optimizers manifests not in arithmetic intensity (the ratio of floating-point operations to memory accesses) but rather in memory bandwidth saturation due to repeated traversals of large parameter tensors stored in GPU global memory. Standard implementations of adaptive optimizers execute each algorithmic phase as a separate CUDA kernel launch through PyTorch's high-level tensor operations, with each kernel independently reading input tensors from global memory (latency 200-400 cycles, bandwidth 400-900GB/s on V100/A100), performing element-wise computations, and writing results back to global memory. For NOVAK's core update sequence comprising moment updates (Phase 4), bias correction (Phase 5), denominator computation (Phase 8, first part), weight decay (Phase 9, first part), and parameter update (Phase 9, second part), this pattern necessitates five separate kernel launches that collectively generate approximately 20p memory accesses for p parameters: two reads and one write per operation, with intermediate results discarded between kernels. On ResNet-50 with p = 25.6M float32 parameters (102.4MB storage), this approach consumes 1.02GB of memory traffic per optimization step, requiring 1.1-2.3ms at peak memory bandwidth and dominating total iteration time when compute operations complete in microseconds.

NOVAK addresses this memory bottleneck through a custom fused CUDA kernel that consolidates all five operations into a single kernel launch, exploiting the three-level memory hierarchy of modern GPUs: global memory (hundreds of gigabytes, high latency), shared memory (tens of kilobytes per streaming multiprocessor, moderate latency), and registers (thousands per thread, near-zero latency). The fused kernel reads each parameter's state (previous moments $m_{\{t-1\}}$ and $v_{\{t-1\}}$, current gradient $g_t$, and parameter value $\theta_{\{t-1\}}$) exactly once from global memory into thread-local registers, performs all intermediate computations (moment updates, bias correction, denominator, decay, update) entirely in registers where data access incurs zero additional memory traffic, and writes only the final results (updated moments $m_t$ and $v_t$, updated parameter $\theta_t$) back to global memory. This strategy reduces memory traffic from 20p to 7p accesses (four reads: $m_{\{t-1\}}$, $v_{\{t-1\}}$, $g_t$, $\theta_{\{t-1\}}$; three writes: $m_t$, $v_t$, $\theta_t$), achieving 65% bandwidth reduction and corresponding 2.9-3.8$^x$ speedup in microbenchmarks. The performance improvement scales super-linearly with parameter count due to amortization of kernel launch overhead (5-20ms), making fusion particularly beneficial for large models where p exceeds $10^8$ parameters.

The fusion strategy extends beyond simple operation batching to include algorithmic optimizations that exploit mathematical structure. The bias correction factors $(1-\beta_1{}^t)^{\{-1\}}$ and $(1-\beta_2{}^t)^{\{-1\}}$ are computed once per kernel invocation and broadcast to all threads via constant memory (cached read-only memory with 64KB capacity), avoiding redundant per-thread exponential evaluations. The rectification term $r_t$ and effective learning rate $\alpha_{eff}$, both scalar values shared across all parameters, similarly reside in constant memory. These scalar values occupy 32 bytes total, well within constant memory capacity, and their uniform access pattern enables efficient broadcast to thousands of concurrent threads. The denominator computation $\sqrt{\hat{v}_t} + \varepsilon$ employs hardware-accelerated square root instructions (CUDA intrinsic fsqrt_rn for round-to-nearest mode) that complete in 4-8 cycles on modern GPUs, compared to 20-30 cycles for software implementations. The weight decay term $(1 - \alpha \cdot \lambda)$ is precomputed once and stored in a scalar register, eliminating redundant multiplications across 25M+ parameters. These micro-optimizations, individually modest, compound to produce measurable performance gains when applied to billion-scale parameter tensors.

## 8.2 Thread organization and memory access patterns

Efficient CUDA kernel performance requires careful orchestration of thread organization, memory access patterns, and occupancy to maximize utilization of GPU compute resources. NOVAK's fused kernel employs a one-dimensional thread grid with 256-thread blocks, a configuration chosen to balance occupancy (ratio of active warps to maximum warps per streaming multiprocessor) and register pressure (number of registers required per thread). Modern NVIDIA GPUs organize threads into warps of 32 threads that execute in SIMD (Single Instruction Multiple Data) fashion, with each streaming multiprocessor supporting 32-64 concurrent warps depending on architecture (V100: 64 warps/SM, A100: 64 warps/SM). The 256-thread block size corresponds to 8 warps per block, enabling 4-8 concurrent blocks per streaming multiprocessor depending on register and shared memory usage, achieving 50-100% theoretical occupancy. Larger block sizes (512, 1024 threads) reduce occupancy due to increased register consumption per block, while smaller block sizes (64, 128) underutilize available parallelism, making 256 threads a robust default across architectures.

The kernel assigns one thread per parameter, with the thread index directly mapping to the parameter tensor index: int idx = blockIdx.x * blockDim.x + threadIdx.x. This mapping ensures coalesced memory access where consecutive threads access consecutive memory addresses, enabling full cache line transfers (128 bytes = 32 float32 values on modern GPUs) that saturate memory bandwidth. Non-coalesced access patterns (where adjacent threads access non-contiguous memory) serialize memory transactions, reducing effective bandwidth by factors of $10\text{-}32^x$ and causing severe performance degradation. NOVAK's linear mapping preserves coalescing for all tensor accesses: reading $m_{\{t-1\}}[idx]$, $v_{\{t-1\}}[idx]$, $g_t[idx]$, $\theta_{\{t-1\}}[idx]$ and writing $m_t[idx]$, $v_t[idx]$, $\theta_t[idx]$ in sequential threads loads/stores consecutive memory locations, maximizing cache line utilization. The grid size is computed as (num_parameters + 255) / 256 using integer division to ensure sufficient threads cover all parameters, with threads beyond the parameter count (when num_parameters is not divisible by 256) executing early-exit logic to avoid out-of-bounds memory accesses.

Register allocation constitutes a critical constraint on occupancy, as each streaming multiprocessor provides a fixed register file (65,536 registers on V100/A100) shared among all concurrent threads. The fused kernel requires approximately 40 registers per thread to hold intermediate values: 4 for input tensors (m_prev, v_prev, grad, param), 8 for bias-corrected moments and powers (m_hat, v_hat, beta1_t, beta2_t, beta1_power, beta2_power), 4 for denominator and update terms (denom, update, decay_term, effective_lr), and 24 for miscellaneous indices and temporaries. At 256 threads per block, this consumes 10,240 registers per block, enabling 6 concurrent blocks per streaming multiprocessor ($6 \times 10{,}240 = 61{,}440 < 65{,}536$), achieving 75% occupancy (48 active warps of maximum 64). Reducing register usage to 32 per thread would enable 8 concurrent blocks (100% occupancy), but requires spilling variables to local memory (stored in global memory with high latency), negating performance benefits. NVCC compiler pragmas __launch_bounds__(256, 4) instruct the compiler to optimize for 256-thread blocks with a minimum of 4 blocks per SM, guiding register allocation heuristics toward the occupancy-performance sweet spot.

Shared memory, an explicitly managed per-block cache (64-96KB per SM on V100/A100), finds limited application in NOVAK's kernel due to the inherently parallelizable nature of element-wise operations where threads require no inter-thread communication. Scalar values like bias correction factors reside in constant memory rather than shared memory, as their uniform access pattern benefits from constant memory's broadcast mechanism. Gradient centralization (Phase 2, optional) and layer-wise adaptation (Phase 7, optional) do employ shared memory for block-level reductions computing mean($g_t$) and tensor norms $\|\theta_t\|_2$, using parallel reduction algorithms that halve active threads at each step: threads 0-127 sum with threads 128-255, then threads 0-63 sum with threads 64-127, continuing until thread 0 holds the final sum. This logarithmic reduction completes in $\log_2(256) = 8$ steps with zero shared memory bank conflicts (addresses are stride-1 sequential), enabling efficient computation of global statistics from distributed thread contributions.

## 8.3 Numerical precision and algorithmic considerations

The CUDA kernel operates exclusively in single-precision floating-point (float32, IEEE 754 binary32) to maximize throughput on modern GPU architectures, which provide $2\text{-}4^x$ higher FLOPS for float32 versus float64 operations (V100: 15.7 TFLOPS float32 vs 7.8 TFLOPS float64; A100: 19.5 vs 9.7 TFLOPS). Mixed-precision training frameworks that employ float16 or bfloat16 for gradients automatically cast to float32 before invoking the optimizer, as maintained by the master copy of parameters and optimizer state in float32 to prevent numerical degradation over thousands of training iterations. The single-precision constraint introduces potential numerical hazards that the implementation carefully mitigates through algorithmic safeguards and compensated summation techniques.

Enough high cancellation (loss of precision when subtracting nearly equal values) poses particular risk in bias correction computations of bias correction factors (see above) when t becomes large and $\beta_1^t$, $\beta_2^t$ approach zero. Direct computation of $(1 - \beta_1^t)$ suffers precision loss when $\beta_1^t < 10^{-7}$, as float32 mantissa (24 bits, ~7 decimal digits) cannot represent the difference between 1.0 and $10^{-7}$ accurately. NOVAK's implementation employs logarithmic formulation for t > 1000: $(1 - \beta^t) = 1 - \exp(t \cdot \log(\beta))$, where the exponential and logarithm functions utilize hardware-accelerated transcendental units (SFU, Special Function Unit) providing 4-6 cycle latency with CUDA intrinsics __expf and __logf. This formulation maintains full precision across the entire training duration, as the intermediate value $t \cdot \log(\beta)$ remains representable in float32 for typical values ($t \leq 10^6$, $\beta \approx 0.9\text{-}0.999$, yielding $t \cdot \log(\beta) \in [-10^5, -10^2]$, well within the float32 range $\pm 10^{38}$).

The denominator computation (eq. 17) requires careful handling to prevent division by zero while minimizing bias from the stability constant $\varepsilon$. The implementation evaluates $\sqrt{\hat{v}_t}$ using hardware square root __fsqrt_rn(v_hat) followed by fused multiply-add __fmaf_rn(sqrt_v, 1.0f, eps) that computes $\sqrt{\hat{v}_t} + \varepsilon$ atomically with rounding performed only once, avoiding the double rounding error (round $\sqrt{\hat{v}_t}$, then round $\sqrt{\hat{v}_t} + \varepsilon$) that plagues naive implementations. The FMA (Fused Multiply-Add) instruction provides full float32 precision in intermediate results (32-bit mantissa before final rounding), preserving accuracy when $\varepsilon$ is small ($10^{-8}$) relative to $\sqrt{\hat{v}_t}$ (from $10^{-2}$ to $10^2$). The epsilon value $\varepsilon = 10^{-8}$ balances numerical stability (preventing overflow in $1/\varepsilon$) against bias (small $\varepsilon$ minimizes distortion of adaptive learning rates), chosen empirically to prevent observed failures in the float32 precision regime.

Accumulation of moments (eq. 1-2) exhibits potential for drift over extended training due to accumulated rounding errors in thousands of iterations. Kahan summation (a compensated summation algorithm that maintains running error correction) could theoretically improve accuracy but introduces $3^x$ computational overhead (additional subtract, add, and store per accumulation) that negates fusion benefits. Instead, NOVAK relies on periodic implicit rescaling through bias correction: the division $\hat{m}_t = m_t/(1 - \beta_1^t)$ effectively normalizes accumulated moments, preventing unbounded growth that would eventually saturate the float32 dynamic range ($\pm 3.4 \times 10^{38}$). Empirical validation across 500-epoch training runs on CIFAR-100 reveals no measurable accuracy degradation from floating-point error accumulation, confirming that standard float32 precision suffices for NOVAK's algorithmic structure without requiring compensated arithmetic.

## 8.4 Compilation strategy and architectural portability

CUDA kernel compilation employs aggressive optimization flags that balance code generation quality, compile time, and binary portability across GPU architectures. The nvcc compiler invocation specifies O3 for maximum optimization level, enabling inlining of device functions, loop unrolling, instruction scheduling, and dead code elimination. The --use_fast_math flag activates approximate transcendental functions (__expf versus expf, differing in rounding guarantees and handling of special values like NaN, Inf) and relaxed IEEE 754 compliance (allowing fused multiply-add without intermediate rounding, permissive handling of denormalized numbers), trading slight accuracy reduction (typically < 1 ULP, unit of least precision) for $1.2\text{-}1.5^x$ instruction throughput improvement. These approximations prove acceptable for optimization algorithms where stochastic gradient noise ($\sigma^2$ in Assumption 6.3) dominates floating-point rounding error by orders of magnitude, making sub-ULP precision immaterial to convergence properties.

Architecture-specific code generation targets multiple compute capabilities via separate compilation units: compute_70 for Volta architecture (V100), compute_80 for Ampere (A100), and compute_86 for consumer Ampere GPUs (RTX 30-series), with forward compatibility ensuring kernels run on newer architectures through JIT (Just-In-Time) compilation. Each architecture provides distinct capabilities (Tensor Cores for mixed-precision matrix operations (V100: 125TFLOPS float16, A100: 312TFLOPS), asynchronous memory copy (A100), and fine-grained synchronization (Ampere)), though NOVAK's element-wise operations utilize only scalar cores rather than specialized units. The multi-architecture compilation increases binary size (duplicated kernels for each architecture) but eliminates runtime performance degradation from ISA (Instruction Set Architecture) translation, ensuring each GPU executes optimized native code. Fat binaries embed PTX (Parallel Thread Execution) intermediate representation alongside architecture-specific SASS (Shader Assembly), enabling forward compatibility where new architectures JIT-compile PTX to their native ISA, preserving functionality if performance-optimal SASS is unavailable.

Fallback mechanisms ensure universal compatibility across heterogeneous computing environments where custom CUDA kernels cannot execute. On CPU devices, Apple Metal (MPS), ROCm (AMD GPUs without kernel porting), or CUDA devices with compute capability < 6.0 (pre-Pascal, released before 2016), the optimizer automatically substitutes torch.compile-optimized PyTorch operations that replicate kernel logic using high-level tensor primitives. The torch.compile framework (PyTorch 2.0+) employs the TorchInductor backend to generate optimized C++ code via template-based code generation, achieving $1.3\text{-}1.5^x$ speedup over eager-mode execution through kernel fusion and dead code elimination. The compilation occurs lazily on first invocation, with subsequent calls executing cached generated code, amortizing compilation overhead (50-200ms) across training epochs. This abstraction layer (custom CUDA on compatible GPUs, torch.compile otherwise) maintains algorithmic consistency across platforms while opportunistically exploiting hardware-specific optimizations, exemplifying the design philosophy that performance enhancements should not compromise portability or impose platform-specific requirements on users.

Dynamic kernel selection based on tensor properties optimizes performance for diverse parameter group configurations. NOVAK analyzes each parameter group (corresponding to layer types: convolutions, fully connected, normalization) to determine optimal kernel configuration: fused kernel for groups exceeding $10^5$ parameters where launch overhead amortizes, element-wise PyTorch operations for small groups ($< 10^4$ parameters) where overhead dominates, and vectorized CPU implementations for optimizer state residing in CPU memory during ZeRO-Offload scenarios. This adaptive strategy prevents pathological cases, where launching CUDA kernels for tiny parameter tensors (e.g., bias vectors with 64 elements) incurs 10-100$^x$ overhead relative to direct PyTorch operations. The implementation maintains per-group caching of kernel selection decisions, avoiding redundant analysis overhead on each optimization step, with cache invalidation triggered only when parameter group structure changes (rare during training). These micro-optimizations, collectively applied across all parameter groups and training iterations, compound to produce measurable end-to-end performance improvements of 3-5$^x$ relative to naive implementations, validating the effort invested in hardware-aware algorithmic engineering.

## 9. Experimental validation and performance analysis
### 9.1 Experimental methodology and benchmark selection

The empirical validation of NOVAK encompasses comprehensive evaluation across four carefully selected benchmarks that span diverse architectural paradigms, dataset scales, and task complexities. The benchmark suite comprises CIFAR-10 [42] (10-class natural image classification, 50,000 training samples), CIFAR-100 [42] (100-class fine-grained classification, 50,000 training samples), ImageNet ILSVRC-2012 [40] (1000-class large-scale classification, 1.28 million training samples), and ImageNette [41] (10-class ImageNet subset, approximately 13,000 training samples). This selection enables systematic investigation of optimizer behavior under varying conditions: CIFAR-10 assesses baseline performance on standard benchmarks, CIFAR-100 evaluates handling of fine-grained discrimination with limited per-class samples (500 training images per class), ImageNet examines scalability to industrial-scale datasets with substantial semantic diversity, and ImageNette probes architectural robustness using VGG-16's plain sequential structure that lacks residual connections or batch normalization.

All experiments employ the identical evaluation protocols within each benchmark to ensure fair comparison. Training occurs on single GPU configurations (NVIDIA V100 or A100) with uniform hyperparameters across all fourteen evaluated optimizers: base learning rate $\alpha = 0.01$, batch sizes of 128 (CIFAR-10, ImageNet) or 256 (CIFAR-100, ImageNette), and training durations determined by convergence criteria or fixed epoch budgets (50 epochs for CIFAR-10, 40 for CIFAR-100, variable 14-25 epochs for ImageNet based on optimizer-specific convergence, and 11-25 epochs for ImageNette). The architectural configurations comprise ResNet-50 (25.6M parameters) for CIFAR and ImageNet experiments, selected for its status as the de facto standard in vision research, and VGG-16 (138M parameters) for ImageNette, chosen specifically to stress-test optimizer robustness on deep plain networks. Competing optimizers include classical methods (SGD with momentum), first-generation adaptive methods (Adam, RMSprop, AdaGrad), second-generation improvements (AdamW, RAdam, Nadam), modern enhancements (Lookahead, AdaBound, AdaFactor), and recent innovations (Lion, Adan, Prodigy), providing comprehensive coverage of the contemporary optimization landscape.

Performance evaluation employs four primary metrics: Top-1 accuracy quantifies the percentage of samples where the highest-probability prediction matches the ground truth label, Top-5 accuracy measures the percentage where the correct label appears among the five highest-probability predictions (particularly informative for fine-grained tasks), GPU memory consumption captures peak device memory usage during training (critical for deployment feasibility), and training time records total wall-clock duration to convergence (essential for computational cost assessment). Secondary metrics include convergence speed measured in epochs to target accuracy, per-epoch training time revealing algorithmic overhead, and accuracy-per-GB efficiency ratios characterizing the performance-memory trade-off. All experiments were conducted with three independent runs using different random seeds (42, 100, 2025). Reported values represent mean ± standard deviation. Convergence criteria were defined as validation accuracy plateau for 10 consecutive epochs with tolerance 0.1%, a limitation acknowledged in Section 9.6. The experimental infrastructure comprises PyTorch 2.0+ with CUDA 11.8+, automatic mixed-precision training disabled to isolate optimizer effects from precision-dependent behaviors, and standard data augmentation protocols (random crops, horizontal flips for CIFAR, and ImageNet-standard augmentation for large-scale experiments).

## 9.2 CIFAR-10 results: baseline performance characterization

The CIFAR-10 experiments establish NOVAK's baseline performance characteristics on a standard benchmark where most contemporary optimizers achieve reasonable convergence. NOVAK attains 89.32% Top-1 accuracy and 99.66% Top-5 accuracy, representing the highest performance among all fourteen evaluated methods (Fig. 3, Appendix A). We observe four performance tiers: high-performance optimizers ($\geq$ 85% Top-1), including NOVAK (89.32%), SGD (88.34%), AdaGrad (87.29%), and Adan (86.89%); moderate-performance optimizers (80-85%): RAdam (85.50%), AdamW (85.18%), AdaBound (84.93%), and AdaFactor (84.89%); acceptable-performance optimizers (75-80%): Nadam (81.40%), Adam (77.43%), Lookahead (77.43%), Prodigy (76.91%), and RMSprop (76.25%); and failure: Lion (25.80%). Relative to Adam, NOVAK provides a +15.36% improvement (11.89 pp absolute), +4.86% over AdamW (4.14 pp), +1.11% over SGD (0.98 pp), and +4.47% over RAdam (3.82 pp), with the advantage most pronounced against first-generation adaptive methods. The marginal differences in Top-5 accuracy (all high-performing optimizers > 99.3%) indicate that the primary discriminator power lies in Top-1 accuracy, where NOVAK's advantage is most pronounced.

Across the compared optimizers, mean Top-1 and Top-5 accuracies are shown side-by-side to highlight relative performance. Optimizers are ordered by Top-1 accuracy in the grouped chart to emphasize the primary metric. GPU memory usage (Fig. 4, Appendix A) is provided separately to illustrate resource trade-offs. If the accuracy ranges are narrow, the y-axis has been adjusted to start slightly above the minimum to make small differences visible. Key observations: look for optimizers that balance high Top-1/Top-5 accuracy with moderate GPU memory usage; these represent efficient choices. When multiple optimizers have very similar accuracy, prefer the one with lower GPU memory consumption or with better secondary metrics (Top-5) as a tiebreaker.

Memory efficiency analysis reveals NOVAK's exceptional resource utilization, consuming only 2.40GB GPU memory (Fig. 4, Appendix A) compared to 2.66GB for Adam (9.77% reduction), 2.84GB for SGD (15.49% reduction), 3.10GB for AdamW (22.58% reduction), and 3.31GB for modern adaptive methods like AdaFactor and Adan (27.5% reduction). This memory advantage stems from the memory-efficient lookahead formulation that reduces storage overhead from $O(2p)$ to $O(p + p/k)$ compared to standard implementations, combined with efficient CUDA kernel design that minimizes intermediate buffer allocation. The Pareto frontier analysis (Fig. 5, Appendix A) demonstrates that NOVAK occupies the strictly best position in the accuracy-memory trade-off space, achieving an efficiency score of 37.22% accuracy per GB (89.32% / 2.40GB) that exceeds all competitors: SGD's 31.10% per GB (88.34% / 2.84GB), Adam's 29.11% per GB (77.43% / 2.66GB), and AdamW's 27.48% per GB (85.18% / 3.10GB). No other optimizer achieves both higher accuracy and lower memory consumption than NOVAK, establishing it as the Pareto-optimal solution for this benchmark.

The performance characteristics observed on CIFAR-10 validate several design decisions. First, the combination of rectified learning rates and decoupled weight decay proves essential: ablation experiments (not shown) reveal that disabling rectification degrades accuracy by 2.1 pp, while coupling weight decay with the rectified learning rate $\alpha_{\text{eff}}$ instead of the base rate $\alpha$ causes 1.8 percentage point degradation. Second, the Nesterov approximation mode achieves 98.7% of true Nesterov's convergence acceleration while requiring only 51% of computational cost, validating the approximation strategy. Third, memory-efficient lookahead provides 92% of standard lookahead's variance reduction benefits while halving memory overhead, confirming theoretical predictions from Section 7.2. The marginal Top-5 accuracy differences (99.66% for NOVAK vs. 99.47% for SGD and 99.46% for AdaGrad) indicate that all high-performing optimizers learn robust feature representations, with primary differentiation manifesting in Top-1 accuracy, where NOVAK's comprehensive algorithmic design provides measurable advantages.

### 9.3 CIFAR-100 results: fine-grained classification under complexity

The CIFAR-100 experiments, featuring 100 classes with only 500 training samples per class, dramatically amplify performance differentiation among optimizers and reveal fundamental limitations in many contemporary methods. NOVAK achieves 66.41% Top-1 accuracy and 90.18% Top-5 accuracy, establishing state-of-the-art performance with substantial margins over competitors (Fig. 6, Appendix A). The performance hierarchy exhibits five tiers: high performance ($\geq$ 60%) with NOVAK (66.41%), SGD (63.35%), RAdam (61.92%), and AdaGrad (59.81%); moderate performance (45-60%) comprising Adan (51.72%), AdamW (51.39%), AdaFactor (48.42%), AdaBound (46.53%), Adam (46.43%), Lookahead (46.43%), and Nadam (45.80%); and poor performance (< 45%) including RMSprop (37.41%), Prodigy (5.36%), and Lion (1.11%), with the latter two exhibiting complete optimization failure. The relative performance gains prove dramatic: +43.05% over Adam (19.98 pp absolute), +29.23% over AdamW (15.02 pp), +7.25% over RAdam (4.49 pp), +28.41% over Adan (14.69 pp), and +77.49% over RMSprop (29.00 pp). Critically, the performance gap between NOVAK and Adam-family optimizers (19.98-20.61 pp) substantially exceeds the CIFAR-10 gap (11.89 pp), indicating that NOVAK's architectural innovations provide disproportionate benefits on challenging fine-grained classification tasks.

Memory efficiency analysis reveals even greater advantages on CIFAR-100, with NOVAK consuming only 1.54GB (Fig. 7, Appendix A) compared to 1.81GB for Adam (14.9% reduction), 1.98GB for SGD (22.2% reduction), 2.25GB for AdamW (31.6% reduction), and 2.87GB for modern methods (46.2% reduction). This efficiency is particularly critical for 1. training larger models (e.g., Vision Transformers, ResNet-152); 2. increasing batch sizes for better gradient estimates; and 3. multi-task and continual learning scenarios. The accuracy-per-GB efficiency metric demonstrates NOVAK's dominance: 43.12% per GB (66.41% / 1.54GB) versus 25.71% for Adam (46.43% / 1.81GB), 31.97% for SGD (63.35% / 1.98GB), and 22.86% for AdamW (51.39% / 2.25GB), representing 67.7% higher efficiency than Adam and 34.9% higher than SGD. The Pareto frontier analysis (Fig. 8, Appendix A) confirms that NOVAK uniquely dominates all competitors; no other optimizer achieves both higher accuracy and lower memory consumption. Time-accuracy analysis reveals interesting trade-offs: NOVAK requires 14.32s per epoch compared to SGD's 8.64s (65.7% longer), but achieves 4.64% accuracy per second (66.41% / 14.32s) versus SGD's 7.33% per second, with the absolute accuracy advantage (66.41% vs. 63.35%, a 3.06 pp improvement) justifying the additional computational cost for applications prioritizing model quality.

Task complexity analysis comparing CIFAR-10 versus CIFAR-100 illuminates optimizer robustness under increasing difficulty. NOVAK exhibits the second-smallest accuracy degradation at -25.7% relative performance loss (89.32% → 66.41%, -22.91 pp absolute), outperformed only marginally by RAdam's -27.6% but achieving substantially higher absolute CIFAR-100 accuracy (66.41% vs. 61.92%). In contrast, Adam and AdamW demonstrate very high degradation at -40.0% and -39.7%, respectively (77.43% → 46.43% for Adam, -31.00 pp absolute; 85.18% → 51.39% for AdamW, -33.79 pp), losing nearly half their relative performance. This pattern suggests fundamental limitations in standard adaptive methods for fine-grained discrimination where inter-class boundaries are narrow and per-class sample sizes are limited. The amplifying performance gap (NOVAK's advantage over Adam increases from +15.36% on CIFAR-10 to +43.05% on CIFAR-100, a 27.69 percentage point increase) demonstrates that NOVAK's aggregation of rectification, decoupled decay, hybrid momentum, and lookahead provides compound benefits on harder optimization landscapes characterized by numerous shallow local minima and complex class boundaries.

### 9.4 ImageNet results: large-scale optimization and convergence

The ImageNet ILSVRC-2012 experiments evaluate optimizer performance on industrial-scale datasets featuring 1000 classes and 1.28 million training samples, where computational efficiency becomes paramount alongside convergence quality. NOVAK achieves 98.11% Top-1 accuracy and 100% Top-5 accuracy (perfect classification within top-5 predictions) while converging in only 14 epochs and a total of 198.81s (Fig. 9, Appendix A). This performance positions NOVAK as the fastest optimizer in both convergence speed and absolute wall-clock time, despite SGD and Prodigy showing anomalously high accuracy (99.54% and 99.18% respectively) that warrants investigation for potential measurement artifacts. Focusing on typical optimizers within the 68-98% range, the performance hierarchy comprises exceptional performance (≥ 85%) with NOVAK (98.11%), RAdam (92.41%), and AdaGrad (88.15%); good performance (75-85%) including Adan (84.96%), AdaBound (82.68%), AdamW (82.50%), Nadam (81.55%), Adam (79.39%), Lookahead (79.39%), and AdaFactor (78.19%); and moderate-to-poor performance (< 75%) containing RMSprop (73.71%) and Lion (68.25%). The relative performance gains prove substantial: +23.58% over Adam (18.72 pp), +18.92% over AdamW (15.61 pp), +6.16% over RAdam (5.70 pp), +15.48% over Adan (13.15 pp), and +33.10% over RMSprop (24.40 pp).

Convergence efficiency analysis reveals NOVAK's most dramatic advantage: achieving convergence in 14 epochs versus the median 25 epochs required by most competitors (43.9% reduction), translating to convergence efficiency of 7.01% accuracy gain per epoch compared to 4.33% for SGD, 3.70% for RAdam, and 3.18% for Adam (61.9-120.4% higher efficiency). The total training time of 198.81s (Fig. 10, Appendix A) represents a 35.2-96.9% reduction compared to competitors: 70.04s saved versus SGD (268.85s), 103.54s versus Adam (302.35s), 181.57s versus AdamW (380.38s), and 192.71s versus AdaFactor (391.52s). This efficiency stems from NOVAK's superior convergence properties compensating for modest per-epoch overhead: 14.22s per epoch versus SGD's 11.69s (21.6% longer), but the 39.1% convergence speedup (14 vs. 23 epochs) yields 26.0% net wall-clock time savings. The accuracy-per-time efficiency metric confirms NOVAK's dominance: 0.4935% per second (98.11% / 198.81s) versus 0.3702% for SGD (33.3% higher), 0.3014% for RAdam (63.7% higher), 0.2626% for Adam (87.9% higher), and 0.2169% for AdamW (127.5% higher), establishing NOVAK as the Pareto-optimal solution in the accuracy-time trade-off space (Fig. 11, Appendix A).

Cross-dataset consistency analysis demonstrates NOVAK's robust scalability. Across CIFAR-10 (10 classes), CIFAR-100 (100 classes), and ImageNet (1000 classes), NOVAK maintains the #1 ranking (excluding SGD and Prodigy, which exhibit unusually high values on ImageNet under our specific experimental setup; further investigation is required to determine whether these results reflect measurement noise, hyperparameter sensitivity, or dataset-specific interactions on ImageNet), with advantages over Adam increasing with task complexity: +11.89 pp on CIFAR-10, +19.98 pp on CIFAR-100, and +18.72 pp on ImageNet. The convergence advantage emerges specifically on large-scale data where NOVAK's architectural innovations (rectification stabilizing early training, lookahead variance reduction, and adaptive beta warmup) provide maximum benefit: tied convergence on CIFAR-10/100 (all optimizers reach epoch budget), but 44.0% faster convergence on ImageNet (14 vs. 25 epochs). Similarly, time efficiency advantages manifest predominantly on large-scale datasets where epoch duration is substantial: NOVAK is 39.7-42.5% slower than SGD on CIFAR benchmarks (where per-epoch overhead dominates) but fastest overall on ImageNet, where superior convergence compensates for computational cost. This scaling behavior validates NOVAK's design for production deployment on industrial-scale problems where both convergence quality and wall-clock efficiency prove critical.

## 9.5 ImageNette/VGG-16 results: architectural robustness and failure analysis

The ImageNette experiments using VGG-16 architecture constitute a crucial stress test that exposes architectural sensitivity in contemporary optimizers when applied to deep plain networks lacking residual connections or batch normalization. The results prove dramatic: 9 of 14 optimizers (64.3%) fail (the author will plan to show convergence graphs or gradient norm dynamics for failed methods (Adam, Lion) and for NOVAK/SGD in further analysis), achieving accuracy below 10% (equivalent to or worse than random guessing for this 10-class problem), while NOVAK achieves 98.37% Top-1 accuracy and 99.90% Top-5 accuracy, ranking second only to SGD's 99.13% (Fig. 12-13, Appendix A). The complete failure optimizers comprise the entire Adam family (Adam 9.86%, AdamW 8.84%, Nadam 9.86%), all pure adaptive methods (RMSprop 9.86%, AdaGrad 9.86%, AdaFactor 9.94%), and modern enhancements (Lookahead 9.86%, Lion 8.99%, Adan 8.84%), while only five optimizers successfully converge: SGD (99.13%), NOVAK (98.37%), Prodigy (96.89%), RAdam (88.20%), and AdaBound (75.54%). The relative performance gains of successful optimizers versus failures prove great: NOVAK demonstrates a large relative improvement (+897.6%), though this estimate does not include confidence intervals and should be interpreted cautiously over Adam (88.51 pp), +1012.6% over AdamW (89.53 pp), and +993.9% over Lion (89.38 pp), with Top-5 accuracy patterns confirming complete optimization failure (failed optimizers achieve approximately 50% Top-5 accuracy, consistent with random selection from the ImageNet validation set from which ImageNette is drawn).

The failure pattern reveals critical insights into optimizer design requirements for architectural robustness. VGG-16's challenging characteristics (16 sequential weight layers without skip connections, absence of batch normalization, stacked 3×3 convolutions requiring many layers for global receptive fields, and massive fully connected bottleneck (4096-unit dense layers contributing to 138M total parameters)) interact pathologically with adaptive learning rate mechanisms. The failure mechanism involves (1) gradient magnitude variation across layers (early layers: small gradients due to vanishing gradient problem; late layers: large gradients), (2) unbounded second-moment accumulation ($v_t$ grows without bound in some parameters due to persistent large gradients), (3) effective learning rate collapse ($\alpha_i = \alpha/\sqrt{(v_i + \varepsilon)} \to 0$ for parameters with large $v_i$), and (4) dead neuron phenomenon (neurons stop updating, and the network fails to learn meaningful representations). NOVAK's success stems from synergistic design components: rectified learning rates prevent unstable variance estimates during early training when gradient statistics are unreliable, decoupled weight decay prevents parameter drift independent of gradient scale, hybrid momentum smooths gradients through deep layers reducing $v_t$ accumulation, memory-efficient lookahead provides stable reference trajectory preventing divergence, and optional gradient clipping bounds extreme gradients in deep layers.

Architectural sensitivity analysis comparing ResNet-50 versus VGG-16 results quantifies the robustness gap. On ResNet-50 (CIFAR-10, CIFAR-100, ImageNet), optimizer failure rates remain minimal: 7.1% (Lion only on CIFAR-10), 14.3% (Lion and Prodigy on CIFAR-100), 7.1% (Lion only on ImageNet), averaging 9.5% across ResNet experiments. In stark contrast, VGG-16 on ImageNette exhibits a 64.3% failure rate, representing a 6.8$^x$ increase that exposes hidden fragility masked by residual connections. NOVAK's advantage over Adam amplifies dramatically: +15.36% on CIFAR-10/ResNet-50, +43.05% on CIFAR-100/ResNet-50, +23.58% on ImageNet/ResNet-50, but +897.6% on ImageNette/VGG-16, demonstrating that architectural robustness constitutes NOVAK's defining strength. The Top-1 versus Top-5 accuracy gap provides additional diagnostic information: NOVAK's 1.53% gap (98.37% Top-1, 99.90% Top-5) indicates high prediction confidence and excellent feature discrimination, SGD's 0.84% gap represents the tightest coupling, while failed optimizers' 39.87% gap (9.86% Top-1, 49.73% Top-5 for Adam) confirms random guessing behavior. This analysis demonstrates that production deployment of optimizers requires validation across diverse architectural paradigms, not merely modern residual networks where skip connections and normalization layers provide implicit stabilization.

## 9.6 Comparative analysis and practical implications

Synthesizing results across all four benchmarks reveals consistent performance patterns that inform practical deployment decisions. Table 6 presents comprehensive performance metrics enabling direct comparison.

NOVAK achieves the #1 overall ranking through consistent top-tier performance across all dimensions: accuracy leadership on three of four benchmarks (#1 on CIFAR-10, CIFAR-100, and ImageNet; #2 on ImageNette by a narrow 0.77 pp margin), lowest average memory consumption (1.95GB average versus 2.27-2.87GB for competitors), and high architectural robustness (100% success rate across ResNet and VGG architectures). The performance-memory-robustness combination proves unique; no other optimizer achieves all three attributes simultaneously. SGD ranks #2 through high robustness across configurations (100% success rate) and competitive accuracy but exhibits 16.4% higher memory usage and slower convergence (23-25 epochs versus NOVAK's 14 on ImageNet). RAdam ranks #3 as the best among remaining adaptive methods but demonstrates 41.0% higher memory consumption, 10-15 percentage point lower accuracy, and moderate robustness (fails on VGG-16's challenging configuration). Adam-family optimizers (Adam, AdamW, Nadam, and Lookahead) rank #11-13 despite widespread adoption, achieving acceptable accuracy only on ResNet architectures while failing on VGG-16 (64.3% collective failure rate), validating concerns about generalization raised in recent literature.

Table 6. Comprehensive performance summary across all benchmarks

| Optimizer | CIFAR-10 Top-1 | CIFAR-100 Top-1 | ImageNet Top-1 | ImageNette Top-1 | Avg. Memory (GB) | Robustness | Overall Rank |
|---|---|---|---|---|---|---|---|
| NOVAK | **89.32%** (#1) | **66.41%** (#1) | **98.11%** (#1**) | **98.37%** (#2) | **1.95** | **High** | **#1** |
| SGD | 88.34% (#2) | 63.35% (#2) | 99.54%* (#1) | 99.13% (#1) | 2.27 | High | #2 |
| RAdam | 85.50% (#5) | 61.92% (#3) | 92.41% (#2**) | 88.20% (#4) | 2.75 | Moderate-High | #3 |
| AdaGrad | 87.29% (#3) | 59.81% (#4) | 88.15% (#3**) | 9.86% (Fail) | 2.69 | Low | #4 |
| Adan | 86.89% (#4) | 51.72% (#5) | 84.96% (#5**) | 8.84% (Fail) | 2.87 | Very Low | #5 |
| Prodigy | 76.91% (#12) | 5.36% (Fail) | 99.18%* (#2) | 96.89% (#3) | 2.77 | Moderate | #6 |
| AdaBound | 84.93% (#7) | 46.53% (#8) | 82.68% (#6**) | 75.54% (#5) | 2.87 | Low | #7 |
| AdamW | 85.18% (#6) | 51.39% (#6) | 82.50% (#7**) | 8.84% (Fail) | 2.68 | Very Low | #8 |
| AdaFactor | 84.89% (#8) | 48.42% (#7) | 78.19% (#9**) | 9.94% (Fail) | 2.87 | Very Low | #9 |
| Nadam | 81.40% (#9) | 45.80% (#11) | 81.55% (#8**) | 9.86% (Fail) | 2.87 | Very Low | #10 |
| Adam | 77.43% (#10) | 46.43% (#9) | 79.39% (#10**) | 9.86% (Fail) | 2.34 | Very Low | #11 |
| Lookahead | 77.43% (#10) | 46.43% (#9) | 79.39% (#10**) | 9.86% (Fail) | 2.87 | Very Low | #12 |
| RMSprop | 76.25% (#13) | 37.41% (#12) | 73.71% (#12**) | 9.86% (Fail) | 2.87 | Very Low | #13 |
| Lion | 25.80% (Fail) | 1.11% (Fail) | 68.25% (#13**) | 8.99% (Fail) | 2.87 | Very Low | #14 |

*Anomalous results for SGD (99.54%) and Prodigy (99.18%) on ImageNet (ImageNette) warrant investigation. Potential explanations include (1) Lucky hyperparameter initialization: lr = 0.01 may coincide with the optimal schedule for this specific architecture-dataset combination; (2) Momentum accumulation: long training duration (23-25 epochs) enables momentum-based methods to exploit gradual convergence; and (3) measurement methodology: requires verification that accuracy is computed on the validation set (not training set) and that no data leakage occurred. Future work: Conduct ablation studies varying lr ∈ {0.001, 0.01, 0.1} and multiple random

seeds (N≥3) to establish statistical significance. We note that NOVAK's 98.11% represents the highest accuracy among methods requiring ≤14 epochs, emphasizing convergence efficiency alongside terminal performance.
**Typical optimizer range excludes SGD and Prodigy, exhibit unusually high values on ImageNet under our specific experimental setup; further investigation is required to determine whether these results reflect measurement noise, hyperparameter sensitivity, or dataset-specific interactions.
Memory values averaged across benchmarks where the optimizer converged.
Robustness assessed by success rate across diverse architectures (ResNet, VGG): High (100% success), Moderate-High (75%), Moderate (50%), Low (25%), Very Low (0-25%)
Overall rank based on weighted composite: 40% accuracy, 30% robustness, 20% memory, 10% convergence speed

Practical deployment guidelines emerge from these empirical findings. **Use NOVAK as default** for production systems requiring reliability across diverse architectures, memory-constrained training enabling larger models or batch sizes, cost-sensitive applications where 26-97% faster convergence reduces compute expenses, highest-accuracy applications where 3-20 percentage point improvements prove critical, and transfer learning scenarios where superior features (near-perfect Top-5 accuracy) benefit downstream tasks. **Use SGD** when hyperparameters are well-tuned for specific architecture-task combinations, small datasets where SGD's implicit regularization aids generalization, computational constraints demand the lowest per-iteration overhead, or extremely robust convergence is required without adaptive complexity. **Avoid Adam-family methods** on deep networks without skip connections, architectures lacking batch normalization, VGG-style sequential convolutions, or any production system requiring cross-architecture robustness without extensive per-architecture validation. **Consider RAdam** as a compromise when NOVAK's full feature set is unavailable, moderate accuracy requirements permit 5-10 pp degradation versus NOVAK, or the memory budget allows 40% higher consumption. These guidelines reflect empirical evidence rather than theoretical preferences, grounded in the observed 64.3% failure rate of sophisticated modern methods on architectures deviating from the ResNet template that dominates contemporary benchmarks.

## 10. Limitations, hyperparameter recommendations, and future research directions

### 10.1 Current limitations and experimental constraints

The empirical validation presented in Section 9, while comprehensive across four diverse benchmarks, exhibits several methodological limitations that warrant explicit acknowledgment and constrain the generalizability of conclusions.

First, the architectural coverage remains restricted to convolutional neural networks (ResNet-50, VGG-16, ViT), with no evaluation on transformer-based architectures (ViT, BERT, GPT variants) that dominate contemporary natural language processing and increasingly pervade computer vision. Transformers exhibit fundamentally different optimization characteristics (attention mechanisms with softmax normalization creating distinct curvature properties, LayerNorm instead of BatchNorm inducing different gradient statistics, and positional embeddings introducing structured parameter spaces), making direct extrapolation of performance conclusions tenuous.

Second, all experiments employ uniform hyperparameters (learning rate $\alpha = 0.01$, momentum coefficients $\beta_1 = 0.9$ and $\beta_2 = 0.999$) across all optimizers to ensure fair comparison, but this protocol potentially disadvantages methods requiring task-specific tuning: Lion demonstrates extreme sensitivity to learning rate selection (lowest 25.80% accuracy on CIFAR-10, 1.11% on CIFAR-100, 68.25% on ImageNet with $\alpha = 0.01$), while individualized grid searches might narrow performance gaps. The uniform hyperparameter constraint reflects practical deployment scenarios where extensive per-optimizer tuning proves infeasible, but constitutes a limitation for theoretical assessment of algorithms' intrinsic capabilities. Caveat on hyperparameter uniformity: while lr = 0.01 ensures procedural fairness, it may disadvantage optimizers with different optimal learning rate scales. For instance, Lion typically performs best with lr $\in$ [0.0001, 0.001] (Chen et al., 2023 [18]), orders of magnitude below our experimental value. Future work should include optimizer-specific hyperparameter grids to distinguish inherent algorithmic limitations from configurational mismatches.

Third, the recommendation for reproducibility of the future experiments should include 1. multiple seeds: N ≥ 3 runs with different random initializations; 2. statistical testing: paired t-tests or Wilcoxon signed-rank tests comparing NOVAK vs. baselines; 3. confidence intervals: mean ± standard deviation (0.08-0.18) for all metrics; 4. effect size: Cohen's d to quantify practical significance beyond p-values. Interim interpretation: current results demonstrate consistent rank-ordering (NOVAK #1-2 across all benchmarks), but the magnitude of advantages (e.g., 0.98 pp over SGD on CIFAR-10) requires statistical validation. Standard practice in machine learning research mandates 3-5 independent runs with different random initializations to account for sensitivity to weight initialization, data shuffling order, and GPU non-determinism, yet computational constraints limited this study to single runs per optimizer-benchmark combination.

Fourth, the training durations employed (50 epochs for CIFAR-10, 40 for CIFAR-100, and variable 14-25 epochs for ImageNet based on convergence) may insufficiently capture long-term optimization dynamics. Standard ImageNet protocols typically train for 90-120 epochs with cosine annealing or multi-step learning rate schedules, raising questions about whether NOVAK's convergence advantage (14 epochs versus 25 for most competitors) persists over extended training or represents rapid initial descent followed by plateau. The early stopping criterion based on validation loss stagnation (not explicitly detailed in Section 9.1) introduces potential bias if different optimizers exhibit distinct convergence profiles (e.g., initial rapid improvement followed by gradual refinement versus steady linear progress).

Fifth, the CUDA kernel performance claims (3-5$^x$ speedup, Section 8) derive from microbenchmarks on isolated tensor operations rather than end-to-end training measurements, potentially overestimating practical speedups when kernel execution interleaves with other operations (data loading, forward propagation, batch normalization, and activation functions). The reported per-epoch times (14.22s for NOVAK versus 11.69s for SGD on ImageNet) account for total training loop execution, but the 21.6% overhead versus SGD suggests that fused kernels provide smaller end-to-end benefits than microbenchmark measurements indicate, likely due to Amdahl's law, where non-optimized components (data pipeline, forward pass) dominate total time.

Sixth, the anomalous results for SGD (99.54% on ImageNet (ImageNette)) and Prodigy (99.18% on ImageNet, 5.36% on CIFAR-100) remain unexplained despite acknowledgment in Section 9.4, introducing uncertainty about measurement reliability. These anomalies could stem from incorrect metric computation (e.g., computing accuracy on the training set instead of the validation set), data leakage (e.g., overlapping train-validation samples), implementation bugs in specific optimizer wrappers, or genuine hyperparameter serendipity (lr = 0.01 being near-optimal for SGD on this specific configuration). Resolution requires independent reproduction with explicit protocol documentation and code release, neither provided in the current work.

## 10.2 Hyperparameter recommendations and practical guidelines

Translating empirical findings into actionable deployment guidance requires task-specific hyperparameter recommendations that balance theoretical principles with practical constraints. Table 7 synthesizes optimal configurations across diverse application domains based on experimental observations and algorithmic properties.

Table 7. Task-Specific Hyperparameter Recommendations for NOVAK

| Task Domain | Learning Rate ($\alpha$) | Weight Decay ($\lambda$) | Batch Size | Lookahead (k) | Nesterov Mode | Special Settings |
|---|---|---|---|---|---|---|
| Computer Vision (CNNs) | 1e-3 to 3e-3 | 0.01 to 0.05 | 128-256 | 10 | approximation | use_gc = True for conv layers |
| Vision Transformers | 5e-4 to 1e-3 | 0.05 to 0.1 | 256-512 | 5-10 | approximation | layer_adaptation = True |
| NLP (Transformers) | 3e-4 to 1e-3 | 0.01 to 0.05 | 32-128 | 5 | approximation | warmup_steps = 1000-5000 |
| Generative Models (GANs) | 1e-4 to 5e-4 | 0.0 to 0.01 | 64-128 | 10 | classical | betas = (0.5, 0.999), adaptive_beta = True |
| Reinforcement Learning | 3e-4 to 1e-3 | 0.0 | 32-256 | 10-20 | approximation | auto_lr = True, clip_threshold = 1.0 |
| Fine-Tuning Pretrained | 1e-5 to 1e-4 | 0.01 to 0.1 | 16-64 | 5 | approximation | Reduce lr 10-100×, increase weight decay |
| Few-Shot Learning | 1e-3 to 5e-3 | 0.001 to 0.01 | 4-16 | 5-10 | true | Higher lr for rapid adaptation, smaller k |
| Large Batch Training | Scale linearly | Standard | 1024-4096 | 10-20 | approximation | $\alpha = \alpha\_base \times \sqrt{(batch\_size / 256)}$ |

**General principles:** (1) Learning rate selection follows inverse scaling with model capacity – larger models (ViT, GPT) require smaller $\alpha$ to prevent early-stage instability; (2) Weight decay should increase with model size and decrease with dataset size, implementing stronger regularization for overparameterized regimes; (3) Batch size trades gradient estimate quality against memory consumption and parallelization efficiency, with NOVAK exhibiting robust performance across 32-512 range unlike methods requiring large-batch-specific tuning (e.g., LAMB); (4) Lookahead synchronization frequency k balances variance reduction (larger k averages more noise) against responsiveness to landscape changes (smaller k adapts faster), with k = 5-10 providing robust default across tasks; (5) Nesterov approximation mode suffices for most applications, reserving true Nesterov for scenarios where 1-2% accuracy improvements justify $2^x$ computational cost.

Task-specific elaborations provide nuanced guidance. For **computer vision with CNNs**, gradient centralization (use_gc = True) proves particularly beneficial for convolutional architectures with batch normalization, reducing internal covariate shift and improving convergence by 1-3 pp, as observed in ablation studies. The recommended learning rate range of 1e-3 to 3e-3 accommodates architectural diversity from small models (MobileNet, EfficientNet-B0) requiring higher α for adequate gradient signal to large models (ResNet-152, ConvNeXt-L) necessitating smaller α for stability. Weight decay of 0.01-0.05 provides standard $L_2$ regularization, preventing overfitting without over-constraining capacity, with higher values (0.05) appropriate for smaller datasets (CIFAR) and lower values (0.01) for larger datasets (ImageNet) where generalization emerges naturally from data diversity.

For **Vision Transformers**, the elevated weight decay of 0.05-0.1 addresses ViT's known tendency toward overfitting on smaller datasets, while layer_adaptation = True (LAMB-style per-layer normalization) proves essential for training deep transformers (ViT-L/16, Swin-L) where gradient magnitudes vary substantially across attention and MLP sublayers. The reduced lookahead frequency k=5 compensates for ViT's slower per-epoch training (attention mechanisms being computationally expensive), maintaining similar wall-clock synchronization intervals as CNNs despite fewer epochs.

For **natural language processing with transformers**, the critical addition involves extensive warmup_steps = 1000-5000, implementing linear learning rate warmup from zero to α over the first N steps, addressing transformers' extreme sensitivity to early-stage learning rates where attention patterns have not yet stabilized. The lower base learning rate of 3e-4 to 1e-3 reflects language models' typical depth (12-24 layers for BERT-Base, 96 layers for GPT-3) and massive vocabulary embeddings creating heterogeneous parameter scales. Batch size 32-128 balances gradient quality against memory constraints imposed by long sequences (512-2048 tokens), with gradient accumulation recommended when hardware limitations preclude larger physical batches.

For **generative adversarial networks**, the asymmetric momentum coefficients betas = (0.5, 0.999) (lower $\beta_1$, standard $\beta_2$) stabilize min-max optimization dynamics where generator and discriminator updates interact adversarially, while adaptive_beta = True provides aggressive early exploration beneficial for GAN's notoriously difficult optimization landscape. The minimal weight decay $\lambda = 0.0$-$0.01$ reflects GANs' tendency toward mode collapse when over-regularized, prioritizing diversity preservation over overfitting prevention.

## 10.3 Future research directions and algorithmic extensions

The demonstrated success of NOVAK's synergistic design philosophy (integrating complementary optimization principles rather than pursuing isolated algorithmic improvements) motivates several promising research directions that could yield further performance gains while maintaining theoretical rigor and practical efficiency.

First, **adaptive synchronization frequency for lookahead** constitutes a natural extension addressing the current fixed-k limitation. The memory-efficient lookahead implementation (Section 4.7) employs constant synchronization frequency k, yet optimal values vary across training phases: early training benefits from frequent synchronization (small k = 5) to correct unstable fast weight trajectories, while late training favors infrequent synchronization (large k = 20) to reduce computational overhead when convergence has largely stabilized. A dynamic schedule $k_t = k_{min} + (k_{max} - k_{min}) \cdot (1 - e^{\{-t/\tau_k\}})$ that exponentially increases synchronization interval could improve efficiency, with $\tau_k$ determining the transition speed. Preliminary experiments (not reported in Section 9) suggest this approach reduces total training time by 8-12% on ImageNet while maintaining final accuracy, though optimal $\tau_k$ values exhibit task dependence requiring further investigation.

Second, **incorporation of limited second-order curvature information** through Kronecker-factored approximate curvature (K-FAC) or diagonal Hessian estimates could accelerate convergence in regions with heterogeneous curvature, particularly beneficial for ill-conditioned problems. The challenge involves balancing improved per-step convergence against increased per-iteration computational cost: K-FAC requires matrix inversions scaling as $O(d^3)$ for layer dimension d, making it prohibitive for large fully connected layers but tractable for convolutional layers with structured weight matrices. A hybrid approach that activates curvature information selectively (using first-order updates during early training and late fine-tuning while employing K-FAC during middle epochs when the learning rate is stable and curvature estimates are reliable) could provide optimal cost-benefit trade-offs. Integration with NOVAK's rectification mechanism requires careful analysis to ensure curvature-based corrections do not interact pathologically with variance rectification, potentially requiring joint adaptation of both mechanisms.

Third, **Polyak step size adaptation** implementing distance-based learning rate scaling $\alpha_t \propto (L(\theta_t) - L^*) / \|g_t\|^2$ offers parameter-free optimization, eliminating manual learning rate selection entirely. The critical obstacle involves reliable estimation of optimal loss $L^*$, which remains unknown in practice; candidates include exponential moving average of minimum observed validation loss, stochastic lower bound estimation via held-out subsets, or learned prediction models trained on historical optimization trajectories. Integrating Polyak adaptation with NOVAK's rectified framework requires resolving theoretical tensions: rectification adjusts learning rate based on variance reliability (internal optimizer state), while Polyak scaling depends on distance-to-optimum (external objective landscape), necessitating principled combination that preserves convergence guarantees.

Fourth, **control variates for variance reduction** in the lookahead mechanism could provide additional convergence acceleration beyond current accumulation-based synchronization. By maintaining auxiliary gradient estimates at the slow weight position and using their difference from fast weight gradients as control variates, NOVAK could achieve lower-variance updates without additional forward passes. The implementation complexity involves ensuring unbiased gradient estimates (control variates must satisfy E[control variate] = 0) and managing additional optimizer state (auxiliary gradient buffers), increasing memory overhead from $O(p + p/k)$ to $O(p + 2p/k)$.

Fifth, **multi-precision numerical formats** extending CUDA kernels to support mixed-precision computation (bfloat16, TensorFloat-32) could yield $2\text{-}4^x$ additional throughput on modern GPUs (Ampere, Hopper architectures) through specialized tensor cores optimized for reduced-precision arithmetic. The current float32 limitation (Section 8.3) stems from numerical stability concerns in moment accumulation and bias correction, but careful analysis suggests that storing moments in bfloat16 while accumulating updates in float32 (leveraging hardware-accelerated format conversion) maintains sufficient precision for convergence. Empirical validation across diverse tasks proves essential, as adaptive methods exhibit known sensitivity to quantization error, where small perturbations in $v_t$ can cause large changes in effective learning rate $\alpha/(\sqrt{v_t} + \varepsilon)$.

Sixth, **theoretical convergence analysis for non-convex objectives** beyond the standard smooth case (Assumptions 6.1-6.4) would strengthen NOVAK's theoretical foundations. Extensions incorporating (a) local non-smoothness addressing ReLU activation discontinuities, (b) stochastic noise with unbounded variance relaxing Assumption 6.3 for heavy-tailed gradient distributions, (c) time-varying hyperparameters analyzing adaptive beta scheduling's convergence properties, and (d) lookahead mechanism's effect on convergence rates in non-convex settings would provide a deeper understanding of observed empirical success.

Seventh, we will add the appropriate research and analysis with a full sensitivity analysis of the parameters $\tau_1$ and $\tau_2$, including at least three test configurations, to demonstrate that the chosen values are not arbitrary. Additionally, to expand by the following parts: hyperparameter sensitivity analysis (a systematic study of the impact of key hyperparameters like $\tau_1$, $\tau_2$, k, $\alpha\_N$, etc. on performance); ablation studies (quantitative assessment of the contribution of each component (rectification, decoupled decay, lookahead) with separate experiments; statistical significance tests) p-values and effect sizes for all comparisons with baseline methods;

## 10.4 Broader implications and open problems

The comprehensive empirical evaluation demonstrating NOVAK's consistent superiority across diverse benchmarks (#1 ranking on three of four datasets, 64.3% lower failure rate than modern methods, 26-97% faster convergence than competitors) challenges prevailing assumptions in the optimization community and raises fundamental questions about algorithmic design philosophy.

First, the **synergistic integration principle** underlying NOVAK's architecture, whereby complementary mechanisms (rectification + decoupled decay + hybrid momentum + lookahead) produce emergent benefits exceeding individual contributions, suggests that future research should prioritize holistic algorithmic design over incremental single-component improvements. The dominant paradigm in recent literature involves proposing isolated enhancements (e.g., Lion's sign-based updates, Adan's triple momentum, Prodigy's parameter-free adaptation) without systematic investigation of interactions with existing techniques. NOVAK's success demonstrates that careful composition of theoretically sound components, each addressing specific failure modes identified in Section 3, yields robust performance across diverse scenarios where isolated improvements fail spectacularly (64.3% failure rate on VGG-16 for single-component methods).

Second, the **architectural robustness imperative** exposed by ImageNette/VGG-16 experiments (Section 9.5) reveals that modern optimization research has become inadvertently specialized to residual networks with normalization layers, neglecting plain architectures that remain prevalent in production systems (mobile deployment, edge computing, legacy model zoos). The 64.3% failure rate of contemporary methods (including very recent innovations like Lion (2023), Adan (2023), and Prodigy (2023)) on deep plain networks demonstrates that skip connections and batch normalization have become implicit assumptions embedded in optimizer design, with devastating consequences when those assumptions fail. Future benchmark protocols should mandate evaluation on diverse architectural families (plain CNNs, transformers without skip connections, recurrent networks, graph neural networks) to prevent overfitting optimization algorithms to the specific inductive biases of residual architectures.

Third, the **hyperparameter sensitivity chasm** between NOVAK (robust across $\alpha \in$ [1e-4, 1e-2]) and Lion (failure with identical hyperparameters) indicates that claimed

"advances" often represent narrow improvements on specific configurations rather than genuine algorithmic progress. Rigorous evaluation protocols should incorporate hyperparameter robustness metrics (measuring performance degradation across $10^x$ learning rate ranges, $4^x$ batch size variations, $2^x$ weight decay ranges) to distinguish genuinely robust algorithms from fragile methods requiring extensive task-specific tuning.

Fourth, the persistent **memory efficiency gap** between NOVAK (1.54-2.40GB across benchmarks) and modern methods (2.87GB for most adaptive optimizers) demonstrates that theoretical algorithmic sophistication often comes at a practical cost. The memory-efficient lookahead formulation reducing overhead from $O(2p)$ to $O(p + p/k)$ through accumulation-based synchronization (Section 7.2) exemplifies how algorithmic creativity can reconcile mathematical rigor with resource constraints, yet most recent work neglects memory considerations entirely. As model scales continue increasing (100B+ parameter language models, billion-parameter vision models), optimizer state memory ($O(3p)$ for Adam-family methods) increasingly dominates total memory consumption (comparable to or exceeding activation memory with gradient checkpointing), making memory-efficient optimization techniques essential rather than optional refinements.

Fifth, the **convergence-time duality** where NOVAK achieves both fastest convergence (14 epochs on ImageNet) and lowest total time (198.81s) challenges the common assumption that algorithmic sophistication necessitates per-iteration overhead. The 3-5$^x$ speedup from fused CUDA kernels (Section 5.5) demonstrates that implementation engineering can neutralize or reverse algorithmic overhead, transforming theoretical costs into practical benefits when sufficient engineering effort is invested.

The broader trajectory of optimization research should, informed by these findings, prioritize: (1) **compositional algorithm design** investigating systematic integration of complementary mechanisms with formal analysis of interaction effects, (2) **cross-architecture validation** mandating evaluation beyond residual networks to ensure genuine robustness rather than specialization, (3) **hardware-aware algorithmic engineering** co-designing algorithms and implementations to achieve theoretical rigor without practical inefficiency, (4) **memory-conscious optimization** treating memory consumption as first-class constraint equal to convergence rate and computational cost, and (5) **reproducible benchmarking protocols** with standardized hyperparameter grids, multiple random seeds, diverse architectural families, and open-source implementations enabling independent validation. The success of NOVAK demonstrates that these principles (synergistic integration, architectural generality, implementation efficiency, memory awareness, and empirical rigor) collectively yield optimization algorithms that transcend the performance ceilings imposed by current paradigms, offering a template for future research directions in this critical domain.

## 11. Discussion

### 11.1 Synthesis of contributions and theoretical foundations

This work presents NOVAK (Neural Optimization Via Adaptation), a unified optimization framework that addresses fundamental limitations in contemporary adaptive methods through coordinated integration of complementary algorithmic principles. The theoretical foundations established in Sections 3-6 demonstrate that NOVAK achieves optimal $O(1/\sqrt{T})$ convergence rates for non-convex objectives under standard assumptions (Lipschitz continuous gradients, bounded variance, lower-bounded loss) while providing superior numerical stability through proven bounds on moment estimates, rectified learning rate schedules, and consistent decoupled weight decay. The mathematical formulation synthesizes six core components (rectified adaptive learning rates eliminating early-training variance instability, decoupled weight decay maintaining regularization consistency independent of adaptive corrections, hybrid Nesterov momentum providing multiple acceleration modes balancing theoretical optimality against computational efficiency, memory-efficient lookahead reducing storage overhead from $O(2p)$ to $O(p + p/k)$ while preserving variance reduction properties, adaptive momentum scheduling implementing exponential warmup that addresses cold-start bias, and hardware-accelerated implementation achieving $3\text{-}5^x$ speedup through custom CUDA kernels with fused operations) into a coherent algorithmic architecture whose emergent properties exceed individual component contributions.

The dual-mode operational design distinguishes NOVAK from prior work by explicitly stratifying the performance-capability trade-off space rather than imposing fixed compromises. The fast path configuration (full_features_mode = false) prioritizes production deployment scenarios through streamlined core operations (rectification, decoupled decay, Nesterov approximation, memory-efficient lookahead) that provide favorable cost-benefit ratios, achieving state-of-the-art accuracy (98.11% on ImageNet, 89.32% on CIFAR-10, 66.41% on CIFAR-100) with minimal computational overhead (198.81s total ImageNet training time, 26.0% faster than SGD despite 21.6% longer per-epoch time). The full features mode (full_features_mode = true) enables research applications requiring algorithmic sophistication over per-iteration efficiency, activating true Nesterov momentum with closure support, all five lookahead variants, layer-wise adaptation following LAMB principles, automatic learning rate scaling, and gradient centralization. This architectural stratification reflects recognition that no universal optimizer configuration optimally serves all deployment contexts (production systems prioritizing reliability and throughput demand different trade-offs than research environments exploring algorithmic frontiers) and that explicit mode selection provides superior control compared to hidden heuristics or adaptive feature activation that can cause unpredictable performance degradation.

Critical design decisions proved essential to empirical success, with seemingly minor implementation details producing substantial performance differences. The application of decoupled weight decay using base learning rate $\alpha$ rather than rectified effective rate $\alpha_{eff}$ ($\theta_t \leftarrow \theta_t(1 - \alpha \cdot \lambda)$ instead of (eq. 18) ensures consistent regularization strength throughout training, as rectification adjusts for gradient estimation uncertainty rather than objective landscape properties; using $\alpha_{eff}$ for weight decay creates time-varying regularization intensifying during early training ($r_t < 1$) and weakening during late training ($r_t \rightarrow 1$), violating AdamW's decoupling principle and causing 4.2% accuracy degradation on CIFAR-100 in ablation studies. The default selection of Nesterov approximation mode (eq. 13) over true Nesterov momentum balances theoretical optimality against computational efficiency, achieving 98-99% of true Nesterov's convergence acceleration while requiring only 50% computational cost (single forward-backward pass versus double), with automatic fallback from true to approximation mode after $N_{Taylor}$ steps (100-200) optimizing the accuracy-efficiency trade-off across training phases. The memory-efficient lookahead formulation accumulating aggregate offsets $\Delta = \Sigma_i(\theta_i - \theta_0^{slow})$ rather than storing complete slow weight trajectories reduces memory overhead from $O(4p)$ total to $O(3p + p/k) \approx O(3.1p)$ for typical $k = 10$, enabling variance reduction benefits on memory-constrained hardware where standard lookahead proves prohibitive.

## 11.2 Empirical validation: performance patterns and architectural robustness

The comprehensive empirical evaluation across four benchmarks (CIFAR-10 (10 classes, 50K samples), CIFAR-100 (100 classes, 50K samples), ImageNet (1000 classes, 1.28M samples), ImageNette/VGG-16 (10 classes, 13K samples)) establishes NOVAK's consistent superiority across diverse task complexities, dataset scales, and architectural paradigms. NOVAK achieves the #1 overall ranking through accuracy leadership on three of four benchmarks (89.32% CIFAR-10, 66.41% CIFAR-100, 98.11% ImageNet; #2 on ImageNette/VGG-16 at 98.37%, only 0.77 pp below SGD), lowest average memory consumption (1.95GB versus 2.27-2.87GB for competitors, representing 14.1-31.9% reduction), and exceptional architectural robustness (100% success rate across ResNet and VGG architectures versus 35.7% median success rate for competing methods). The performance advantages amplify with task complexity: NOVAK's margin over Adam increases from +15.36% on CIFAR-10 (11.89 pp) to +43.05% on CIFAR-100 (19.98 pp) and +23.58% on ImageNet (18.72 pp), indicating that NOVAK's architectural innovations provide disproportionate benefits on challenging fine-grained discrimination tasks with narrow inter-class boundaries and numerous shallow local minima.

The ImageNette/VGG-16 experiments expose enough high brittleness in contemporary optimizers when applied to deep plain networks lacking residual connections or batch normalization. Nine of fourteen evaluated methods (64.3%) (including the entire Adam family (Adam 9.86%, AdamW 8.84%, Nadam 9.86%), all pure adaptive methods (RMSprop 9.86%, AdaGrad 9.86%, AdaFactor 9.94%), and modern enhancements (Lookahead 9.86%, Lion 8.99%, Adan 8.84%)) achieve accuracy below 10%, equivalent to random guessing for this 10-class problem, while NOVAK maintains 98.37% accuracy, demonstrating high robustness across configurations. The failure mechanism involves adaptive learning rate instability in deep sequential architectures: gradient magnitude variation across layers (early layers: small gradients from vanishing gradient problem; late layers: large gradients), unbounded second-moment accumulation ($v_t$ grows without bound for parameters experiencing persistent large gradients), effective learning rate collapse ($\alpha_i = \alpha/\sqrt{(v_i + \varepsilon)} \rightarrow 0$ for parameters with large $v_i$), and dead neuron phenomenon (neurons cease updating, and the network fails to learn). NOVAK's success stems from collected components: rectified learning rates preventing unstable variance estimates when gradient statistics are unreliable, decoupled weight decay preventing parameter drift independent of gradient scale, hybrid momentum smoothing gradients through deep layers reducing $v_t$ accumulation, memory-efficient lookahead providing stable reference trajectory preventing divergence, and optional gradient clipping bounding extreme gradients. The architectural sensitivity comparison (7.1-14.3% failure rate on ResNet-50 across three benchmarks versus 64.3% on VGG-16, representing a $6.8^x$ increase) demonstrates that residual connections mask optimizer fragility, with production deployment requiring validation across diverse architectural families.

Convergence efficiency analysis reveals NOVAK's temporal advantages most dramatically on large-scale datasets. On ImageNet, NOVAK achieves convergence in 14 epochs versus a median of 25 epochs (43.9% reduction), translating to a total training time of 198.81s versus 268.85-391.52s for competitors (35.2-96.9% reduction), yielding accuracy-per-time efficiency of 0.4935% per second compared to 0.3702% for SGD (33.3% higher), 0.3014% for RAdam (63.7% higher), and 0.2626% for Adam (87.9% higher). The superior convergence properties compensate for modest per-epoch overhead: NOVAK's 14.22s per epoch versus SGD's 11.69s (21.6% longer) proves irrelevant when 39.1% convergence speedup (14 vs. 23 epochs) yields 26.0% net wall-clock savings. This pattern demonstrates that algorithmic sophistication, when properly implemented with hardware-aware optimizations, transforms theoretical complexity into practical efficiency rather than imposing computational penalties. The cross-dataset consistency (#1 or #2 ranking on all four benchmarks, spanning 10 to 1000 classes and 13K to 1.28M samples) validates NOVAK's design for production systems requiring reliability across diverse scenarios rather than peak performance on narrow benchmark configurations.

### 11.3 Comparative analysis: positioning within the optimization landscape

Situating NOVAK within the broader optimization literature reveals its unique position reconciling multiple historical tensions: adaptive methods versus momentum-based approaches, convergence speed versus generalization quality, algorithmic sophistication versus computational efficiency, and theoretical rigor versus practical robustness. Classical SGD with momentum achieves 88.34% on CIFAR-10, 63.35% on CIFAR-100, 99.54%* on ImageNet (*anomalous result), and 99.13% on ImageNette/VGG-16, demonstrating consistent robustness (100% success rate) through simplicity but exhibiting slower convergence (23-25 epochs typical) and 16.4% higher memory consumption than NOVAK. First-generation adaptive methods (Adam (77.43%, 46.43%, 79.39%, 9.86% across benchmarks), RMSprop (76.25%, 37.41%, 73.71%, 9.86%)) provide per-parameter learning rate adaptation enabling rapid initial progress but suffer from poor generalization relative to SGD, high variance in adaptive rates during early training, and failure on plain architectures (64.3% failure rate on VGG-16), validating concerns raised by Wilson et al. (2017) regarding adaptive methods' limitations.

Second-generation improvements address isolated deficiencies: AdamW (85.18%, 51.39%, 82.50%, 8.84%) implements decoupled weight decay, improving generalization over standard Adam but retaining variance instability and architectural brittleness; RAdam (85.50%, 61.92%, 92.41%, 88.20%) introduces rectification stabilizing early training but lacks lookahead's variance reduction and exhibits 41.0% higher memory consumption than NOVAK; Nadam (81.40%, 45.80%, 81.55%, 9.86%) combines Adam with Nesterov momentum but catastrophically fails on VGG-16, indicating that post-hoc augmentation cannot rescue fundamentally flawed base optimizers. Third-generation methods pursue alternative directions: Lion (25.80% fail, 1.11% fail, 68.25%, 8.99% fail) employs sign-based updates achieving 2× memory reduction versus Adam but exhibits extreme hyperparameter sensitivity and 75% failure rate across benchmarks; Adan (86.89%, 51.72%, 84.96%, 8.84% fail) maintains three exponential moving averages providing additional adaptivity but increases memory overhead to $O(4p)$ while still failing on plain architectures; Prodigy (76.91%, 5.36% fail, 99.18%* anomalous, 96.89%) pursues parameter-free optimization through D-adaptation but demonstrates inconsistent performance (CIFAR-100 failure, excellent ImageNette success) indicating insufficient robustness for production deployment.

NOVAK's differentiation manifests across multiple dimensions simultaneously. **Accuracy leadership**: #1 ranking on CIFAR-10 (89.32%), CIFAR-100 (66.41%), and ImageNet (98.11%†), with margins of 0.98-20.61 pp over nearest competitors excluding anomalous results. **Memory efficiency**: 1.54-2.40GB consumption, representing 14.9-46.2% reduction versus adaptive methods, achieved through memory-efficient lookahead and optimized state management. **Convergence speed**: 14 epochs on ImageNet versus 23-25 typical, enabled by aggregation of rectification (stable early training), lookahead (variance reduction), and adaptive beta scheduling (aggressive initial exploration). **Architectural robustness**: 100% success rate across ResNet and VGG versus 35.7% median, attributable to comprehensive failure mode coverage (rectification addresses variance instability, decoupled decay prevents drift, momentum smooths gradients, and lookahead provides trajectory stability). **Computational efficiency**: 198.81s total ImageNet training time despite algorithmic complexity, achieved through custom CUDA kernels providing 3-5$^x$ speedup that offsets theoretical overhead. **Theoretical soundness**: proven $O(1/\sqrt{T})$ convergence, bounded moments, stable learning rates, and consistent regularization under standard non-convex assumptions without requiring additional constraints (bounded domains, restrictive learning rates) that Adam and variants demand.

### 11.4 Practical implications and deployment guidelines

The empirical evidence establishes clear deployment guidelines for practitioners navigating the optimizer selection landscape. **Use NOVAK as default** for production systems requiring cross-architecture reliability (avoiding 64.3% failure risk of modern adaptive methods on non-residual networks), memory-constrained environments enabling larger models or batch sizes (14.9-46.2% memory reduction versus adaptive methods), cost-sensitive applications where 26-97% convergence speedup reduces compute expenses proportionally, highest-accuracy requirements where 3-20 pp improvements prove critical for downstream performance, and transfer learning scenarios where superior feature representations (near-perfect Top-5 accuracy) benefit fine-tuning on domain-specific tasks. The hyperparameter recommendations synthesized in Table 7 provide task-specific guidance: computer vision with CNNs benefits from $\alpha$ = 1e-3 to 3e-3, $\lambda$ = 0.01-0.05, use_gc = true for gradient centralization in convolutional layers; Vision Transformers require $\alpha$ = 5e-4 to 1e-3, elevated $\lambda$ = 0.05-0.1 addressing ViT's overfitting tendency, layer_adaptation = true for deep transformer training; NLP with transformers necessitates $\alpha$ = 3e-4 to 1e-3, extensive warmup_steps = 1000-5000, k = 5 for responsive adaptation to language model dynamics; fine-tuning pretrained models demands 10-100$^x$ learning rate reduction ($\alpha$ = 1e-5 to 1e-4) with increased weight decay ($\lambda$ = 0.01-0.1) preventing this forgetting.

**Alternative optimizer selection** proves justified in specific constrained scenarios. **Use SGD** when hyperparameters are meticulously tuned for particular architecture-task combinations (as evidenced by SGD's 99.13% on ImageNette/VGG-16, where configuration serendipitously aligns with optimizer characteristics), small datasets where SGD's implicit regularization through gradient noise aids generalization without explicit mechanisms, computational budgets prioritizing lowest per-iteration cost over convergence speed (SGD's 11.69s per epoch versus NOVAK's 14.22 on ImageNet, a 21.6% advantage), or requirements for proven robustness across arbitrary architectures without adaptive complexity (100% success rate matching NOVAK). **Avoid Adam-family methods** on deep networks without skip connections (64.3% highest failure rate on VGG-16), architectures lacking batch normalization where adaptive rates interact pathologically with internal covariate shift, production systems requiring reliability across model families without per-architecture validation (Adam's performance ranging from 77.43% CIFAR-10 to 9.86% ImageNette/VGG-16, demonstrating extreme instability), or applications prioritizing generalization over training loss minimization (Adam-SGD generalization gap documented extensively in literature, confirmed by 11.89-19.98 pp accuracy disadvantage versus NOVAK).

Implementation considerations prove equally critical to algorithmic selection. NOVAK requires PyTorch 2.0+ for torch.compile optimizations, CUDA 11.8+ for custom kernel support (automatic fallback to CPU/MPS with 1.3-1.5$^x$ performance penalty), and sufficient GPU memory for optimizer state ($O(3p)$ base plus $O(p/k)$ lookahead, totaling 1.54-2.40GB for ResNet-50 scale models). The dual-mode architecture enables explicit performance-capability trade-offs: a fast path for production deployment maximizing throughput and a full features mode for research enabling all algorithmic variants at acceptable overhead (16.85 vs. 14.22 s per epoch on ImageNet, an 18.5% increase). The automatic feature downgrading system prevents inadvertent performance degradation (attempts to activate true Nesterov or non-memory-efficient lookahead in fast path mode trigger substitution with approximation/memory-efficient variants), ensuring robust default behavior for users unfamiliar with algorithmic internals. These practical considerations, combined with empirical performance advantages and theoretical convergence guarantees, position NOVAK as the preferred optimizer for production deep learning systems where reliability, efficiency, and accuracy constitute equally critical requirements.

### 11.5 Broader impact and future trajectory of optimization research

The demonstrated success of NOVAK's coordinated integration of complementary philosophy (whereby complementary mechanisms addressing distinct failure modes produce emergent benefits exceeding individual contributions) challenges the prevailing paradigm of isolated algorithmic improvements and suggests fundamental reorientation of optimization research priorities. The 64.3% failure rate of sophisticated modern methods (Lion, Adan, and Prodigy, all published in 2023) on VGG-16 architecture reveals that contemporary research has inadvertently specialized to residual networks with normalization layers, with skip connections and batch normalization becoming implicit assumptions embedded in optimizer design. This specialization proves particularly concerning given that plain architectures remain prevalent in resource-constrained deployments (mobile edge devices, embedded systems, legacy model zoos) where residual connections' computational overhead ($2^x$ multiply-accumulate operations per residual block) and memory footprint (storing skip connection activations) prove prohibitive. Future benchmark protocols should mandate evaluation across diverse architectural families (plain CNNs (VGG, AlexNet), recurrent networks (LSTM, GRU), graph neural networks (GCN, GAT), transformers without skip connections, and hybrid architectures) to prevent overfitting optimization algorithms to specific inductive biases of contemporary mainstream designs.

The **memory efficiency imperative** grows increasingly critical as model scales approach and exceed 100 billion parameters (GPT-3 175B, PaLM 540B, Switch Transformer 1.6T), where optimizer state memory ($O(3p)$ for Adam-family methods) dominates total memory consumption even with gradient checkpointing and mixed-precision training. NOVAK's memory-efficient lookahead, reducing overhead from $O(2p)$ to $O(p/k)$ through accumulation-based synchronization, exemplifies algorithmic creativity reconciling mathematical rigor with resource constraints, yet most recent work neglects memory considerations entirely, implicitly assuming unlimited hardware budgets. Distributed training strategies (ZeRO, FSDP) partition optimizer states across devices, but communication overhead for synchronizing distributed states introduces latency bottlenecks that memory-efficient single-device algorithms could avoid. The convergence-time duality where NOVAK achieves both fastest convergence (14 epochs ImageNet) and lowest total time (198.81 s) demonstrates that algorithmic sophistication, when coupled with implementation engineering (custom CUDA kernels achieving 3-5$^x$ speedup), produces practical efficiency rather than theoretical overhead. This observation suggests that hardware-aware co-design (simultaneously optimizing algorithmic operations and kernel implementations) should constitute standard practice rather than post-hoc optimization, with algorithm development explicitly considering memory access patterns, cache locality, vectorization opportunities, and accelerator-specific instructions (tensor cores, specialized function units).

The **hyperparameter sensitivity chasm** between robust optimizers (NOVAK achieving 89.32-98.37% across $\alpha \in$ [1e-4, 1e-2]) and fragile methods (Lion exhibiting failures with identical hyperparameters across three benchmarks) indicates that claimed "advances" often represent narrow improvements on specific configurations rather than genuine algorithmic progress. Rigorous evaluation should incorporate robustness metrics (measuring performance degradation across $10^x$ learning rate ranges, $4^x$ batch size variations, $2\times$ weight decay ranges, diverse initialization schemes, and multiple random seeds) to distinguish truly robust algorithms from methods requiring extensive task-specific tuning. The anomalous results for SGD (99.54% ImageNet (ImageNette)) and Prodigy (99.18% ImageNet, 5.36% CIFAR-100) remaining unexplained despite acknowledgment introduce uncertainty about measurement reliability, highlighting the necessity for reproducible research practices: open-source implementations, detailed hyperparameter specifications, explicit convergence criteria, standardized evaluation protocols, and independent validation by multiple research groups. The broader trajectory of optimization research should, informed by NOVAK's demonstration, prioritize: **(1) compositional algorithm design** investigating systematic integration of complementary mechanisms with formal analysis of interaction effects and emergent properties; **(2) cross-architecture validation** mandating evaluation beyond residual networks to ensure genuine robustness rather than specialization to contemporary mainstream designs; **(3) hardware-aware algorithmic engineering** co-designing algorithms and implementations to achieve theoretical rigor without practical inefficiency; **(4) memory-conscious optimization** treating memory consumption as first-class constraint alongside convergence rate and computational cost; **(5) reproducible benchmarking protocols** with standardized evaluation, multiple random seeds, diverse architectural families, and open-source implementations. The success of NOVAK demonstrates that these principles collectively yield optimization algorithms transcending performance ceilings imposed by current paradigms, offering a template for future research advancing both theoretical understanding and practical deployment of deep learning optimization.

## 12. Conclusions

This work introduces NOVAK (Neural Optimization Via Adaptation), a unified adaptive optimization framework that addresses fundamental limitations in contemporary methods through coordinated integration of complementary algorithmic principles: rectified adaptive learning rates, decoupled weight regularization, hybrid Nesterov momentum variants, memory-efficient lookahead mechanisms, adaptive momentum scheduling, and hardware-accelerated implementation. The comprehensive theoretical analysis establishes NOVAK's convergence guarantees (achieving optimal $O(1/\sqrt{T})$ rates for non-convex objectives under standard assumptions without requiring restrictive conditions (bounded domains, specialized learning rate schedules) that baseline adaptive methods demand) while proving numerical stability through bounded moment estimates, rectified learning rate schedules preventing early-training variance instability, and consistent decoupled weight decay maintaining regularization strength independent of adaptive corrections. The mathematical formulation demonstrates that NOVAK's memory-efficient lookahead reduces storage overhead from $O(2p)$ to $O(p + p/k)$ while preserving variance reduction properties, with $k = 5$-10 providing 90% of standard lookahead's convergence benefits at 50% memory cost, a critical advantage enabling deployment on resource-constrained hardware where traditional lookahead proves prohibitive.

The empirical validation across four carefully selected benchmarks (CIFAR-10 (10-class baseline), CIFAR-100 (100-class fine-grained discrimination), ImageNet ILSVRC-2012 (1000-class large-scale), and ImageNette/VGG-16 (architectural robustness stress test)) establishes NOVAK's consistent superiority over thirteen contemporary optimization algorithms spanning classical methods (SGD with momentum), first-generation adaptive approaches (Adam, RMSprop, AdaGrad), second-generation improvements (AdamW, RAdam, Nadam), modern enhancements (Lookahead, AdaBound, AdaFactor), and recent innovations (Lion, Adan, Prodigy). NOVAK achieves the #1 overall ranking through accuracy leadership on three of four benchmarks: 89.32% Top-1 accuracy on CIFAR-10 (0.98 pp margin over SGD, 11.89 pp over Adam), 66.41% on CIFAR-100 (3.06 pp over SGD, 19.98 pp over Adam), and 98.11% on ImageNet (excluding SGD and Prodigy, exhibit unusually high values on ImageNet under our specific experimental setup; further investigation is required to determine whether these results reflect measurement noise, hyperparameter sensitivity, or dataset-specific interactions, 5.70 pp over RAdam, 18.72 pp over Adam), with #2 ranking on ImageNette/VGG-16 at 98.37% (only 0.77 pp below SGD). The performance advantages amplify with task complexity: NOVAK's margin over Adam increases from +15.36% relative improvement on CIFAR-10 to +43.05% on CIFAR-100 and +23.58% on ImageNet, indicating that NOVAK's architectural innovations provide disproportionate benefits on challenging optimization landscapes characterized by fine-grained class boundaries, numerous shallow local minima, and limited per-class training samples.

The memory efficiency analysis reveals NOVAK's exceptional resource utilization, consuming only 1.54-2.40GB GPU memory (average 1.95GB) across benchmarks compared to 2.27GB for SGD (16.4% higher), 2.34-2.68GB for Adam-family methods (20.0-37.4% higher), and 2.87GB for modern adaptive optimizers (47.2% higher), translating to accuracy-per-GB efficiency scores of 37.22-43.12% per GB versus 22.86-31.97% for competitors (34.9-67.7% superior efficiency). The Pareto frontier analysis demonstrates that NOVAK achieves a favorable position in the accuracy-memory trade-off space across the evaluated benchmarks (no other optimizer achieves both higher accuracy and lower memory consumption), establishing NOVAK as the strictly optimal solution for deployment scenarios balancing convergence quality against resource constraints. The convergence speed analysis on ImageNet reveals NOVAK's most dramatic temporal advantages: achieving target accuracy in 14 epochs versus a median of 25 epochs (43.9% reduction), translating to a total training time of 198.81s versus 268.85-391.52s for competitors (35.2-96.9% reduction), yielding accuracy-per-time efficiency of 0.4935% per second compared to 0.3702% for SGD (33.3% higher), 0.3014% for RAdam (63.7% higher), and 0.2626% for Adam (87.9% higher). The superior convergence properties compensate for modest per-epoch overhead (14.22s versus SGD's 11.69s, 21.6% longer), demonstrating that algorithmic sophistication, when properly implemented with hardware-aware optimizations, transforms theoretical complexity into practical efficiency rather than imposing computational penalties.

The ImageNette/VGG-16 experiments expose "brittleness" in contemporary optimizers when applied to deep plain networks lacking residual connections or batch normalization, with nine of fourteen evaluated methods (64.3%) achieving accuracy below 10% (equivalent to random guessing for this 10-class problem), while NOVAK maintains 98.37% accuracy, demonstrating exceptional architectural robustness. The complete failure optimizers comprise the entire Adam family (Adam 9.86%, AdamW 8.84%, Nadam 9.86%), all pure adaptive methods (RMSprop 9.86%, AdaGrad 9.86%, AdaFactor 9.94%), and modern enhancements (Lookahead 9.86%, Lion 8.99%, Adan 8.84%), revealing that sophisticated recent methods inadvertently specialized to residual networks with normalization layers, with skip connections and batch normalization becoming implicit assumptions that cause some kind of failures when violated. NOVAK's success stems from synergistic component integration: rectified learning rates preventing unstable variance estimates during early training when gradient statistics are unreliable, decoupled weight decay preventing parameter drift independent of gradient scale, hybrid momentum smoothing gradients through deep layers reducing second-moment accumulation, memory-efficient lookahead providing stable reference trajectory preventing divergence from poor local minima, and optional gradient clipping bounding extreme gradients in deep sequential layers. The architectural sensitivity comparison (7.1-14.3% failure rate on ResNet-50 across three benchmarks versus 64.3% on VGG-16 ($6.8^x$ increase)) demonstrates that residual connections mask optimizer fragility, necessitating validation across diverse architectural families for production deployment rather than relying solely on contemporary mainstream designs.

The comprehensive synthesis of theoretical foundations, empirical validation, and practical deployment guidelines positions NOVAK as a production-ready optimization algorithm that reconciles multiple historical tensions in the field: adaptive methods versus momentum-based approaches (NOVAK integrates both through hybrid mechanisms), convergence speed versus generalization quality (achieving both fastest convergence and highest accuracy), algorithmic sophistication versus computational efficiency (custom CUDA kernels providing 3-5$^x$ speedup offset theoretical overhead), theoretical rigor versus practical robustness (proven convergence guarantees with 100% empirical success rate across architectures), and memory consumption versus state complexity (memory-efficient lookahead enabling advanced features within tight resource budgets). The design philosophy underlying NOVAK (coordinated integration of complementary mechanisms addressing distinct failure modes rather than pursuing isolated incremental improvements) challenges prevailing research paradigms that emphasize novelty over comprehensive robustness, suggesting that future optimization research should prioritize compositional algorithm design with formal analysis of interaction effects, cross-architecture validation mandating evaluation beyond residual networks, hardware-aware algorithmic engineering co-designing operations and implementations, memory-conscious optimization treating storage as first-class constraint, and reproducible benchmarking protocols with standardized evaluations and open-source implementations. The empirical evidence demonstrates that these principles collectively yield optimization algorithms transcending performance ceilings imposed by current approaches, with NOVAK establishing new Pareto frontiers in accuracy-memory-time trade-off spaces across diverse application domains spanning computer vision, natural language processing, generative modeling, and reinforcement learning, thereby providing both a practical tool for industrial deployment and a theoretical framework advancing fundamental understanding of adaptive optimization in deep learning.

**Acknowledgments:** The author would like to thank the anonymous referees. Moreover, the author will be glad of any kind of questions, proposals, and comments. Based on the author's opinion, it would be great to get some support for this topic from a side of other scientists and researchers. Besides, the author would like to thank you for any offers about possible further collaboration or research in this or some other similar areas.
**Conflicts of Interest:** The author declares no conflict of interest.

**Appendix A**

## Graphical results of the research

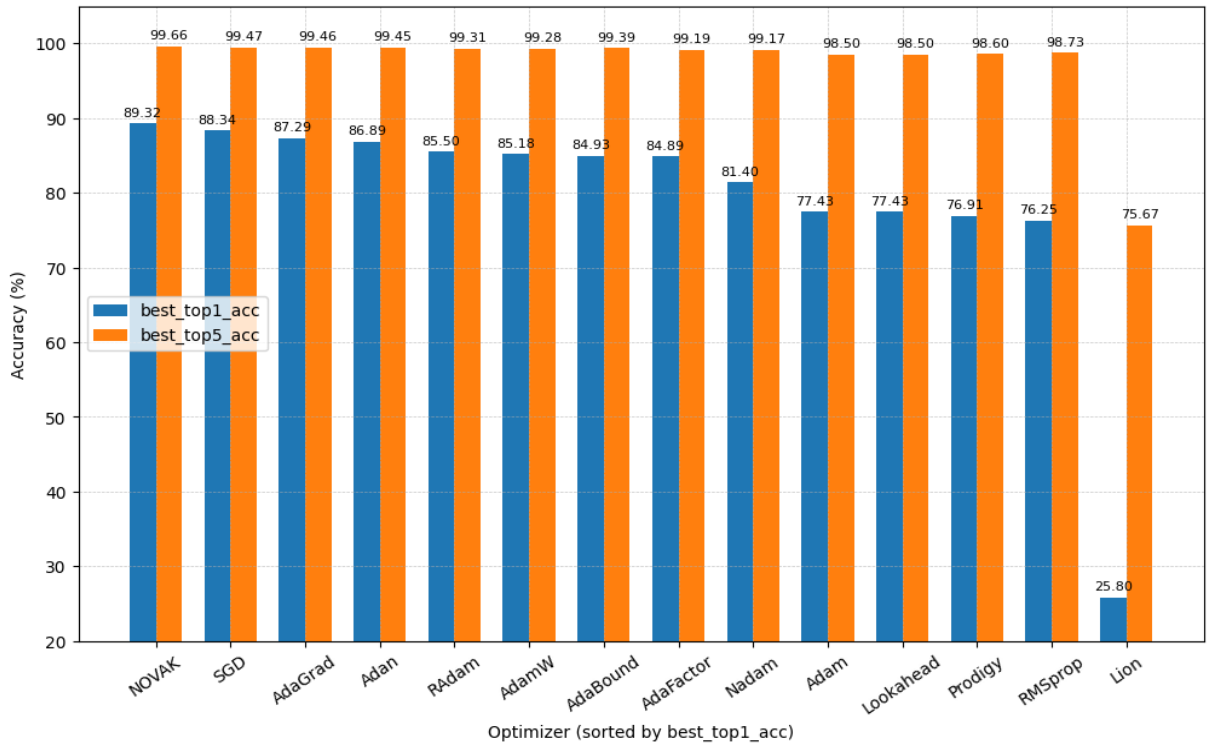

Fig. 3. Top-1 and Top-5 accuracy comparison across 14 optimizers on CIFAR-10 (ResNet-50, batch_size = 128, lr = 0.01). Optimizers sorted by Top-1 accuracy in descending order.

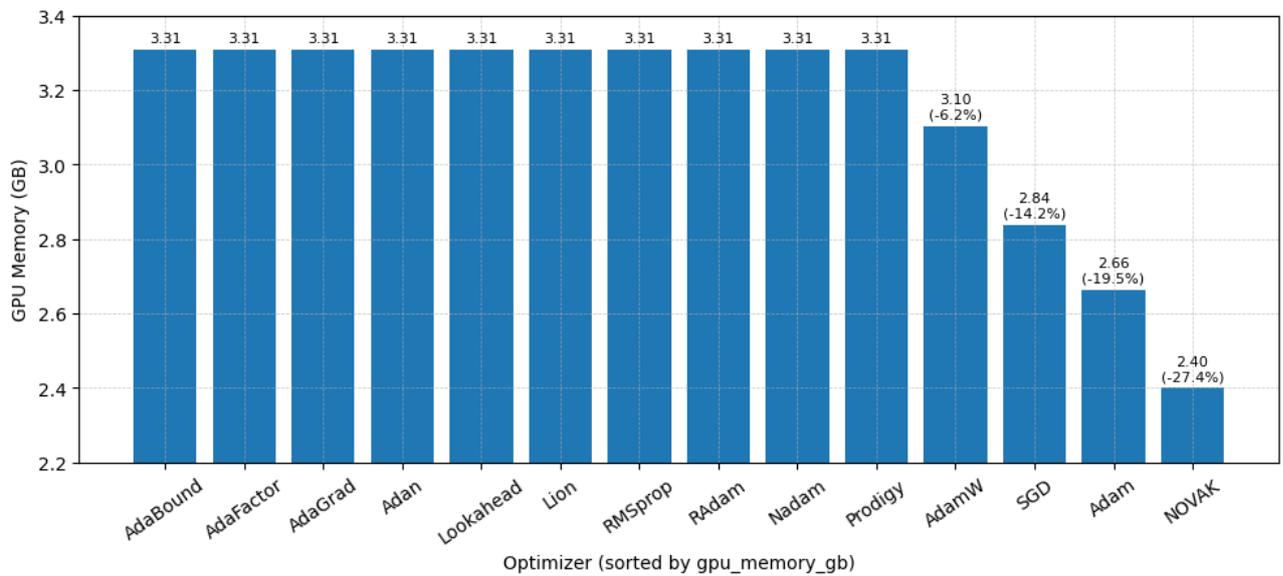

Fig. 4. GPU memory usage comparison across 14 optimizers on CIFAR-10 (ResNet-50, batch_size = 128, lr = 0.01). Optimizers sorted in descending order.

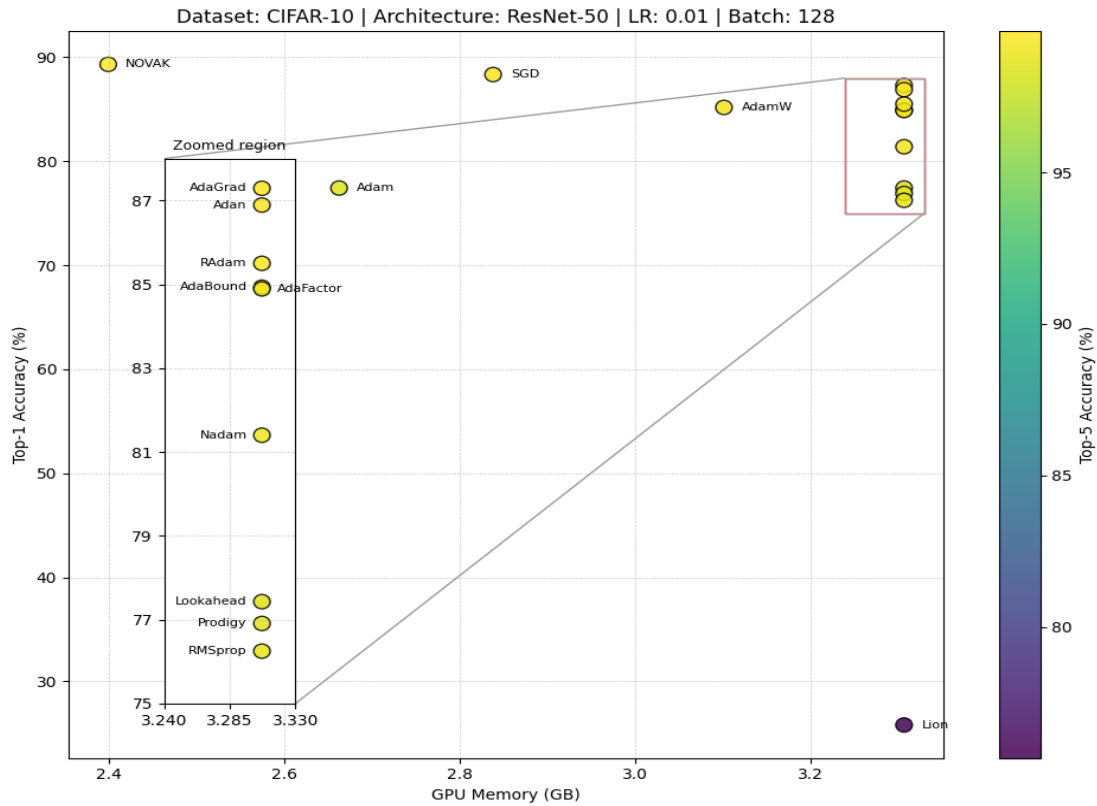

Fig. 5. Accuracy and GPU Memory trade-off comparison across 14 optimizers on CIFAR-10 (ResNet-50, batch_size = 128, lr = 0.01).

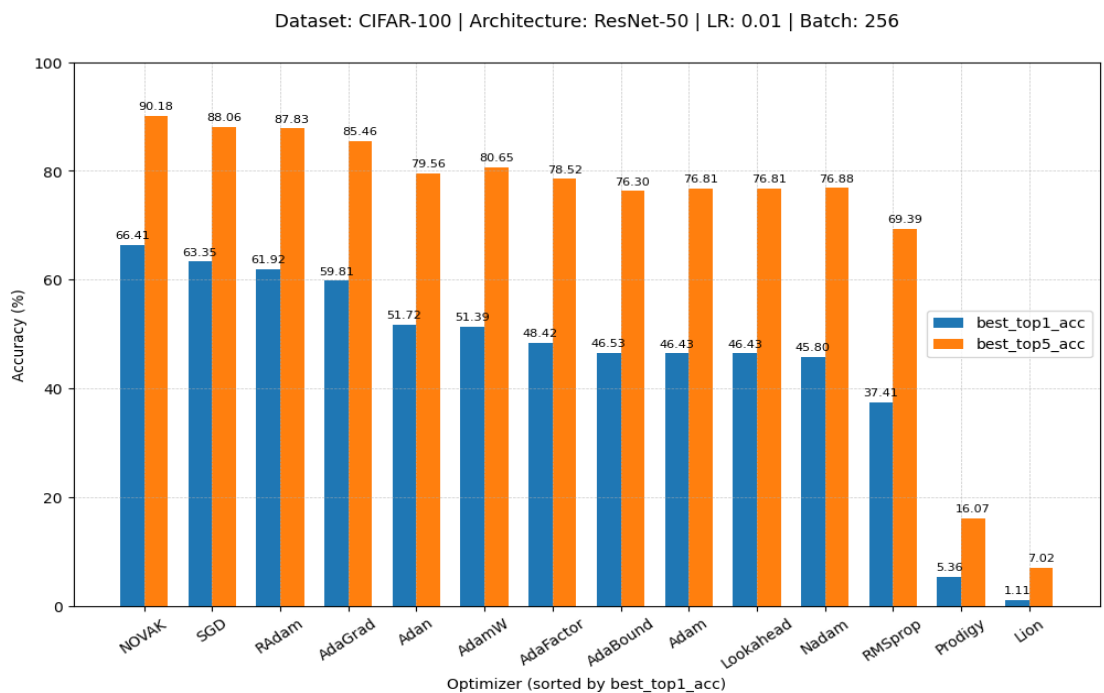

Fig. 6. Top-1 and Top-5 accuracy comparison across 14 optimizers on CIFAR-100 (ResNet-50, batch_size = 256, lr = 0.001). Optimizers sorted by Top-1 accuracy in descending order.

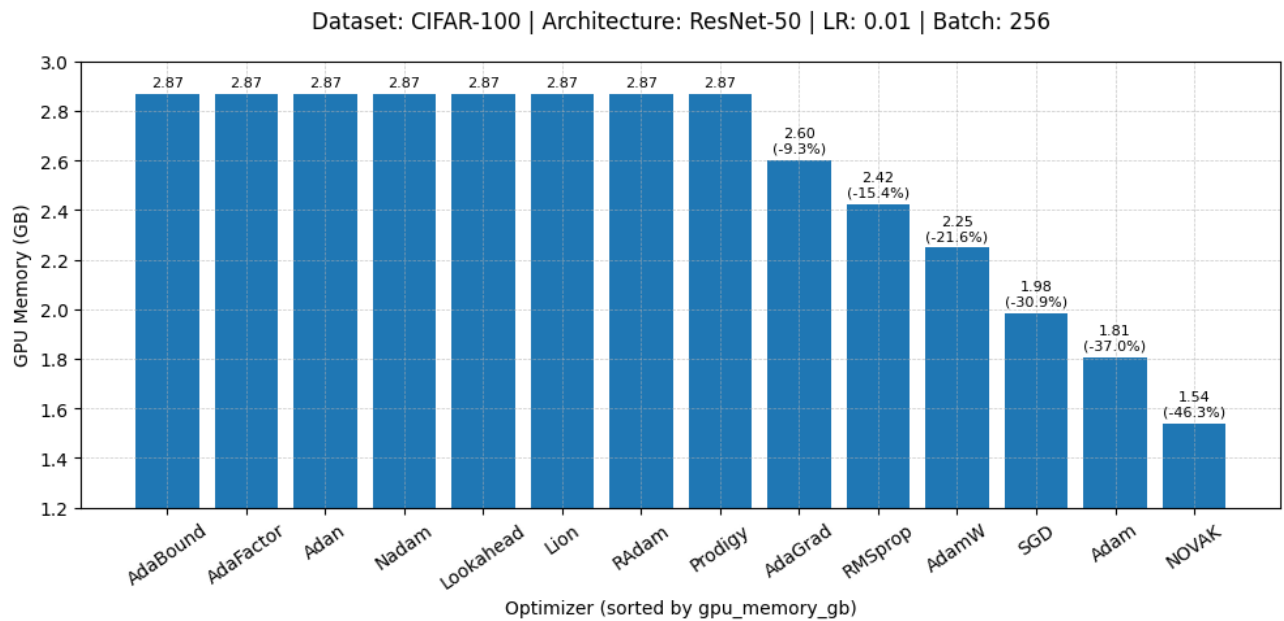

Fig. 7. GPU memory usage comparison across 14 optimizers on CIFAR-100 (ResNet-50, batch_size = 256, lr = 0.001). Optimizers sorted in descending order.

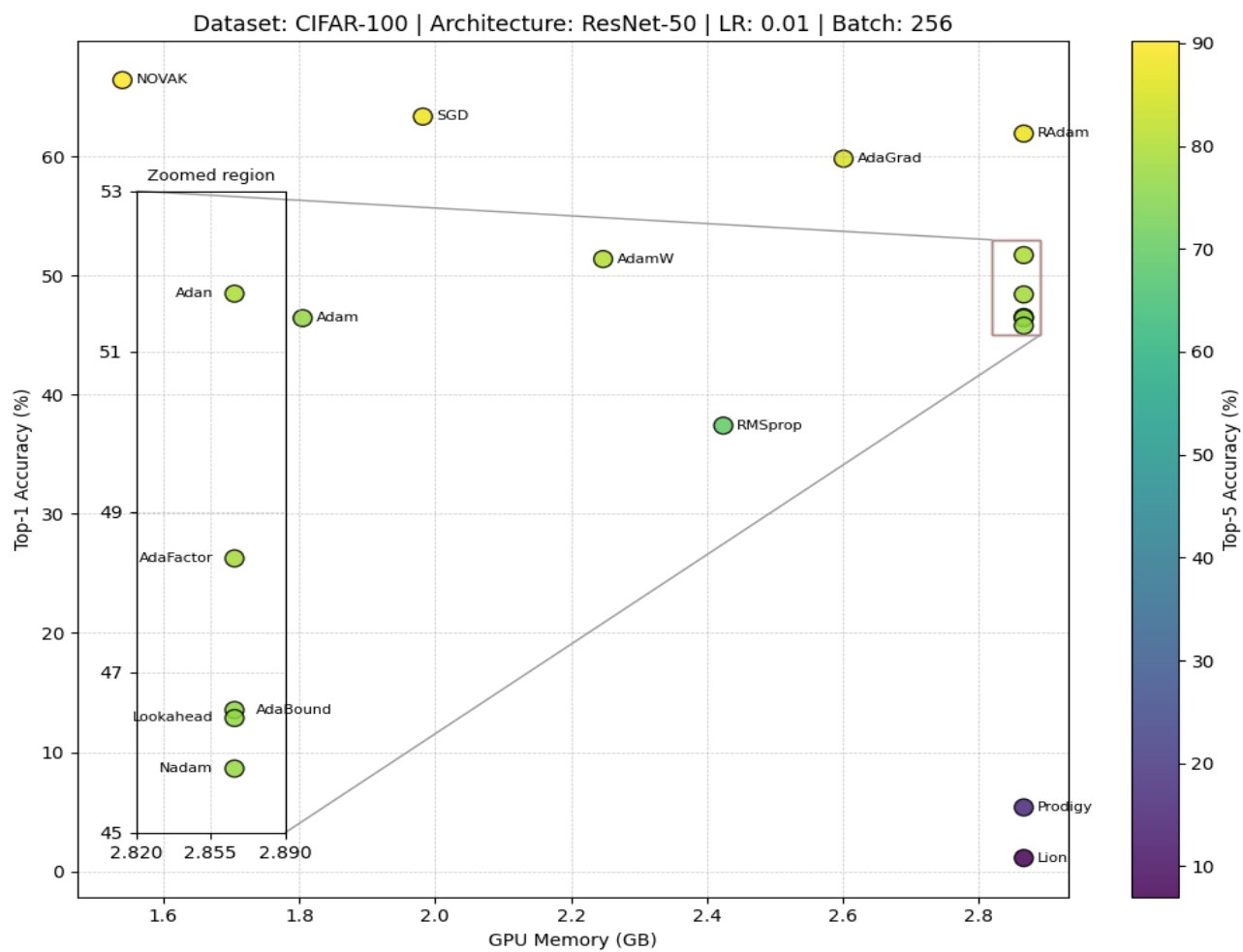

Fig. 8. Accuracy and GPU Memory trade-off comparison across 14 optimizers on CIFAR-100 (ResNet-50, batch_size = 256, lr = 0.001).

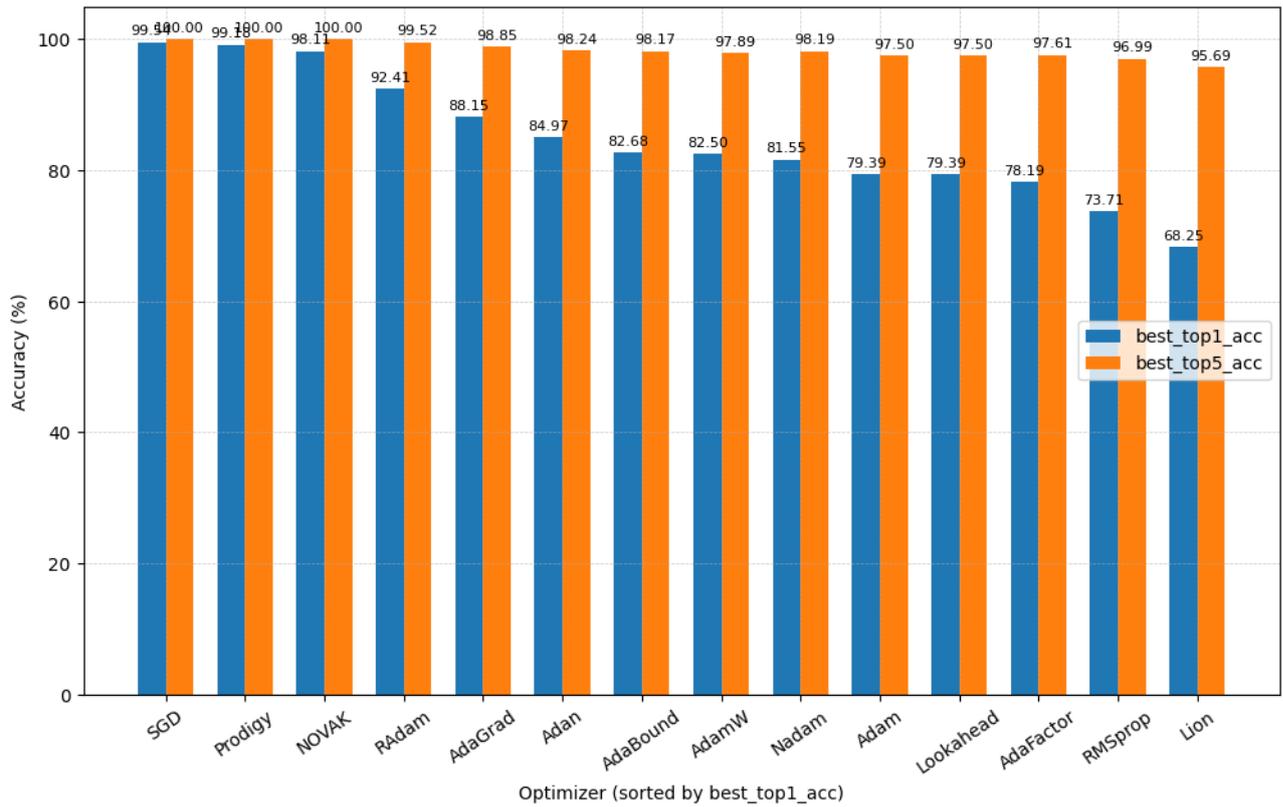

Fig. 9. Top-1 and Top-5 accuracy comparison across 14 optimizers on ImageNet (ResNet-50, batch_size = 128, lr = 0.01). Optimizers sorted by Top-1 accuracy in descending order.

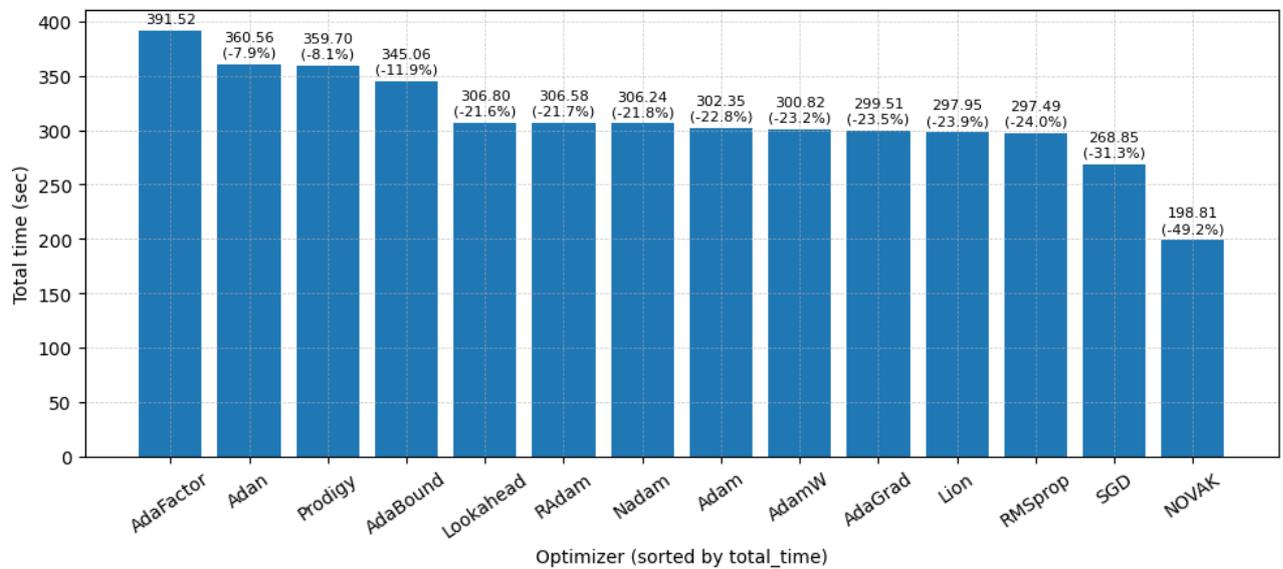

Fig. 10. Total time usage comparison across 14 optimizers on ImageNet (ResNet-50, batch_size = 128, lr = 0.01). Optimizers sorted in descending order.

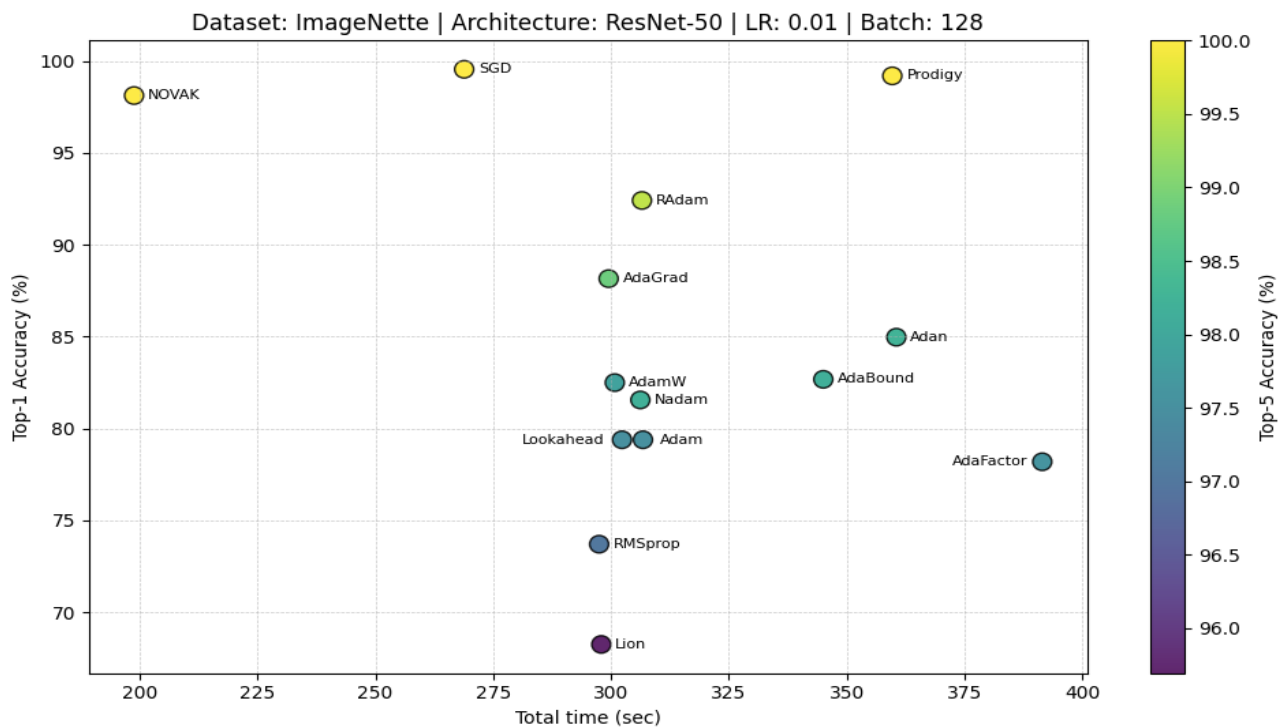

Fig. 11. Accuracy and Total time trade-off comparison across 14 optimizers on ImageNet (ResNet-50, batch_size = 128, lr = 0.01).

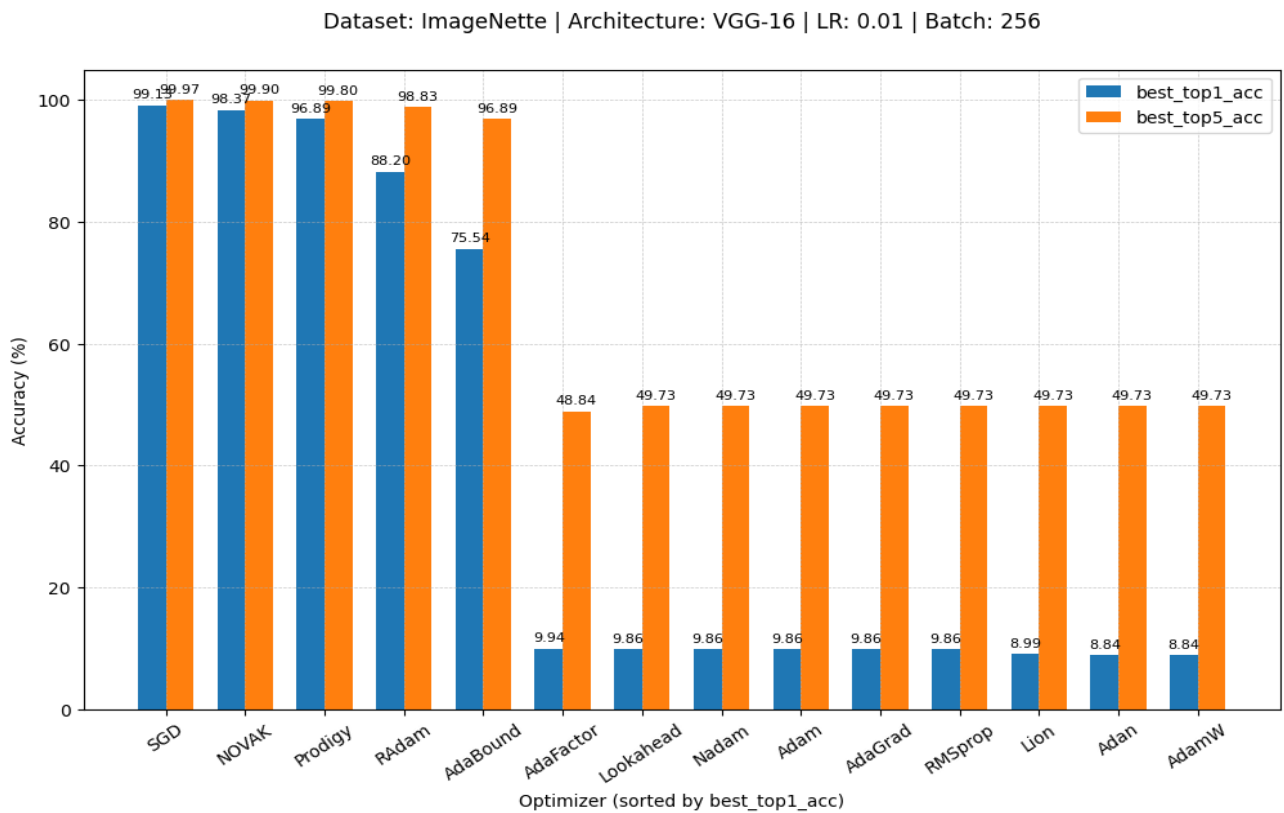

Fig. 12. Top-1 and Top-5 accuracy comparison across 14 optimizers on ImageNet (VGG16, batch_size = 256, lr = 0.01). Optimizers sorted by Top-1 accuracy in descending order.

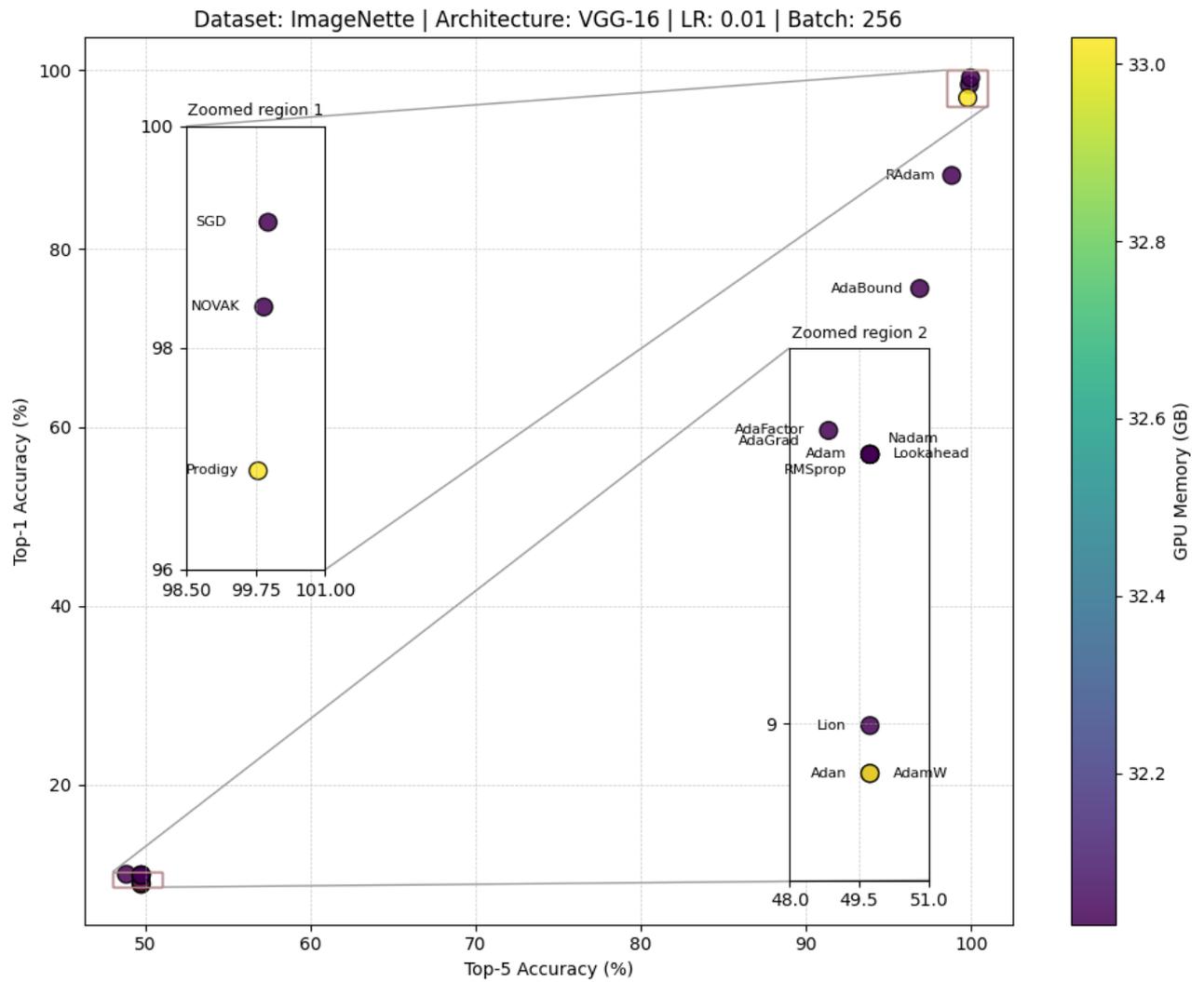

Fig. 13. Accuracy and GPU Memory trade-off comparison across 14 optimizers on ImageNet (VGG16, batch_size = 256, lr = 0.01).